\pdfoutput=1
\documentclass[11pt]{article}

\usepackage[final]{acl}

\usepackage{times}
\usepackage{latexsym}

\usepackage[T1]{fontenc}
\usepackage[utf8]{inputenc}

\usepackage{microtype}

\usepackage{inconsolata}

\usepackage{amsmath}
\usepackage{amsfonts}
\usepackage{multirow}
\usepackage{paralist}
\usepackage{enumitem}
\usepackage{graphicx}
\usepackage{booktabs}
\usepackage{float}
\usepackage{subcaption}

\title{
A Closer Look into Mixture-of-Experts in Large Language Models
}

\graphicspath{{./figures}}

\newcommand\blfootnote[1]{%
  \begin{NoHyper}%
  \renewcommand\thefootnote{}\footnote{#1}%
  \addtocounter{footnote}{-1}%
  \end{NoHyper}%
}

\author{Ka Man Lo $^{\ast}$ \\
  University of Macau \\\And
  Zeyu Huang $^{\ast}$ \\
  University of Edinburgh \\\And
  Zihan Qiu $^{\ast}$ \\
  Tsinghua University \\\AND
  Zili Wang \\
  INF Technology \\\And
  Jie Fu $^{\dag}$ \\
  Shanghai AI Lab \\}

\begin{document}

\maketitle

\begin{abstract}
Mixture-of-experts (MoE) is gaining increasing attention due to its unique properties and remarkable performance, especially for language tasks. 
By sparsely activating a subset of parameters for each token, MoE architecture could increase the model size without sacrificing computational efficiency, achieving a better trade-off between performance and training costs.
However, the underlying mechanism of MoE still lacks further exploration, and its modularization degree remains questionable.
In this paper, we make an initial attempt to understand the inner workings of MoE-based large language models. Concretely, we comprehensively study the parametric and behavioral features of four popular MoE-based models and reveal some intriguing observations, including 
\begin{inparaenum}[1)]
    \item Neurons act like fine-grained experts;
    \item The router of MoE usually selects experts with larger output norms;
    \item The expert diversity increases as the layer increases, while the last layer is an outlier, which is further validated by an initial experiment.
\end{inparaenum}
Based on the observations, we also provide suggestions for a broad spectrum of MoE practitioners, such as router design and expert allocation. We hope this work could shed light on future research on the MoE framework and other modular architectures.
Code is available at \url{https://github.com/kamanphoebe/Look-into-MoEs}.
\end{abstract}

\blfootnote{$^{\ast}$ Equal contribution.}
\blfootnote{$^{\dag}$ Corresponding author.}

\section{Introduction}

The advent of Large Language Models (LLMs) revolutionized the field of Natural Language Processing. 
LLM researchers are continually pushing the boundaries of Language Models by scaling up both model size and the column of training data, significantly enhancing the capabilities of these models.
This escalation in training cost and complexity necessitates innovative solutions to better balance between pre-training efficiency and model performance. 
One emerging solution to this end is the Mixture-of-Experts (MoE)~\cite{shazeer2017outrageously} architecture.
The MoE framework facilitates the computational efficiency of the model by dynamically routing inputs to a subset of experts, allowing for substantial model scaling while maintaining training costs and leading to numerous influential advancements in the field~\cite{reid2024gemini, jiang2024mixtral, dai2024deepseekmoe, qwen_moe}.

Beyond efficiency, another attractive trait of MoE architecture is its modular design and learning paradigm. 
This modularization allows for flexible and potentially more generalizable handling of diverse data and tasks within a single model by assigning them to specialized experts.
Despite its widespread adoption, it remains an open question whether current MoE-based LLMs truly leverage this modularity in knowledge distribution and expert behaviors.
In other words, is MoE a simple ensemble of homogeneous experts or a modular combination of heterogeneous experts? 
Answering this question comprehensively is non-trivial. 
Therefore, in this paper, we take the first step by investigating four popular MoE-based LLMs (Mixtral 8x7B~\cite{jiang2024mixtral}, Mixtral 8x22B, DeepSeekMoE~\cite{dai2024deepseekmoe}, and Grok-1\footnote{https://github.com/xai-org/grok-1}) from two critical perspectives: model parameters and model behaviors. 
We aim to explore common and distinct features and behaviors among different experts, further shedding light on the inner mechanisms of MoE-based models.

%这里概括一下我们具体做的事情，然后再把takeaway放到这里来，intro就差不多了。
Specifically, we examine the correlation between experts' parameters, gates, and their output features given text inputs.
Before diving into deeper analyses, we briefly summarize some of our empirical conclusions (detailed in \S~\ref{sec:discussion}) and observations:
\begin{itemize}[leftmargin=5mm, noitemsep]
    \item Neurons in the Feed-Forward Network (FFN) layer are fine-grained experts.
    Both the gate embedding matrix and the expert projection matrix $W_{\text{act}}$ perform the choosing operation: the former determines the expert selection while the latter controls the neuron activation.
    We observe that the similarity heat maps exhibit correlations, suggesting that, from the perspective of $W_{\text{act}}$, the expert neurons can be considered as ``tiny'' experts, each represented by a single neuron.
    \item Increasing the number of experts in deeper layers while reducing it in the last layer.
    This is experimented in Fig.~\ref{fig:dynamic_moe}.
    Our observations indicate that the similarities between the parameters and outputs of the experts consistently decrease with increasing layer number, followed by a sudden increase in the last layer.
    % This suggestion comes from the lower/higher similarities between experts' parameters and outputs observed in the deep/last layers.
    \item Using the norm as the routing mechanism is a reasonable choice. 
    For both Mixtral 8x7B and DeepSeekMoE, we observe that the gate typically selects experts with larger output norms.
    \item When analyzing the correlation between experts, measuring the similarities between weight matrices is, to some extent, equivalent to assessing the average similarities of expert outputs.
    % In our experiments, their similarity heat maps are akin to each other. 
    % \item Compared to specific initialization schemes, training MoE from scratch is more likely to facilitate expert diversity.
    % This stems from the observations that stronger correlations (\textit{e.g.}, higher similarities) exist between the parameters and behaviors of Mixtral experts. 
    % In contrast, DeepSeekMoE and Grok-1, which are trained from scratch, do not exhibit such correlations.
    \item Training MoE from scratch promotes greater expert diversity than specific initialization schemes. This stems from the observations that stronger correlations (\textit{e.g.}, higher similarities) between parameters and behaviors in Mixtral experts. In contrast, DeepSeekMoE and Grok-1, which are trained from scratch, do not show these correlations.
\end{itemize}

\section{Preliminary: Mixture-of-Experts}

Mixture-of-Experts models enhance transformers by replacing the original FFNs with \( N \) parallel FFNs combined with a router. 
These \( N \) FFNs are called experts and denoted as \( E_n \) for \( n \in [1, N] \). 
The gate \( g(\cdot; W_g, k) \), parameterized by \( W_g \) and an integer \( \operatorname{k} \), assigns the input \( x \) to a score distribution over the experts, \( g(x; W_g, \operatorname{k}) \in \mathbb{R}^N \). 
Typically, the gate \( g \) consists of a simple linear layer followed by a \(\operatorname{softmax}\) and a \(\operatorname{Top-k}\) function.

Given \( x \in \mathbb{R}^{d_{\text{hid}}} \), the output \( y \in \mathbb{R}^{d_{\text{hid}}} \) is the weighted sum of the outputs from all experts:

\[
y = \sum_{n \in N} g_n(x; W_g, \operatorname{k}) E_n(x)
\]

When \( \operatorname{k} \) for \(\operatorname{Top-k}\) is smaller than \( N \), only a subset of experts is involved in the computation. This is known as Sparse Mixture-of-Experts (SMoE).

The experts $E_n$ of the models investigated in this paper follow the style in LLaMA~\cite{touvron2023llama}, which consists of three linear layers and operates as (the subscript $n$ is omitted for brevity):
\begin{equation}
   E(x)=W_{\text{down}}(W_{\text{up}}x \odot \sigma(W_{\text{act}}x)) 
\end{equation}
where $\odot$ represents element-wise multiplication and $\sigma$ represents the activation function.
Given the three projection matrices \( W_{\text{up}}, W_{\text{act}} \in \mathbb{R}^{d_{\text{mid}} \times d_{\text{hid}}} \) and \( W_{\text{down}} \in \mathbb{R}^{d_{\text{hid}} \times d_{\text{mid}}} \), we define a neuron as the combination of the row vectors \( W_{\text{up}}[i, :] \) and \( W_{\text{act}}[i, :] \), along with the column vector \( W_{\text{down}}[:, i] \). Thus, each expert contains \( d_{\text{mid}} \) neurons, each with size $d_{\text{hid}}$.

\section{Overview}

\begin{table*}[thbp]
    \small
    \centering
    % % \vskip -0.3in
    \begin{tabular}{llccccc}
        \hline
        \multirow{2}{*}{\textbf{Model}} & \multirow{2}{*}{\textbf{Abbreviation}} & \multirow{2}{*}{\textbf{\# MoE layers}} & \multirow{2}{*}{\textbf{\# experts}} & \multirow{2}{*}{\textbf{Top-k}} & \textbf{Hidden size} & \textbf{Intermediate size} \\
        & & & & & ($d_{\text{hid}}$) & ($d_{\text{mid}}$) \\
        \hline
        Mixtral 8x7B & Mixtral & 32 & 8 & 2 & 4096 & 14336 \\
        Mixtral 8x22B & Mixtral-22 & 56 & 8 & 2 & 6144 & 16384 \\
        Mistral 7B & Mistral & 32 & N/A & N/A & 4096 & 14336 \\
        DeepSeekMoE & DeepSeek & 27 & 64 routed + 2 shared & 6 & 2048 & 1408 \\
        Grok-1 & Grok & 64 & 8 & 2 & 6144 & 32768 \\
        \hline
    \end{tabular}
    % \vskip -0.1in
    \caption{\footnotesize Basic information of models used for analysis. The abbreviations are used throughout our paper.}
    \label{tab:models}
    % \vskip -0.2in
\end{table*}

Our experiments are conducted on several open-source MoE models, namely Mixtral 8x7B, Mixtral 8x22B~\footnote{For the Mixtral 8x22B model, we only conduct most of the analyses mentioned in the main context (excluding those in the appendix) due to time limit.}, DeepSeekMoE, and Grok-1.
We choose these models due to their widespread use and impressive performance across various domains.
Additionally, they exhibit complementary characteristics across several key attributes, enabling a robust comparative analysis using control variables.
Details are discussed in Append~\ref{append:models}.
To further study the similarities and differences between a standard transformer and a MoE model, we include Mistral 7B~\cite{jiang2023mistral} as one of our investigated models. 
Basic information about these models, along with the abbreviations used throughout our paper, is summarized in Tab.~\ref{tab:models} and Tab.~\ref{tab:models2}.
The analysis is divided into two sections: one focusing on model parameters (\textbf{static}) and the other on model behaviors in response to text input (\textbf{dynamic}).
% TODO: explain from a high-level perspective why we choose to investigate them and their connections.

\textit{Unless otherwise stated (\S~\ref{exp:out_sim}), cosine similarity is employed for all experiments involving similarity measurements.}
While we acknowledge the existence of other metrics, we primarily use cosine similarity as it is a widely adopted approach~\cite{sun2024transformer, zhang2021moefication}.
% Due to the page limit, we have included only part of the figures in the main context; please refer to the appendix for all the graphs.

\section{Analysis of Static Parameters}

% In this section, we study MoE experts from three different perspectives, including weight matrices, neurons, and hidden states.
From a high-level perspective, a model's knowledge is encoded in its parameters, making the investigation of weight matrices a natural approach.
In this section, we study the correlation between the parameters of:
\begin{inparaenum}[i)]
    \item MoE experts (and FFNs for Mistral), and
    \item gate embeddings;
\end{inparaenum}
which are two vital components of the MoE architecture.

\subsection{Weight Matrices of Experts} 
\label{sec:expert-matrix}

% The experts of our chosen MoE models follow the structure of feed-forward networks (FFNs) in LLaMA~\cite{touvron2023llama}, which consists of three linear layers and performs operations as:
% \begin{equation}
%    \text{Expert}(x)=W_{\text{down}}(W_{\text{up}}x \odot \operatorname{Act}(W_{\text{act}}x)) 
% \end{equation}
% where $\odot$ denotes element-wise multiplication and the activation function is abbreviated as $\operatorname{Act}$.
% Given the three weight matrices \( W_{\text{up}}, W_{\text{act}} \in \mathbb{R}^{d_{\text{mid}} \times d_{\text{hid}}} \) and \( W_{\text{down}} \in \mathbb{R}^{d_{\text{hid}} \times d_{\text{mid}}} \), we define a neuron as the combination of the row vectors \( W_{\text{up}}[i, :] \) and \( W_{\text{act}}[i, :] \), along with the column vector \( W_{\text{down}}[:, i] \). Thus, each expert contains \( d_{\text{mid}} \) neurons.
% we state that the expert comprises $d_{mid}$ neurons.
% Hence, a neuron is constructed by a row vector of $W_{\text{up}}$ and of $W_{\text{act}}$, and a column vector of $W_{\text{down}}$.

MoE models replace FFNs in standard transformers with experts.
Following~\citet{geva2020transformer, qiu2024empirical}, the projection matrices of the experts can be regarded as keys and values:
the column vectors of $W_{\text{down}}$ represent potential outputs; 
the row vectors of $W_{\text{up}}$ produce weights for each possible output; 
the row vectors of $W_{\text{act}}$ determine whether to activate the corresponding neurons.
Thus, examining the weight matrices provides a straightforward way to understand the expert behaviors. 
We analyze both the matrix and neuron levels to gain insights from different perspectives.

\subsubsection{Matrix-level} \label{exp:mat_sim}

\begin{figure*}[thbp]
    % \vskip -0.3in
    \centering
    $\vcenter{\hbox{\includegraphics[width=.09\linewidth]{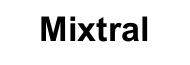} }}$
    $\vcenter{\hbox{\includegraphics[width=.36\linewidth]{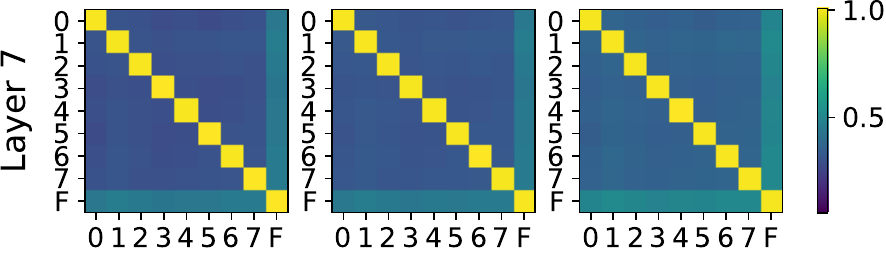} }}$
    $\vcenter{\hbox{\includegraphics[width=.36\linewidth]{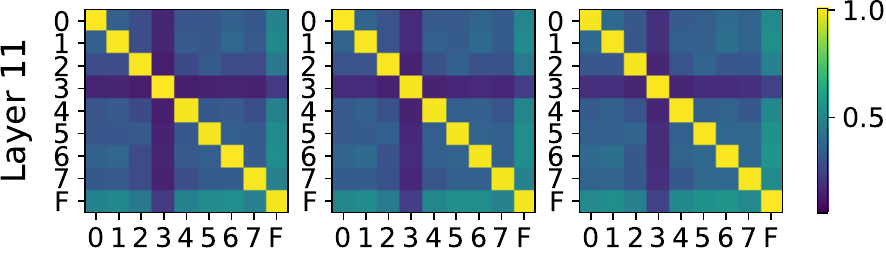}}}$ \\
    $\vcenter{\hbox{\includegraphics[width=.09\linewidth]{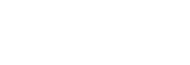} }}$ 
    $\vcenter{\hbox{\includegraphics[width=.36\linewidth]{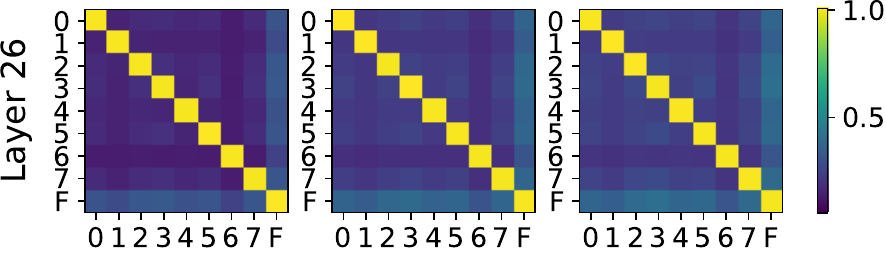} }}$
    $\vcenter{\hbox{\includegraphics[width=.36\linewidth]{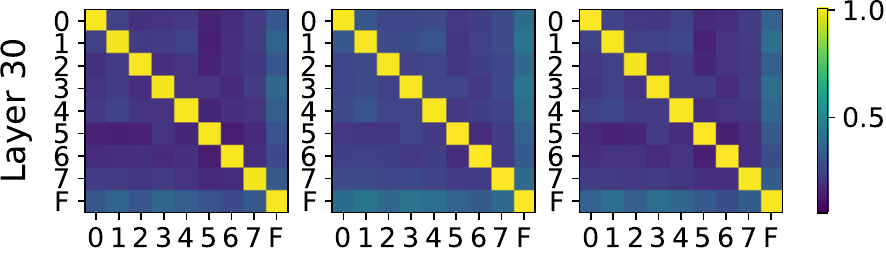}}}$ \\
    \rule{\linewidth}{0.5pt}\vspace{1mm} \\
    $\vcenter{\hbox{\includegraphics[width=.09\linewidth]{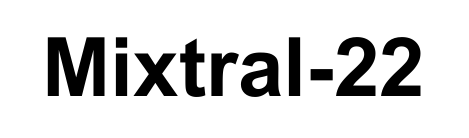} }} $
    $\vcenter{\hbox{\includegraphics[width=.36\linewidth]{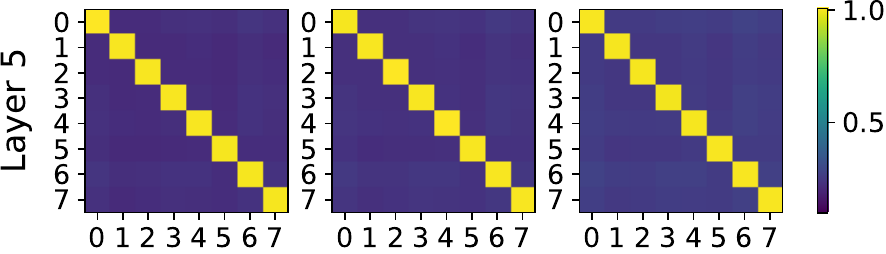} }}$
    $\vcenter{\hbox{\includegraphics[width=.36\linewidth]{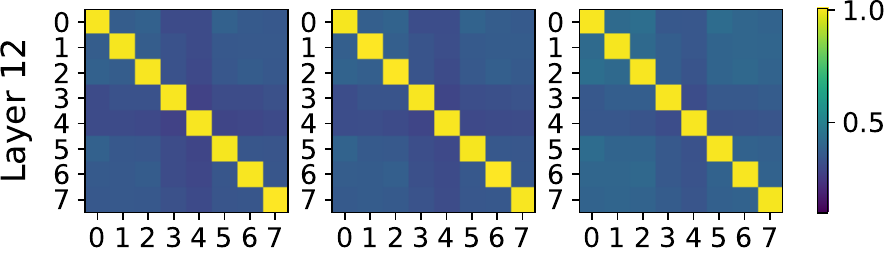}}}$ \\
    $\vcenter{\hbox{\includegraphics[width=.09\linewidth]{blank.pdf} }}$ 
    $\vcenter{\hbox{\includegraphics[width=.36\linewidth]{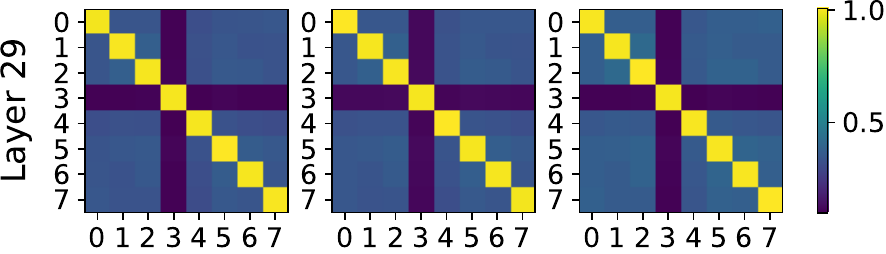} }}$
    $\vcenter{\hbox{\includegraphics[width=.36\linewidth]{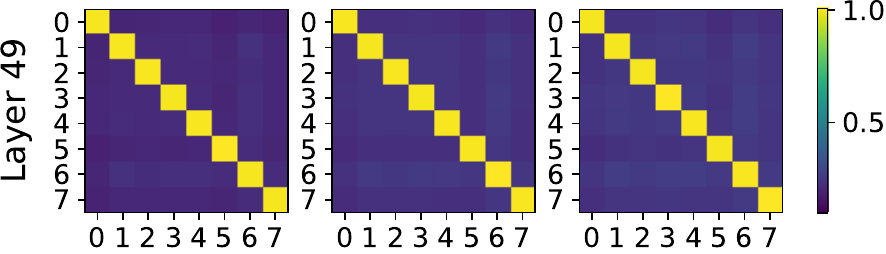}}}$ \\
    \rule{\linewidth}{0.5pt}\vspace{1mm} \\
    $\vcenter{\hbox{\includegraphics[width=.09\linewidth]{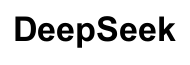} }} $
    $\vcenter{\hbox{\includegraphics[width=.35\linewidth]{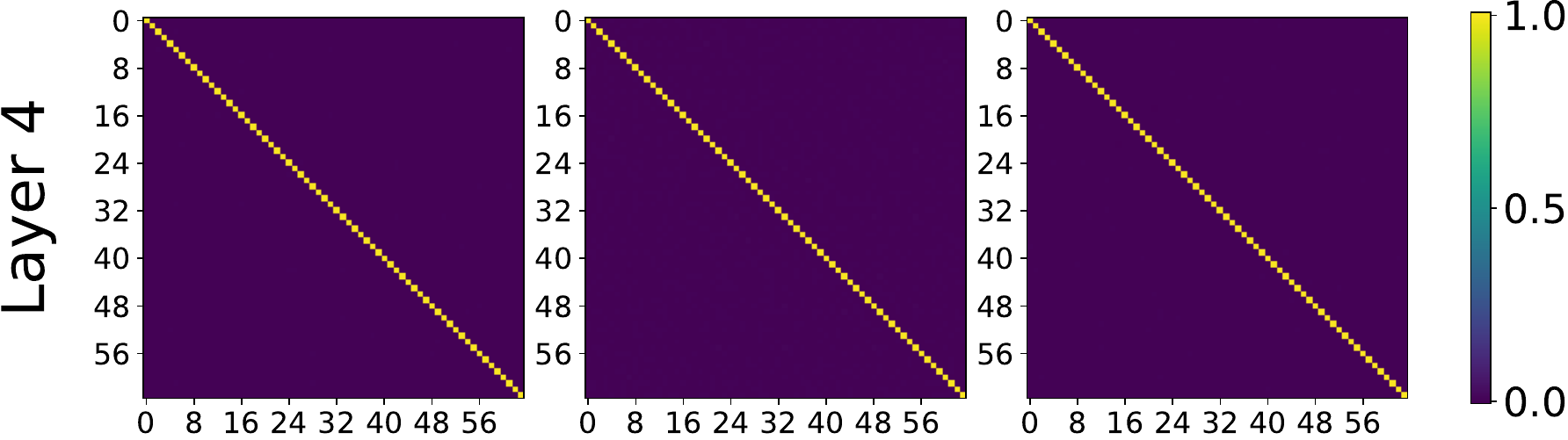} }}$ \hspace{1mm}
    $\vcenter{\hbox{\includegraphics[width=.35\linewidth]{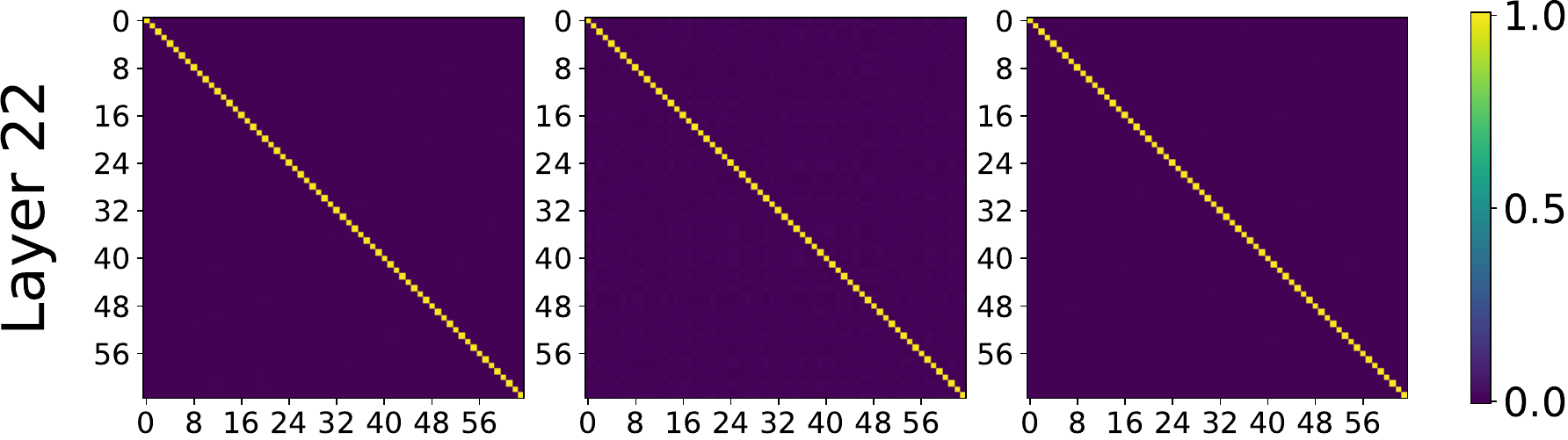}}}$ \\
    \vspace{1mm}\rule{\linewidth}{0.5pt}\vspace{1mm} \\
    $\vcenter{\hbox{\includegraphics[width=.09\linewidth]{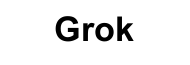} }}$
    $\vcenter{\hbox{\includegraphics[width=.36\linewidth]{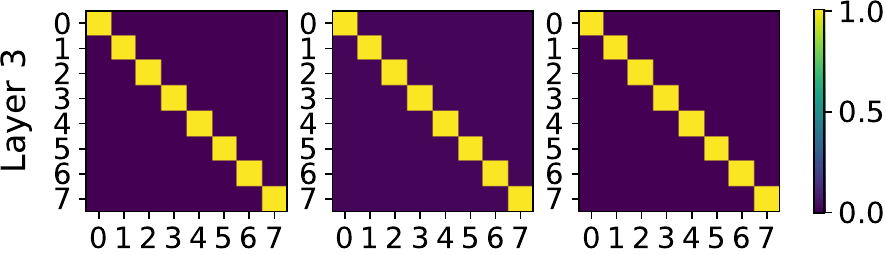} }}$
    $\vcenter{\hbox{\includegraphics[width=.36\linewidth]{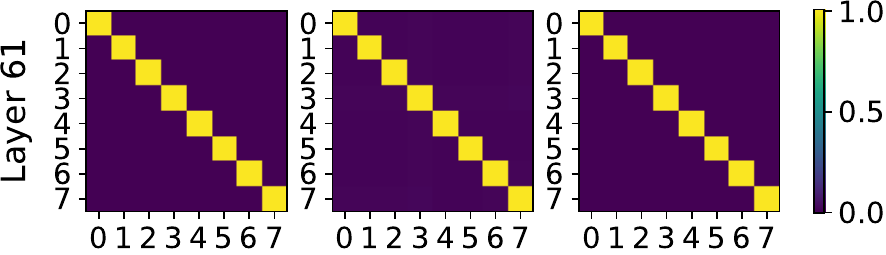}} }$ \\
    % \vskip -0.1in
    \caption{\footnotesize Matrix-level similarity heat maps of expert weight matrices. Each layer contains three heat maps, corresponding to $W_{\text{up}}$, $W_{\text{act}}$, and $W_{\text{down}}$, respectively. The tick numbers refer to the expert indices. ``F'' denotes the Mistral FFN.}
    \label{fig:mat_sim}
\end{figure*}

In this part, we explore the similarity of the three projection matrices $W_{\text{up}}$, $W_{\text{act}}$, and $W_{\text{down}}$ among all experts in each layer. 
The similarity is calculated based on the flattened matrices and is illustrated in Fig.~\ref{fig:mat_sim}.
We denote ``F'' as the Mistral FFN and ``SE'' as the DeepSeek shared expert.
\textit{Note that the figures for different models do not share the same color bar.}

\noindent\textbf{Common~\footnote{The observations shared by \textit{all} of our investigated models are written in the Common part.}.}
The heat maps of the three matrices exhibit similar patterns.
Directly flattening the large weight matrices leads to high-dimensional vectors, so we use principal components analysis (PCA) to reduce these vectors to two-dimensional space.
The resulting figures also show that, for Mixtral and DeepSeek, the expert distribution across the three weight matrices is generally comparable.
Details on the PCA results are presented in Append~\ref{append:mat_pca}.

\noindent\textbf{Mixtrals and Mistral.}
The cosine similarities between Mixtral experts ($S_{\text{ee}}$) primarily range from 0.2 to 0.4, while the similarities between the experts and the Mistral FFN ($S_{\text{ef}}$) are about 0.6.
Yet the values tend to be lower in the deeper layers (22$^{\text{nd}}$-30$^{\text{th}}$ for Mixtral and  35$^{\text{th}}$-50$^{\text{th}}$ for Mixtral-22).
A ``dark cross'' can be observed in some layers and corresponds to outliers in the 2D space projected by PCA, indicating that the associated expert is relatively distinct from the others.
Interestingly, this cross appears most frequently in Expert 3 for Mixtral, suggesting that this expert may have learned some unique attributes.
It is noteworthy that the cross usually extends across the entire heat map, including the last row of the FFN.
Thus, when an Mixtral expert differs from other experts, it is also less similar to the Mistral FFN. 

\noindent\textbf{DeepSeek and Grok.}
The shared experts of DeepSeek are implemented as a single MLP block with a larger hidden size than the routed experts, preventing direct comparison of their flattened vectors; thus, we omit them from this experiment.
Fig.~\ref{fig:mat_sim} demonstrates that the similarities between the DeepSeek routed experts and Grok experts are close to zero.
While Mixtrals' training method remains unrevealed, it is known that DeepSeek and Grok are trained from scratch.
This suggests that Mixtrals may have been trained using special schemes, resulting in less diverse experts compared to those trained from scratch~\cite{wu2022residual}.

\subsubsection{Neuron-level} \label{exp:neuron_sim}

% Examining neurons can facilitate the understanding regarding experts from a fine-grained perspective. 
In \S~\ref{sec:expert-matrix}, we measure the parameter similarity between experts at the matrix level.
However, the calculation of cosine similarity is position-dependent.
If the neurons of two experts are similar but in different orders, the similarity of their weight matrices will be significantly lower than expected.
To address this, we propose two approaches to investigate the correlation at the neuron level: averaging and reordering.
Averaging simply averages the rows (for $W_{\text{up}}$ and $W_{\text{act}}$) or the columns (for $W_{\text{down}}$) of the weight matrices, and then calculates the cosine similarity of the resulting vectors across experts.
For reordering, we apply the Jonker-Volgenant algorithm~\cite{jonker1988shortest}, which is typically used for solving linear assignment problems, to find the optimal order of neurons so that the cosine similarity between two experts is maximized.

We describe the results of the reordering method below and provide the details of the averaging approach in Append~\ref{append:averaging}.
Additionally, the projection of neurons into low-dimensional spaces using PCA can be found in Append~\ref{append:neuron_pca}.
Due to the heavy computation, we only select several layers for the reordering calculation.
Note that the matrices are reordered separately.
We measure Kendall's $\tau$ coefficient between the index sequences before and after reordering, whose value increases when the two sequences exist strong agreement.
Tab.~\ref{tab:reorder} depicts the common similarity growth after reordering and the average Kendall's coefficient $\bar\tau$ over the selected layers. 
The order of Mixtral neurons changes little (resulting in a large $\tau$), and hence nearly unchanged similarities.
Despite the substantial similarity increase for DeepSeek and Grok after reordering, their overall values remain around 1e-2.

% \textbf{Mixtral and Mistral.}
% The order of neurons hardly changes, and the similarities grow at the order of 1e-3, meaning that the original ordering almost attains the highest similarities.
% % As a result, the heat maps for three matrices are nearly identical to the figures in \S~\ref{exp:mat_sim}.

% \textbf{DeepSeek and Grok.}
% Reordering the neurons increases the similarities of all matrices from the order of 1e-4 to 1e-2.
% Despite the vast increment, the overall values are still very small.

\begin{table}[htbp]
    \small
    % \vskip -0.05in
    \centering
    \begin{tabular}{lcc}
        \toprule
        \textbf{Model} & \textbf{Order of Growth} & $\bar\tau$ \\
        \midrule
        Mixtral & 1e-3 & 0.75 \\
        DeepSeek & 100 & -0.0002 \\
        Grok & 100 & -0.0003 \\
        \bottomrule
    \end{tabular}
    % \vskip -0.1in
    \caption{\footnotesize Reordering results of expert neurons.}
    \label{tab:reorder}
    % \vskip -0.1in
\end{table}

\subsection{Gate Embedding}

The gate embedding of our investigated MoE models is implemented as a linear layer $W_g$ with size $\mathbb{R}^{N}\times\mathbb{R}^{d_{\text{hid}}}$, where $N$ is the number of experts.
% Similar to the definition of expert neuron, we define a gate neuron as a row vector of the embedding matrix, hence $n_{exp}$ neurons in a gate in total.
The gate serves as a crucial component of MoE, making it essential to study its attributes to understand MoE functionality better.
In addition, since each row vector in the gate embedding determines expert selection, some correspondence may exist between $W_g$ and the expert weights. 
% some relationship may exist between the embedding matrix and the experts. 

To investigate this, we measure the similarities between the gate embedding vectors $W_g[n,:]$ for $n \in [1, N]$.
For computational simplicity, we compare them with the neuron-level averaging (instead of the reordering) heat maps of experts presented in Append~\ref{append:averaging}, with qualitative analyses detailed in Append~\ref{append:gate_sim}.
Specifically, we found that, for all four MoE models, the patterns in the heat maps of gate vectors and of expert neurons $W_{\text{act}}[i,:]$ are partially alike in some layers (\textit{i.e.}, the same coordinates in both heat maps exhibit relatively higher or lower values simultaneously).

Therefore, we further conduct a quantitative analysis of their similarity values.
In particular, we perform linear regression on the paired similarity dataset $(X, Y)$, where $X$ denotes the similarities of $W_g[n,:]$, and $Y$ denotes the neuron-level similarities of $W_{\text{up}}$, $W_{\text{act}}$, or $W_{\text{down}}$.
Tab.~\ref{tab:r2} describes the average square of Pearson correlation coefficients over all layers ($R_{\text{avg}}^2$), while Tab.~\ref{tab:R} lists the Pearson correlation coefficient ($R$) for each layer.
As shown in Tab.~\ref{tab:r2}, the correlation between the similarities of the gate vectors and those of $W_{\text{act}}$ is significantly stronger than that with $W_{\text{up}}$ and $W_{\text{down}}$.
For the $(X, Y_{\text{act}})$ pair, although Mixtral and DeepSeek have similar $R_{\text{avg}}^2$ values, the $R^2$ of Mixtrals fluctuate between 0.1 and 0.7, while the $R^2$ of DeepSeek remains close to 0.4.
Furthermore, we can see from Tab.~\ref{tab:R} that $(X, Y_{\text{act}})$ for both Mixtral and DeepSeek show positive correlations, whereas $(X, Y_{\text{act}})$ for Grok turn to negative correlations starting from the intermediate (after 25$^{\text{th}}$) layers.
We note that the function of $W_g$ and $W_{\text{act}}$ is analogous: the former determines expert selection while the latter is responsible for choosing which neurons to activate.
Therefore, they may learn similar knowledge to effectively perform the \textit{choosing} operation, which explains the observed correlation.

\begin{table}[thbp]
    \small
    \centering
    \begin{tabular}{lccc}
        \toprule
        \textbf{Model} & $(X, Y_{\text{up}})$ & $(X, Y_{\text{act}})$ & $(X, Y_{\text{down}})$ \\
        \midrule
        Mixtral & 0.06 & 0.33 & 0.07 \\
        Mixtral-22 & 0.13 & 0.26 & 0.13 \\
        DeepSeek & 0.00 & 0.40 & 0.00 \\
        Grok & 0.04 & 0.15 & 0.04 \\
        \bottomrule
    \end{tabular}
    % \vskip -0.1in
    \caption{\footnotesize Average square of Pearson correlation coefficients over all layers ($R_{\text{avg}}^2$) for three paired dataset.}
    \label{tab:r2}
    % \vskip -0.1in
\end{table}

\subsection{Summary}
\label{sec:summary-static}

Here, we conclude the key observations from the analysis of static parameters:
\begin{inparaenum}[i)]
    \textbf{\item} Mixtral might contain expert(s) with unique attributes, as evidenced by the frequent presence of dark crosses in Fig.~\ref{fig:mat_sim}.
    \textbf{\item} The similarities of DeepSeek and Grok expert weight matrices are generally lower than those in Mixtrals. 
    As mentioned in \S~\ref{exp:mat_sim}, the matrix-level similarities of DeepSeek and Grok experts are typically close to zero, whereas Mixtrals' expert similarities average around 0.3. 
    \textbf{\item} The weights of different experts become less similar in deeper layers, as observed in the Mixtrals' heat maps in Fig.~\ref{fig:mat_sim}.
    \textbf{\item} $W_{\text{up}}$, $W_{\text{down}}$, and $W_{\text{act}}$, share similar patterns in their similarity heat maps (Fig.~\ref{fig:mat_sim}).
    \textbf{\item} The similarities of $W_g$ and of $W_{\text{act}}$ show either positive or negative association.
    Tab.~\ref{tab:r2} depicts the $R_{\text{avg}}^2$ values, where the pairing of $W_g$ and $W_{\text{act}}$ achieves the highest correlation across all four models.
\end{inparaenum}

% \begin{itemize}[leftmargin=5mm, noitemsep]
%     \item Mixtral might contain expert(s) with special attributes. 
%     Dark crosses can be frequently found in Fig.~\ref{fig:mat_sim}.
%     \item The similarities of DeepSeek and Grok expert weight matrices are generally lower than those of Mixtral. 
%     As mentioned in \S~\ref{exp:mat_sim}, the matrix-level similarities of DeepSeek and Grok experts are usually zero, whereas the Mixtral expert similarities reach about 0.3 on average. 
%     \item Different experts' weights are less similar in deep layers.
%     This can be observed from the Mixtral heat maps in Fig.~\ref{fig:mat_sim}.
%     \item $W_{\text{up}}$, $W_{\text{down}}$, and $W_{\text{act}}$, share similar patterns in their similarity heat maps (Fig.~\ref{fig:mat_sim}).
%     \item The similarities of the gate embeddings and of $W_{\text{act}}$ show either positive or negative association.
%     Tab.~\ref{tab:r2} depicts the $R_{avg}^2$ values, where the gate embeddings and $W_{\text{act}}$ pair achieve the highest for all three models. 
% \end{itemize}

\section{Analysis of Dynamic Behaviours}

\begin{figure*}[htbp]
% \vskip -0.3in 
    \centering
    $\vcenter{\hbox{\includegraphics[width=.08\linewidth]{mixtral.pdf} \hspace{3.cm}}} $
    $\vcenter{\hbox{\includegraphics[width=.08\linewidth]{mixtral_22.pdf} \hspace{3.cm}}} $
    $\vcenter{\hbox{\includegraphics[width=.08\linewidth]{deepseek.pdf} }} $ \\
    $\vcenter{\hbox{\includegraphics[width=.32\linewidth]{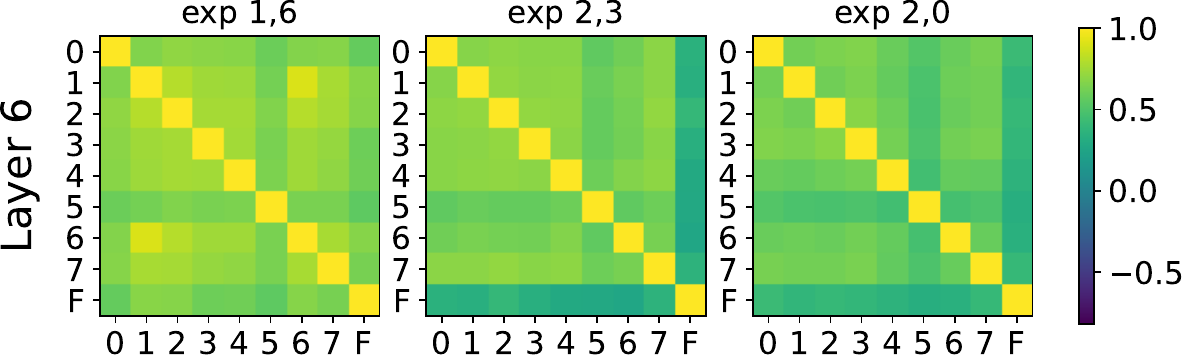}}}$
    $\vcenter{\hbox{\includegraphics[width=.32\linewidth]{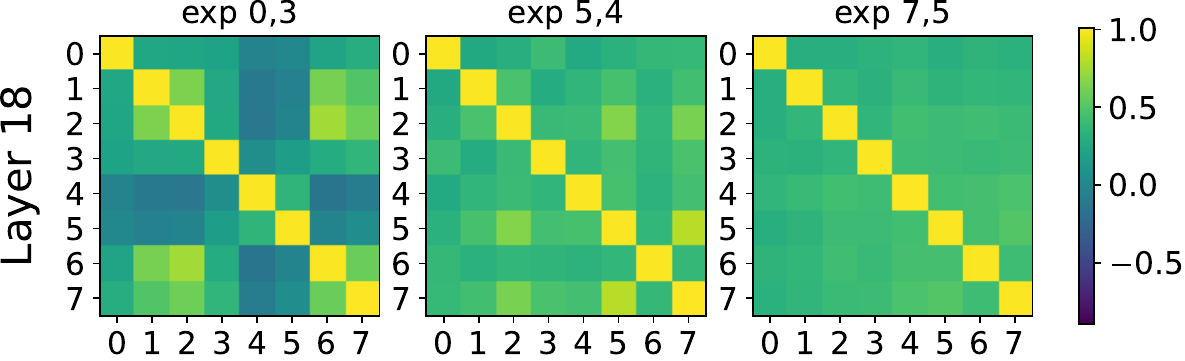}}}$
    $\vcenter{\hbox{\includegraphics[width=.32\linewidth]{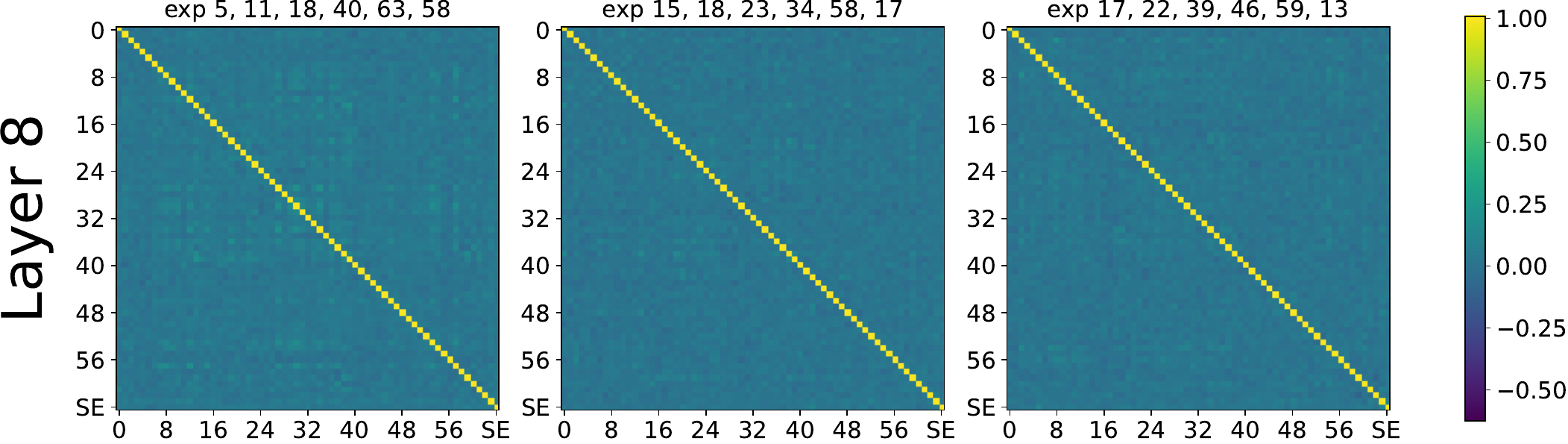}}}$ \\
    $\vcenter{\hbox{\includegraphics[width=.32\linewidth]{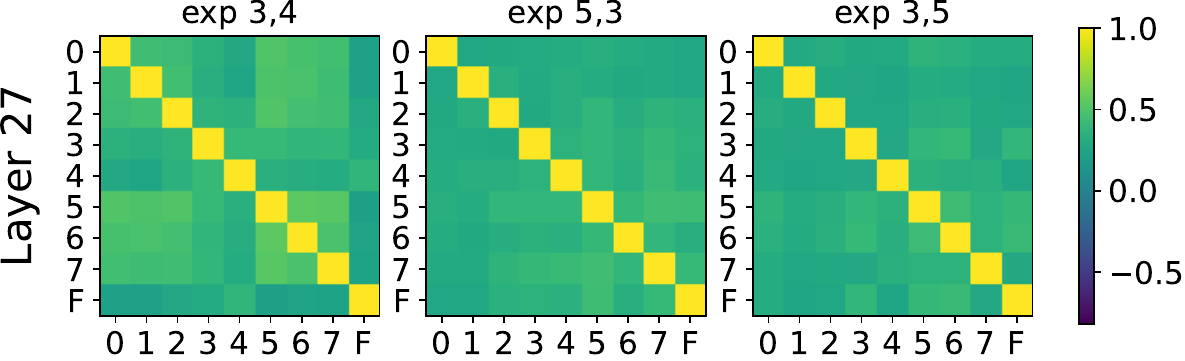}}}$
    $\vcenter{\hbox{\includegraphics[width=.32\linewidth]{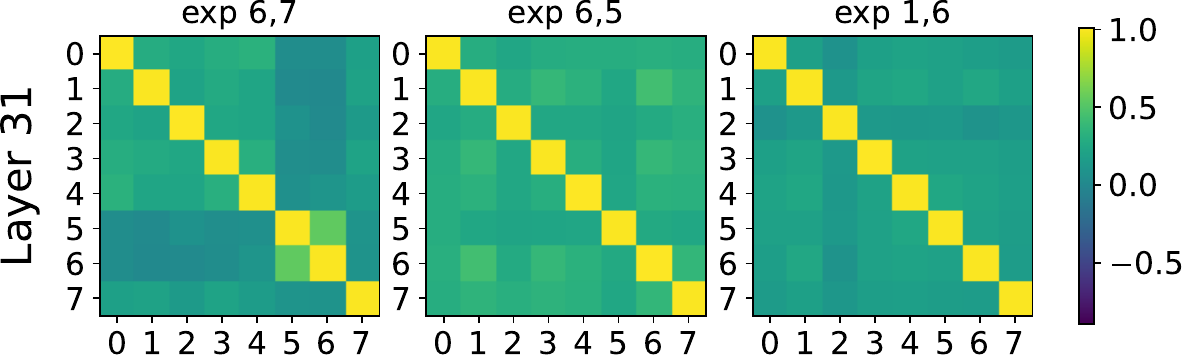}}}$ 
    $\vcenter{\hbox{\includegraphics[width=.32\linewidth]{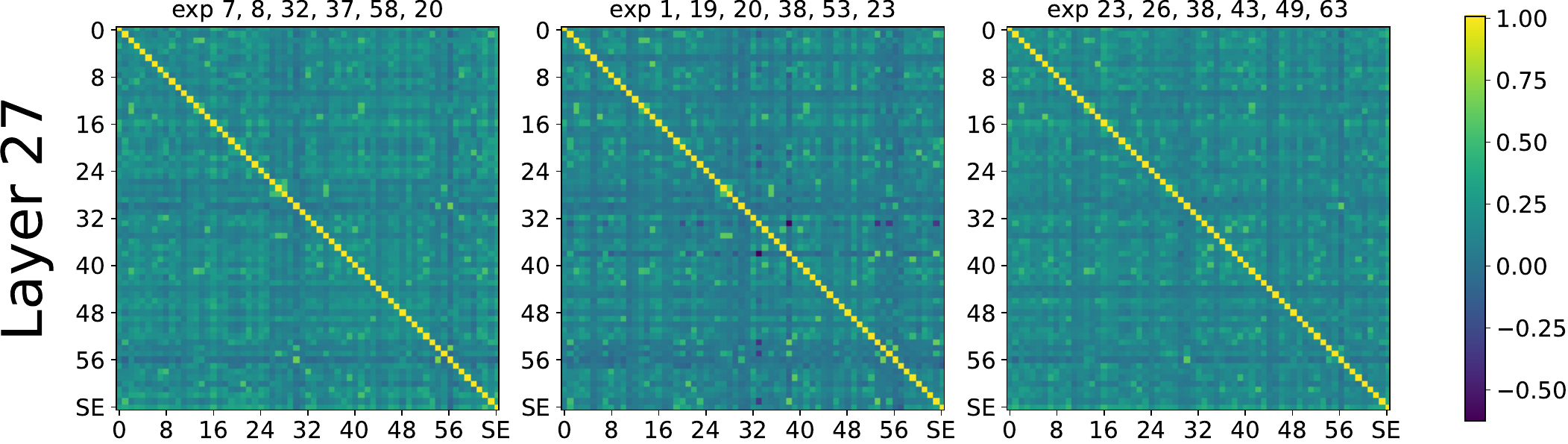}}}$ \\
    $\vcenter{\hbox{\includegraphics[width=.08\linewidth]{blank.pdf} \hspace{9.cm}}} $
    $\vcenter{\hbox{\includegraphics[width=.08\linewidth]{grok.pdf} }} $ \\
    \hspace{.5mm} $\vcenter{\hbox{\includegraphics[width=.32\linewidth]{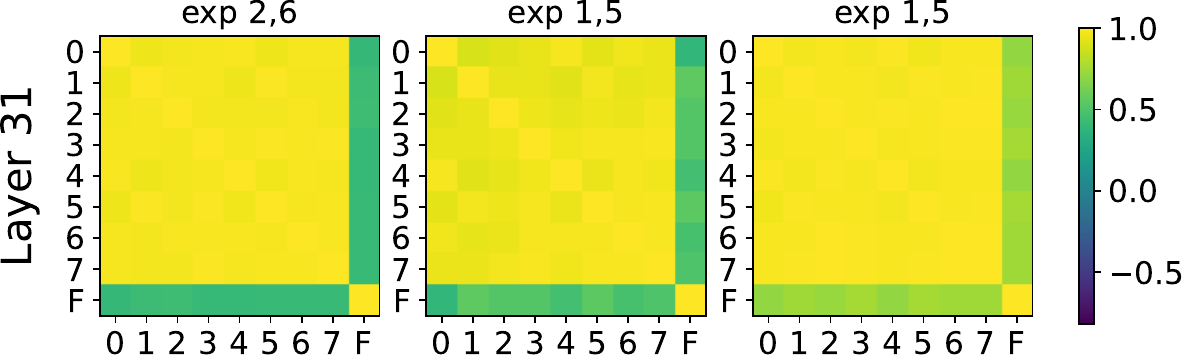}}}$
    $\vcenter{\hbox{\includegraphics[width=.32\linewidth]{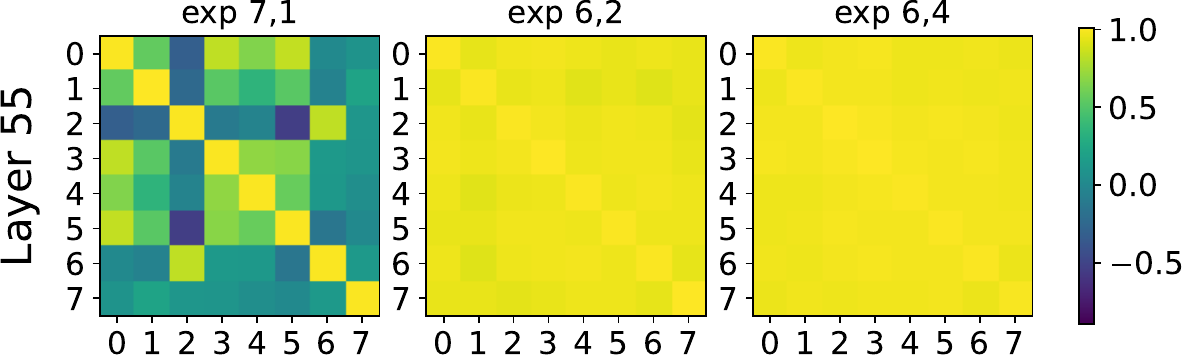}}}$
    $\vcenter{\hbox{\includegraphics[width=.33\linewidth]{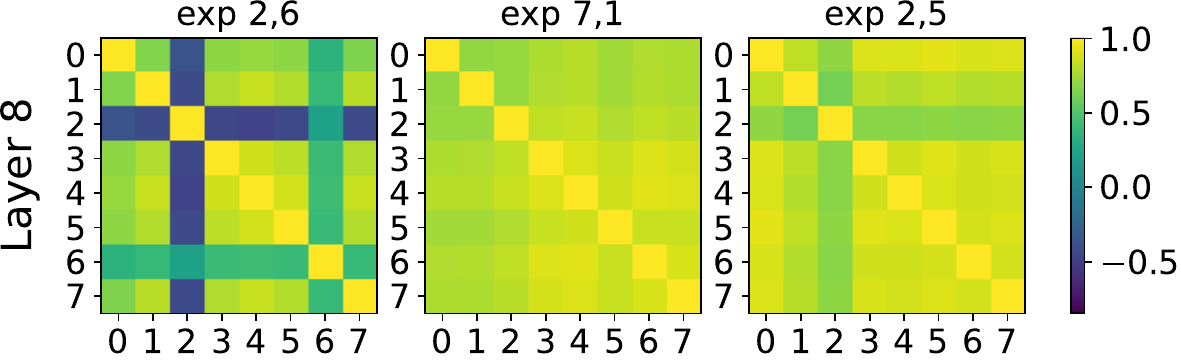}}}$ \\
    % \vskip -0.1in 
    \caption{\footnotesize Similarity heat maps of expert output features using the short input. The top $k$ experts for each token are shown on top of each heat map. The tick numbers refer to the expert indices. ``F'' and ``SE'' denote the Mistral FFN and the DeepSeek shared expert, respectively.}
    \label{fig:out_sim}
\end{figure*}

\begin{figure*}[t!]
% \vskip -0.1in 
    \centering
    $\vcenter{\hbox{\includegraphics[width=.08\linewidth]{mixtral.pdf} }}$ \\
    $\vcenter{\hbox{\includegraphics[width=.14\linewidth]{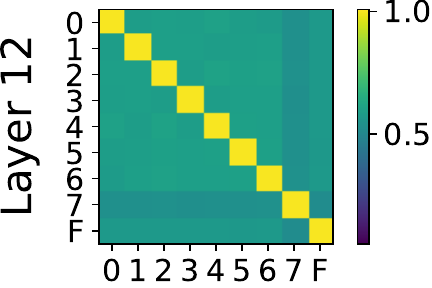} }}$ 
    $\vcenter{\hbox{\includegraphics[width=.33\linewidth]{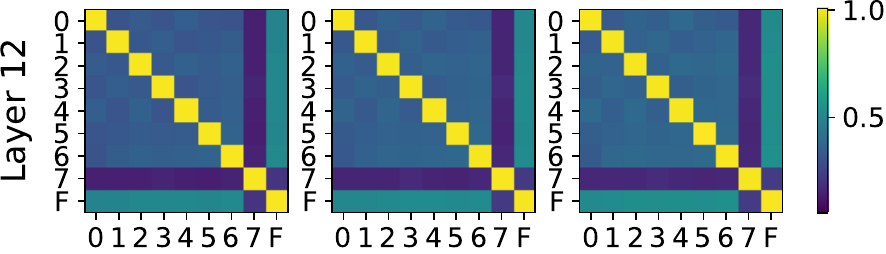} }}$ 
    $\vcenter{\hbox{\includegraphics[width=.14\linewidth]{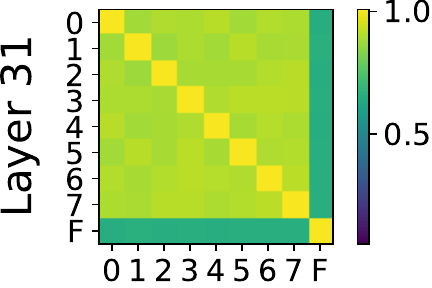} }}$
    $\vcenter{\hbox{\includegraphics[width=.33\linewidth]{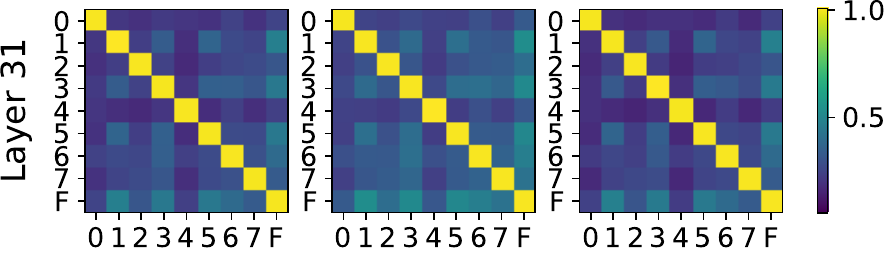} }}$ \\
    $\vcenter{\hbox{\includegraphics[width=.08\linewidth]{mixtral_22.pdf} }}$ \\
    $\vcenter{\hbox{\includegraphics[width=.14\linewidth]{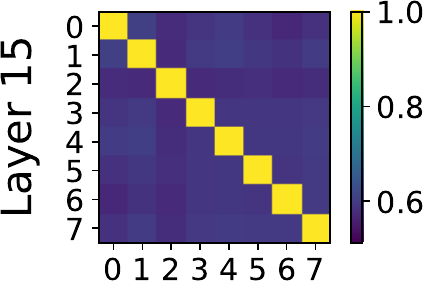} }}$ 
    $\vcenter{\hbox{\includegraphics[width=.33\linewidth]{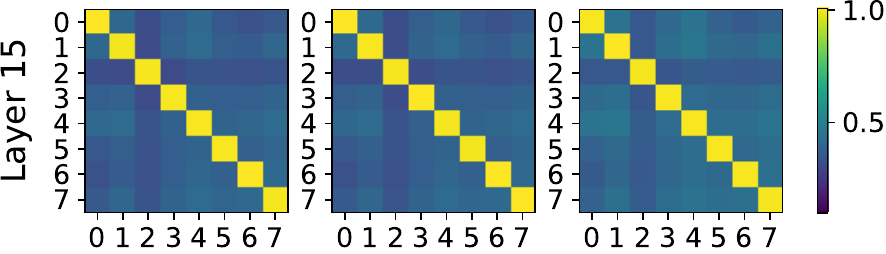} }}$ 
    $\vcenter{\hbox{\includegraphics[width=.14\linewidth]{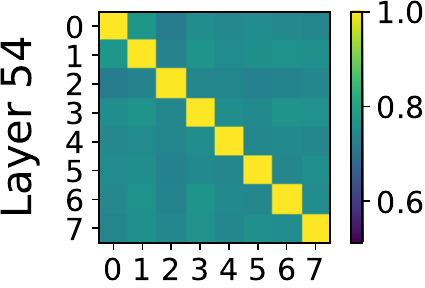} }}$
    $\vcenter{\hbox{\includegraphics[width=.33\linewidth]{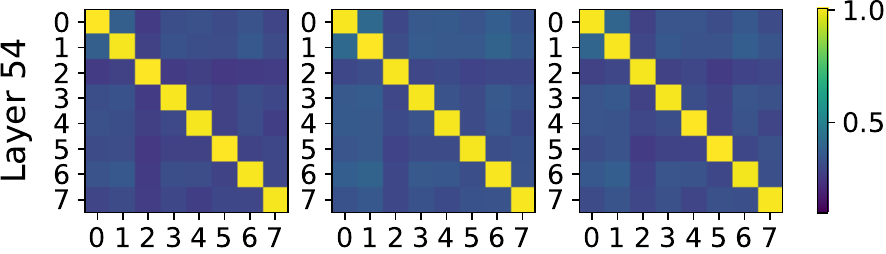} }}$ \\
    % \vspace{1.mm}\rule{\linewidth}{0.5pt} \\
    $\vcenter{\hbox{\includegraphics[width=.08\linewidth]{deepseek.pdf} }}$ \\
    $\vcenter{\hbox{\includegraphics[width=.13\linewidth]{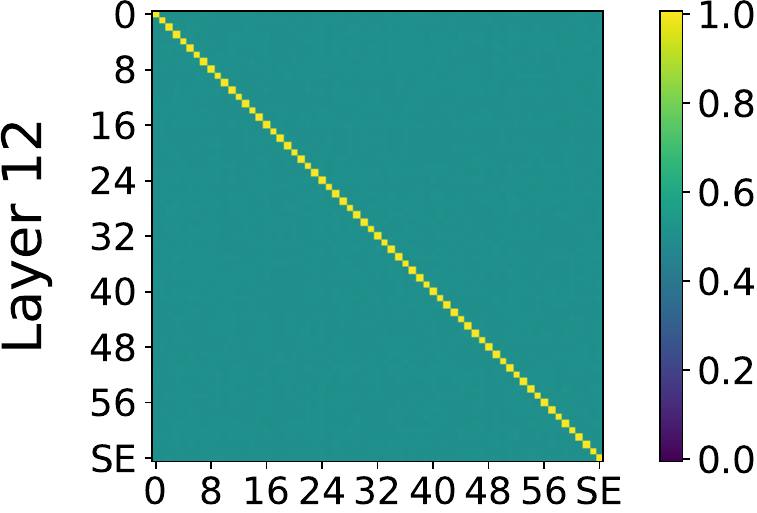} }}$
    $\vcenter{\hbox{\includegraphics[width=.33\linewidth]{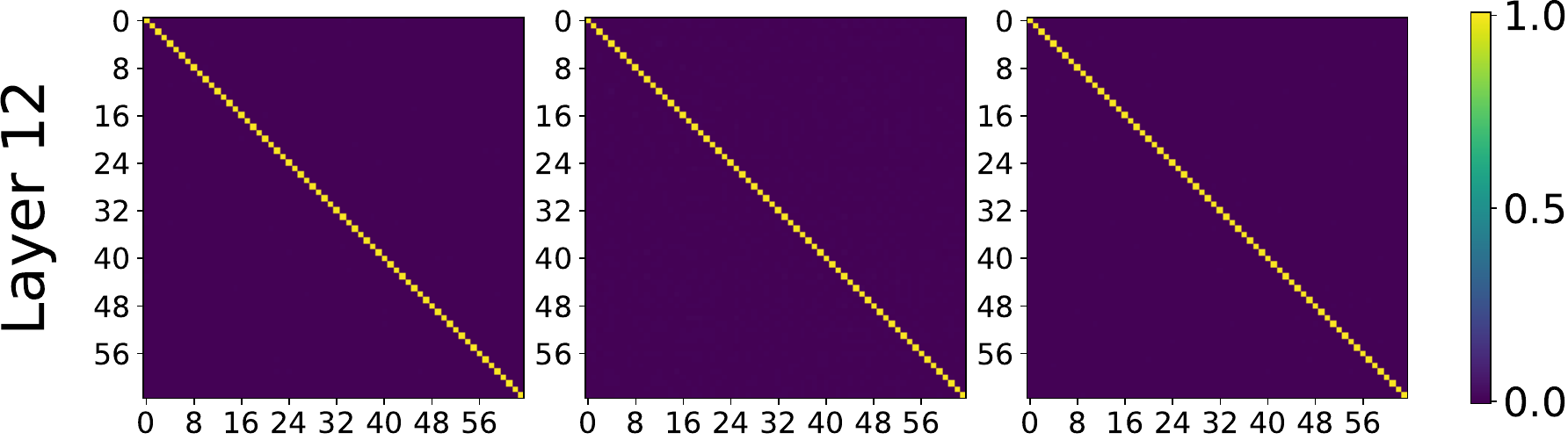} }}$ \hspace{1mm}
    $\vcenter{\hbox{\includegraphics[width=.13\linewidth]{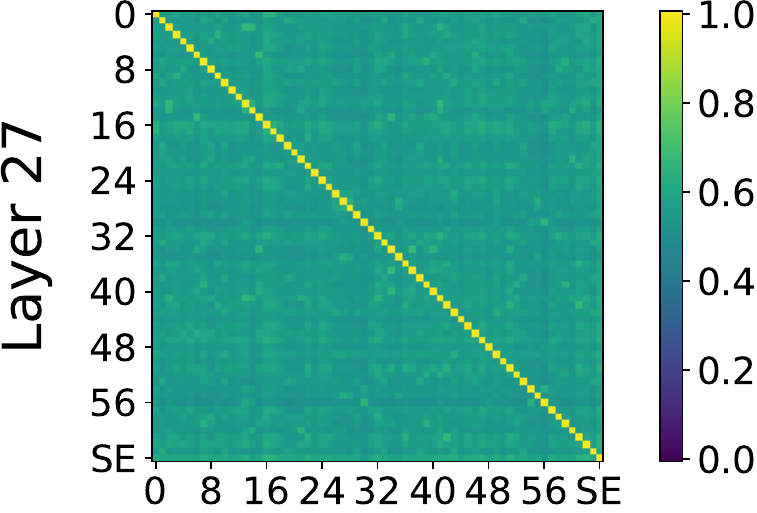} }}$ 
    $\vcenter{\hbox{\includegraphics[width=.33\linewidth]{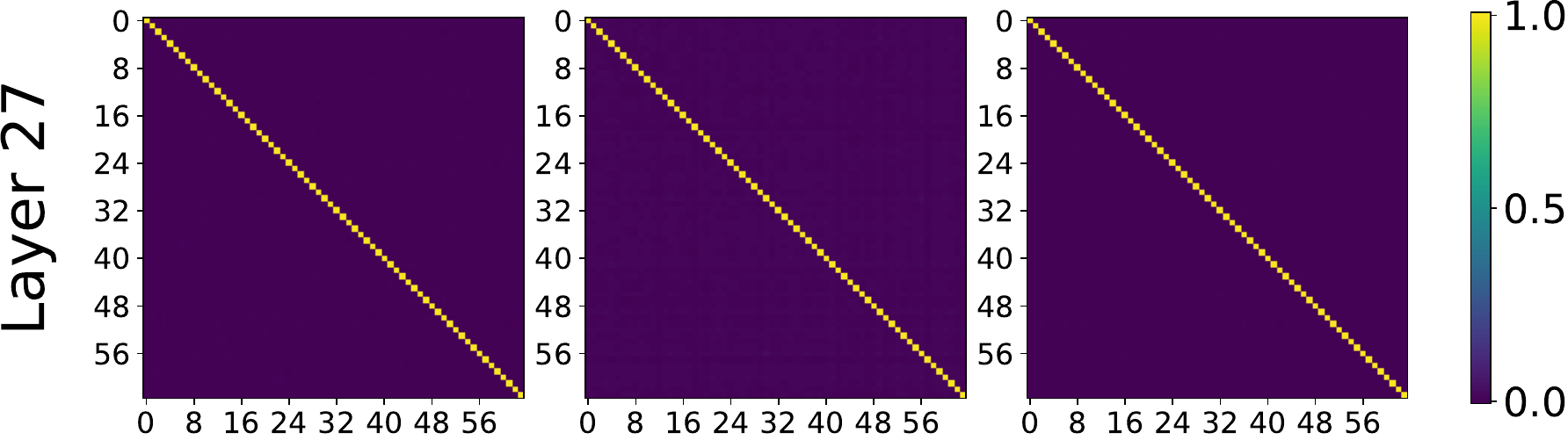} }}$ \\
    % \vspace{1.mm}\rule{\linewidth}{0.5pt} \\
    $\vcenter{\hbox{\includegraphics[width=.08\linewidth]{grok.pdf} }}$ \\
    $\vcenter{\hbox{\includegraphics[width=.14\linewidth]{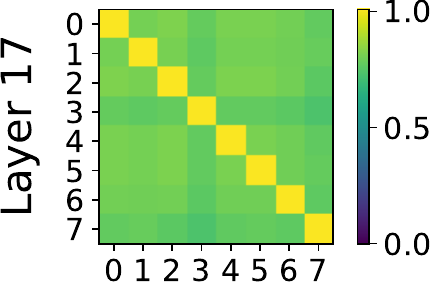} }}$
    $\vcenter{\hbox{\includegraphics[width=.33\linewidth]{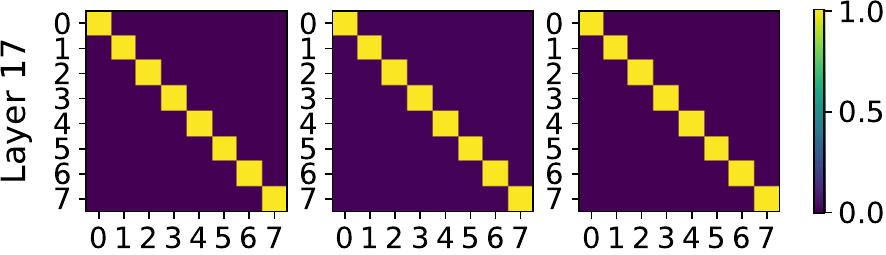} }}$
    $\vcenter{\hbox{\includegraphics[width=.14\linewidth]{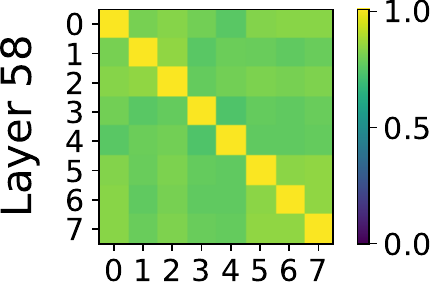} }}$
    $\vcenter{\hbox{\includegraphics[width=.33\linewidth]{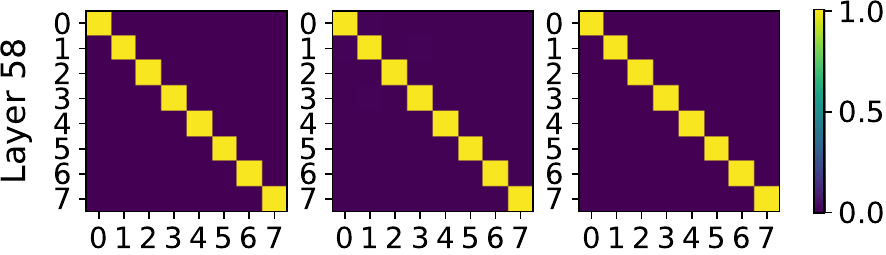} }}$\\
    % \vskip -0.1in 
    \caption{\footnotesize Average similarity heat maps of expert output features using the long input, plotted along with the matrix-level similarity heat maps. The tick numbers refer to the expert indices. ``F'' denotes the Mistral FFN.}
    \label{fig:out_avgsim}
\end{figure*}

The previous experiments examine the MoE models via their parameters, without involving any input.
In this section, we feed text sequences into the MoE models to further study their actual behaviours given various inputs.
Specifically, we analyze the outputs of the experts and gates.

To this end, two stages are required for inference. 
In the first stage, we simply pass the input $x$ through the network using the original Top-k setting and store the output $z_{i}$ of every layer $i$. 
In the second stage, we iterate through the layers. 
During the $i$-th iteration, we feed $z_{i-1}$ into the $i$-th layer (for the first layer, $x$ is employed as the input), set $\text{Top-k}=\operatorname{ALL}$, and record the outputs from all the experts in the $i$-th layer. 
Note that each layer has its own individual forward pass in the second stage. 
Intuitively, our goal is to examine the experts’ behaviors when provided with the original inputs.

\textbf{Input data.}
We utilize a \textit{short input} and a \textit{long input} for the experiments in this section.
For the short input, we employ the first few words of the input from another MoE-related work~\cite{cai2024survey}~\footnote{The specific tokens are \textit{<s>}, \textit{As}, \textit{an}, \textit{open}, \textit{source}, \textit{alternative}, \textit{to}, where the start of the sentence symbol \textit{<s>} does not applicable for the Grok tokenizer.}.
For the long input, we adopt 10 sequences from the test set of the WikiText-103~\cite{merity2016pointer} dataset, totaling approximately 1100 tokens.
The sequences in WikiText-103 cover a variety of domains, with the 10 sequences we used spanning topics such as music, design, and construction.
To ensure the robustness of our findings, we repeat experiments requiring the long input (\S~5.1, \S~5.2) using additional datasets with over 80K tokens, including GSM8K~\cite{cobbe2021gsm8k} and Magicoder-
Evol-Instruct-110K~\cite{wei2024magicoder}.
See Append~\ref{append:dataset} for details.
\textit{The observations of these additional, subject-specific datasets align with the results described in the main context, demonstrating the universality of our conclusions.}

We also conduct experiments for analyzing intermediate states of experts and routing patterns.
Due to the page limit, these experiments are presented in Append~\ref{exp:intermediate} and Append~\ref{exp:chosen_exp}, respectively.

\subsection{Outputs of Experts} 
\label{exp:out_sim}

Since experts are ideally learned to specialize in different aspects, it is natural to question the similarities and differences between the outputs of selected and non-selected experts.
In this experiment, we measure the correlation between the output feature vectors of experts.
We plot the similarity heat maps for three tokens in the short input (Fig.~\ref{fig:out_sim}) and the average heat map across all tokens in the long input (Fig.~\ref{fig:out_avgsim}).
For the long input, we use \textit{angular similarity} instead of cosine similarity for measurement, as the similarities need to be averaged, ensuring that the values range from 0 to 1:
\begin{equation}
    \operatorname{angular\_sim} = 1 - \frac{\arccos{(\operatorname{cosine\_sim}})}{\pi}.
\end{equation}
For clarity, the average similarity heat maps are plotted alongside the matrix-level similarity graphs of the expert weight matrices. Fig.~\ref{fig:out_avgsim_extra} further depicts the results from additional datasets, which are consistent with those of the long input.

\noindent\textbf{Mixtrals and Mistral.}
The graphs for the short input indicate that the outputs from chosen experts tend to be more similar, possibly due to their generally larger norms, which we will discuss in \S~\ref{exp:out_norm}. 
Overall similarities are relatively low in the deeper (22$^{\text{nd}}$-27$^{\text{th}}$ for Mixtral and 30$^{\text{nd}}$-50$^{\text{th}}$ for Mixtral-22) layers, whereas many values exceed 0.8 in the last few layers.
Furthermore, dark crosses often appear in the graphs, with the experts corresponding to these dark crosses often being more similar to the Mistral FFN (\textit{i.e.}, bright color in the last row).
For the long input, the average heat maps show patterns akin to neuron-level similarity graphs, including the presence of dark crosses.
The similarities also decrease with increasing layer depth, except in the last layer.
In addition, we have $S_{\text{ee}}>S_{\text{ef}}$ for both inputs.
Most of these observations align with the previous analyses of static parameters (\S~\ref{sec:summary-static}), implying that measuring the similarity of weights, in some aspects, is equivalent to measuring the average similarity of outputs.

\noindent\textbf{DeepSeek.}
Given the short input, most similarities are around zero, while the values in the last layer are significantly larger.
Again, the similarities between experts chosen by the gate are likely to be higher, although this difference occurs much less frequently than in Mixtrals.
The average similarities for the long input also approach zero.
Moreover, the number of ``small rectangular'' with relatively light color in the graphs decreases as the layer depth increases (except for the last layer), meaning that the average similarities gradually decline. 

\noindent\textbf{Grok.}
Surprisingly, the similarities between the output features remain high for all tokens in the short input, indicating the experts exhibit similar behaviours.
However, the similarities of their weight matrices are mostly zeros (\S~\ref{exp:mat_sim}).
We speculate that this may be due to the relatively large size of each Grok expert, allowing each to learn comprehensive knowledge and behave similarly despite having distinct parameters.
When averaging the similarities for the long input, some of the resulting average heat maps display patterns similar to those of the $W_{\text{act}}$ figures.
This relationship aligns with the observations made for Mixtrals.

\subsection{Norms of Expert Outputs and Gate Scores} 
\label{exp:out_norm}

\begin{figure*}[t!]
    \centering
    $\vcenter{\hbox{\includegraphics[width=.09\linewidth]{mixtral.pdf} }}$
    $\vcenter{\hbox{\includegraphics[width=.89\linewidth]{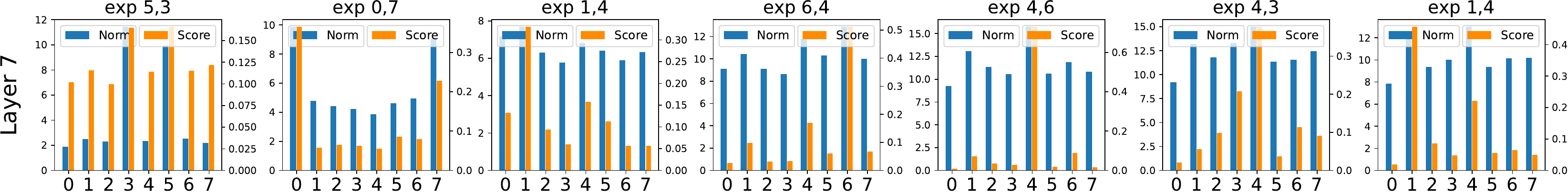} }}$ \\
    % \vspace{1.mm}\rule{\linewidth}{0.5pt}\vspace{1.mm}
    $\vcenter{\hbox{\includegraphics[width=.09\linewidth]{mixtral_22.pdf} }}$
    $\vcenter{\hbox{\includegraphics[width=.89\linewidth]{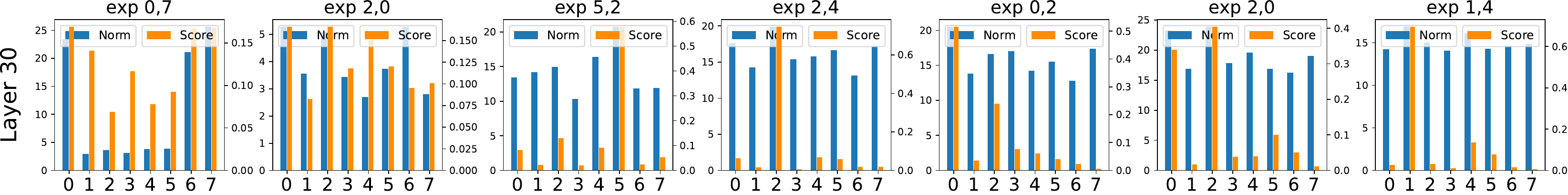} }}$ \\
    $\vcenter{\hbox{\includegraphics[width=.09\linewidth]{deepseek.pdf} }}$
    $\vcenter{\hbox{\includegraphics[width=.89\linewidth]{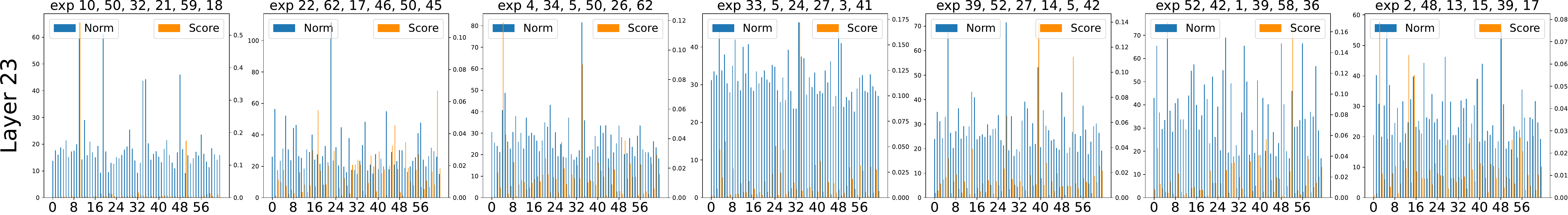} }}$ \\
    % \vspace{1.mm}\rule{\linewidth}{0.5pt}\vspace{1.mm}
    $\vcenter{\hbox{\includegraphics[width=.09\linewidth]{grok.pdf}}}$
    $\vcenter{\hbox{\includegraphics[width=.89\linewidth]{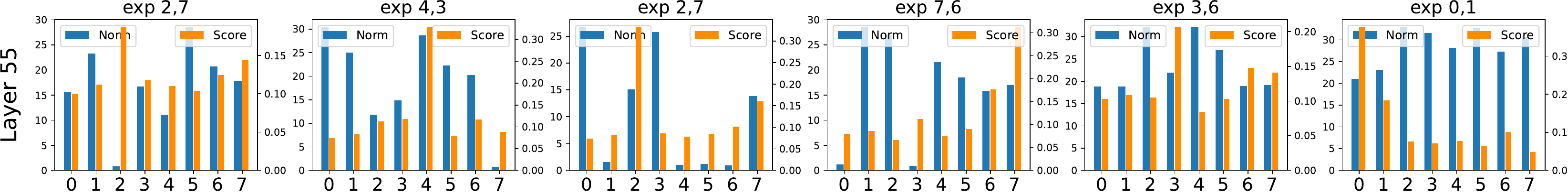} }}$
    \caption{\footnotesize The experts' L2 norms and the gate scores of the short input. Each token's  $k$ experts are shown on top of each heat map. Each number in the horizontal axis refers to an expert index.}
    \label{fig:norm}
\end{figure*}

In \S~\ref{exp:out_sim}, we find that the outputs from chosen experts tend to be more alike.
To investigate the possible reasons for this observation, we employ the short input to study the relationship between the experts' L2 norm and the gate decision in this experiment. 
The calculated norms, along with the gate scores, are plotted in Fig.~\ref{fig:norm}.
In Append~\ref{append:norm}, we repeat this experiment using the long input and additional datasets, and the results also support the ``higher norm, higher score'' observation.

\noindent\textbf{Mixtrals.}
We found that the two experts chosen by the gate usually output feature vectors with the highest norms, which reveals that the norm might be one of the key factors in gate decisions.
This finding agrees with the router's design in CompeteSMoE~\cite{pham2024competesmoe}, which selects experts based on their output norms.
It also helps explain why the outputs of the chosen Mixtrals and DeepSeek experts tend to be more alike (\S~\ref{exp:out_sim}). 
In Fig.~\ref{fig:norm}, we observe that the gate scores assigned to the top-1 experts are usually much higher than those of the others, including the second place.
This demonstrates that the gate is learned to strengthen the confidence of its decision during training.
% While the chosen experts output larger norms, the gate scores are not strictly proportional to the norms for the remaining experts.
% For instance, the expert with the lowest score might not output the smallest norm. 
% Meanwhile, the norm of the feature vector outputted by Mistral FFN is always the lowest, which may correspond to the observation of $S_{ee}>S_{ef}$ in \S~\ref{exp:out_sim}.
On the other hand, the deeper the layer, the larger the norm, which is similar to the growth in standard models~\cite{shleifer2021normformer}.
% This is also similar to our observation in 
% which might be because of the larger gradient during backpropagation.

\noindent\textbf{DeepSeek.}
In contrast to the observation about Mixtrals' experts, the gate decision appears to depend less obviously on the output norms of DeepSeek experts.
However, the top-1 experts often score much higher than the remaining candidates. 
The magnitude of the norms increases with depth, although the increment is less pronounced than in Mixtrals.
In the last layer, the variance of norms becomes greater.
% In addition, the output norms of shared experts are always the lowest, except for the last layer.

\noindent\textbf{Grok.}
While the scores of the top-1 experts are higher than those of the others, no correspondence between the norms and the gate scores is observed.
One possible reason could be the relatively low activation ratios of GeLU (see Append~\ref{exp:intermediate}), which may lead to a weaker dependence on the norm for gate decisions. 
Besides, unlike Mixtrals and DeepSeek, the magnitude of the norms hardly changes across depth, and some of the norm values can be less than 1, which is rare in the other two models.

\begin{figure}[tbp]
    \centering
    \includegraphics[width=.65\linewidth]{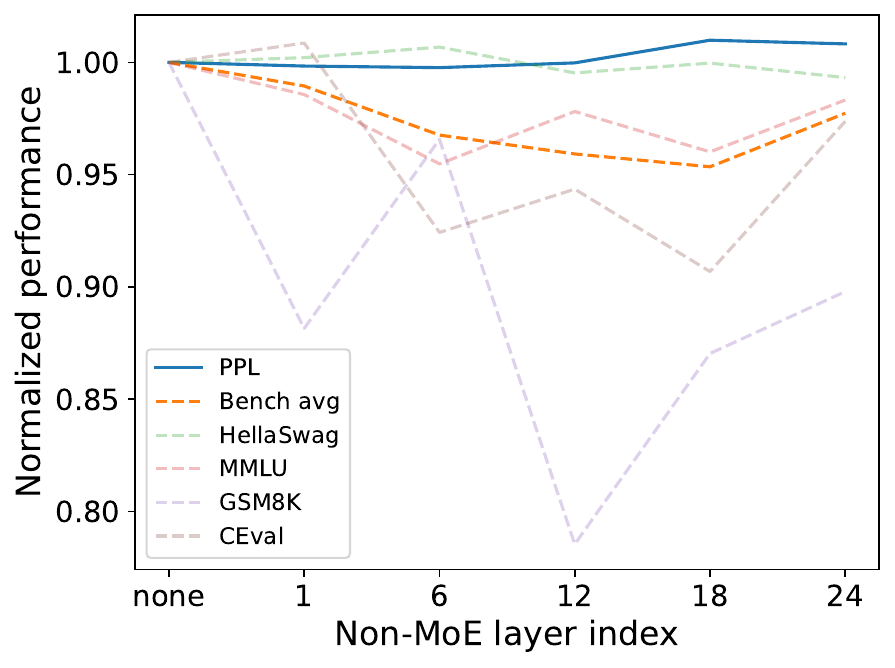}
    \caption{\footnotesize Normalized model performance across benchmarks for the dynamic expert numbers experiment. Solid line represents lower is better, while dashed line represents higher is better. ``Bench avg'' refers to the average performance over the four benchmarks evaluated.}
    \label{fig:dynamic_moe}
\end{figure}

\subsection{Summary}

The observations of dynamic behaviours are concluded below:
\begin{inparaenum}[i)]
    \textbf{\item} The outputs of Mixtrals and DeepSeek experts in deep (last) layers are less (much) alike.
    This can be seen in the heat maps for both the short (Fig.~\ref{fig:out_sim}) and long (Fig.~\ref{fig:out_avgsim}) inputs.  
    \textbf{\item} The average heat maps of expert outputs resemble the neuron-level similarity graphs (Fig.~\ref{fig:out_avgsim}), implying that weight similarity measurements can reflect output similarity.
    \textbf{\item} Grok experts exhibit high output similarity (Fig.\ref{fig:out_sim}), likely due to their larger sizes.
    \textbf{\item} For Mixtrals and DeepSeek, experts generating feature vectors with larger norms tend to receive higher gate scores, as shown in Fig.~\ref{fig:norm}.
    We further verified this observation in Fig.~\ref{fig:reg_norm}. 
    % \textbf{\item} In Mixtral, neurons of different experts located at the same indices have similar activation values.
    % This is evidenced by the presence of vertical lines in the Mixtral heat maps (Fig.~\ref{fig:inter}).
\end{inparaenum}

% \begin{itemize}[leftmargin=5mm, noitemsep]
%     \item The outputs of Mixtral and DeepSeek experts in deep/last layers are less/much alike.
%     This can be observed from the heat maps for both the short (Fig.~\ref{fig:out_sim}) and long (Fig.~\ref{fig:out_avgsim}) inputs.  
%     \item The average heat maps of expert outputs have similar patterns to the neuron-level similarity graphs (Fig.~\ref{fig:out_avgsim}), implying that measuring the similarity of weights, in some aspects, is equivalent to measuring the average similarity of outputs.
%     \item The outputs of Grok experts are highly similar (Fig.~\ref{fig:out_sim}), which may be due to their large sizes.
%     \item For both Mixtral and DeepSeek, experts that output feature vectors with larger norms are likely to achieve higher gate scores as shown in Fig.~\ref{fig:norm}.
%     We further verified this observation in Fig.~\ref{fig:norm} in the appendix. 
%     \item For Mixtral, neurons of different experts positioned at the same indices have similar activation values.
%     In Fig.~\ref{fig:inter}, some vertical lines can be found in the Mixtral heat maps. 
% \end{itemize}

\section{Discussion}
\label{sec:discussion}

Based on our analyses, we offer several suggestions for MoE models across various aspects.

% \begin{itemize}[leftmargin=5mm, noitemsep]
\noindent\textbf{Neuron-level experts.}
Intuitively, the gate embedding matrix $W_g$ determines expert selection while $W_{\text{act}}$ is responsible for choosing which neurons to activate.
Meanwhile, we find that the similarities of $W_g$ and of $W_{\text{act}}$ show association.
This implies that neurons may function as more fine-grained experts. 
Therefore, operations on experts, such as division, construction, and composition, should be further studied at the micro level.
For instance, MoEfication~\cite{zhang2021moefication} and EMoE~\cite{qiu2024unlocking} construct MoE experts by splitting the MLP layers of a dense model, suggesting our findings from a similar perspective.

\noindent\textbf{Model architecture.}
Given that the similarities between experts tend to be relatively low (high) in deep (last) layers, one can consider increasing the number of experts in the deeper layers while reducing it in the last layers.
In addition, since the gate frequently selects experts with larger output norms, employing norm-based routing mechanism is a reasonable approach.
Empirical evidence from \citet{pham2024competesmoe} supports this effectiveness.

We conduct an initial experiment to provide practical experience into our suggestion regarding dynamic expert numbers across layers.
Specifically, we train six MoE models from scratch, each containing 24 layers and 3.6B total parameters, using approximately 120B tokens.
One of the six models is composed of 24 MoE layers, while the others comprise only 23 MoE layers, with one conventional non-MoE layer positioned at different indices.
Details of the model architecture are provided in Tab.\ref{tab:dynamic_moe_architecture}.
As displayed in Fig.\ref{fig:dynamic_moe} and Tab.\ref{tab:dynamic_moe}, the average model performance (\textit{i.e.}, PPL and Bench avg) gradually degrades as the non-MoE layer index increases, whereas a slight improvement appears when the non-MoE layer is placed at the last position (24$^{\text{th}}$).
This highlights the growing importance of multiple expert networks in deeper layers, excluding the last one, which aligns with our observations and suggestions.

\noindent\textbf{Correlation measurement.}
% When analyzing the correlation between experts, measuring the similarities of their weight matrices yields results that are partially equivalent to measuring the similarities of their output feature vectors over considerable tokens. 
% Hence, assessing weight matrices can provide a broader overview, while inspecting the outputs of various tokens individually benefits fine-grained investigation.
Analyzing expert correlations through weight matrix similarities yields partially equivalent results to those from output feature vector similarities across considerable tokens. 
Thus, assessing weight matrices offers a broader overview, while examining individual token outputs allows for more detailed analysis.

\noindent\textbf{Training scheme.}
The training method for Mixtral has not been publicly announced.
However, we observed certain characteristicss shared by Mixtral experts (\textit{e.g.}, relatively high similarities of weight matrices), and a notable relationship between these experts and the Mistral FFN (\textit{e.g.}, similar intermediate states in Fig.~\ref{fig:inter}).
Consequently, we conjecture that the Mixtral model may be trained using special initialization schemes other than from scratch, \textit{e.g.}, upcycling~\cite{komatsuzaki2022sparse} from Mistral, that is, copying all experts from the FFN.  
On the contrary, the experts of DeepSeek and Grok, which are known to be trained from scratch, show weaker correlations than Mixtral experts in our experiments.
Similarly,~\citet{wei2024skywork} tracks changes in expert similarities throughout the training process, observing that upcycled experts exhibit greater similarity compared to those randomly initialized.
Hence, we speculate that training a MoE model from scratch shows stronger potential to facilitate the diversification of experts compared with certain initialization approaches.

% \end{itemize}

\section{Related Work}

Due to the page limit, we focus on existing works analyzing MoEs.
An extended related work section for MoE LLMs can be found in Append~\ref{append:related_work}. 

Most existing works analyze MoE from the router's perspective by observing expert selections.
Early works have observed the unstable choices in the router~\cite{zuo2021taming, chi2022representation, dai2022stablemoe}.
More recent studies find the standard routers do not show clear specialization at the domain level~\cite{jiang2024mixtral, dai2024deepseekmoe} and primarily route based on token ID instead of high-level semantics~\cite{xue2024openmoe}.
\citet{shi2024unchosen} shows that Top-2 and Rank-k routing result in different model behaviours and proposes a new self-contrast decoding method to determine the next-token distribution based on this finding.

Other works investigate the expert's similarity~\cite{wu2022residual}, uncovering and utilizing redundancies among experts for efficient inference~\cite{li2023merge, lu2024not}.
\citet{zhang2024diversifying} reveals the redundancy within experts and perform pruning based on their similarities.
\citet{liu2023towards,qiu2023emergent} notice the connection between routing connection and expert computation, and utilize the average of the experts' first-layer weights to guide routing.
\citet{pham2024competesmoe} proposes adding the expert's output norm as a supervision signal for routing training.
~\citet{chen2022towards} empirically and theoretically proves that a two-layer MoE CNN is able to learn cluster-center features via specializing experts to specific portions of the data.
While these works provide insights into MoE from one or two viewpoints, our work offers a systematic analysis and comparison focusing on transformer-based MoE LLMs.

\textit{As mentioned in previous sections, several existing works share some relevance to our findings, and thus can be seen as supportive.
However, their proposed ideas and methods are different from ours.} 
For instance, rather than revealing the nature of the preference for large output norms in (conventional top-k) routing, as we analyze, CompeteSMoE~\cite{pham2024competesmoe} designs a norm-based router to introduce this tendency manually; 
MoEfication~\cite{zhang2021moefication} splits MLP layers of a dense model to construct MoE experts, while our study highlights that the neurons of an expert can be seen as tiny experts.
Moreover, many of our observations are novel, such as the correlation between the router embedding matrix and the expert weight matrix, as well as the equivalence between parameter and output measurement for experts. 
Therefore, we believe that our work offers valuable insights into MoE LLMs for the community.

\section{Conclusion}

In this paper, we initially attempt to investigate the inner working mechanisms of MoEs by studying the parameters and outputs of four different MoE models.
We summarize our empirical observations and propose practical suggestions across various aspects.
While it is premature to conclude whether MoEs genuinely learn heterogeneous experts, some of our experiments indicate that specific architectural designs (\textit{e.g.}, the number of experts) and training frameworks may facilitate expert specialization.
We hope this work can provide inspiring insights and serve as a valuable foundation for future research on MoE and other modular architectures.

\section{Limitations}

The limitations of our work include:
\begin{inparaenum}[1)]
    \item Although the models we investigated cover several common designs of MoE, our analysis does not encompass all aspects (e.g., other routing strategies like top-1 routing or model architectures that place MoE layers at every other layer);
    \item Despite the availability of other metrics, we primarily adopt cosine similarity in our experiments involving similarity measurement, as it is a widely used approach~\cite{pham2024competesmoe, chen2022towards};
    \item We mainly focus on the pretrained base model but seldom explore the behaviours of models after fine-tuning. Analyzing the changes in expert behaviours during the fine-tuning process could yield valuable insights.
\end{inparaenum}

\bibliography{main}

\begin{thebibliography}{41}
\expandafter\ifx\csname natexlab\endcsname\relax\def\natexlab#1{#1}\fi

\bibitem[{Cai et~al.(2024)Cai, Jiang, Wang, Tang, Kim, and Huang}]{cai2024survey}
Weilin Cai, Juyong Jiang, Fan Wang, Jing Tang, Sunghun Kim, and Jiayi Huang. 2024.
\newblock A survey on mixture of experts.
\newblock \emph{arXiv preprint arXiv:2407.06204}.

\bibitem[{Chen et~al.(2022)Chen, Deng, Wu, Gu, and Li}]{chen2022towards}
Zixiang Chen, Yihe Deng, Yue Wu, Quanquan Gu, and Yuanzhi Li. 2022.
\newblock Towards understanding the mixture-of-experts layer in deep learning.
\newblock \emph{Advances in neural information processing systems}, 35:23049--23062.

\bibitem[{Chi et~al.(2022)Chi, Dong, Huang, Dai, Ma, Patra, Singhal, Bajaj, Song, Mao et~al.}]{chi2022representation}
Zewen Chi, Li~Dong, Shaohan Huang, Damai Dai, Shuming Ma, Barun Patra, Saksham Singhal, Payal Bajaj, Xia Song, Xian-Ling Mao, et~al. 2022.
\newblock On the representation collapse of sparse mixture of experts.
\newblock \emph{Advances in Neural Information Processing Systems}, 35:34600--34613.

\bibitem[{Cobbe et~al.(2021)Cobbe, Kosaraju, Bavarian, Chen, Jun, Kaiser, Plappert, Tworek, Hilton, Nakano, Hesse, and Schulman}]{cobbe2021gsm8k}
Karl Cobbe, Vineet Kosaraju, Mohammad Bavarian, Mark Chen, Heewoo Jun, Lukasz Kaiser, Matthias Plappert, Jerry Tworek, Jacob Hilton, Reiichiro Nakano, Christopher Hesse, and John Schulman. 2021.
\newblock Training verifiers to solve math word problems.
\newblock \emph{arXiv preprint arXiv:2110.14168}.

\bibitem[{Dai et~al.(2024)Dai, Deng, Zhao, Xu, Gao, Chen, Li, Zeng, Yu, Wu et~al.}]{dai2024deepseekmoe}
Damai Dai, Chengqi Deng, Chenggang Zhao, RX~Xu, Huazuo Gao, Deli Chen, Jiashi Li, Wangding Zeng, Xingkai Yu, Y~Wu, et~al. 2024.
\newblock Deepseekmoe: Towards ultimate expert specialization in mixture-of-experts language models.
\newblock \emph{arXiv preprint arXiv:2401.06066}.

\bibitem[{Dai et~al.(2022)Dai, Dong, Ma, Zheng, Sui, Chang, and Wei}]{dai2022stablemoe}
Damai Dai, Li~Dong, Shuming Ma, Bo~Zheng, Zhifang Sui, Baobao Chang, and Furu Wei. 2022.
\newblock Stablemoe: Stable routing strategy for mixture of experts.
\newblock \emph{arXiv preprint arXiv:2204.08396}.

\bibitem[{Geva et~al.(2020)Geva, Schuster, Berant, and Levy}]{geva2020transformer}
Mor Geva, Roei Schuster, Jonathan Berant, and Omer Levy. 2020.
\newblock Transformer feed-forward layers are key-value memories.
\newblock \emph{arXiv preprint arXiv:2012.14913}.

\bibitem[{Jiang et~al.(2023)Jiang, Sablayrolles, Mensch, Bamford, Chaplot, Casas, Bressand, Lengyel, Lample, Saulnier et~al.}]{jiang2023mistral}
Albert~Q Jiang, Alexandre Sablayrolles, Arthur Mensch, Chris Bamford, Devendra~Singh Chaplot, Diego de~las Casas, Florian Bressand, Gianna Lengyel, Guillaume Lample, Lucile Saulnier, et~al. 2023.
\newblock Mistral 7b.
\newblock \emph{arXiv preprint arXiv:2310.06825}.

\bibitem[{Jiang et~al.(2024)Jiang, Sablayrolles, Roux, Mensch, Savary, Bamford, Chaplot, Casas, Hanna, Bressand et~al.}]{jiang2024mixtral}
Albert~Q Jiang, Alexandre Sablayrolles, Antoine Roux, Arthur Mensch, Blanche Savary, Chris Bamford, Devendra~Singh Chaplot, Diego de~las Casas, Emma~Bou Hanna, Florian Bressand, et~al. 2024.
\newblock Mixtral of experts.
\newblock \emph{arXiv preprint arXiv:2401.04088}.

\bibitem[{Jonker and Volgenant(1988)}]{jonker1988shortest}
Roy Jonker and Ton Volgenant. 1988.
\newblock A shortest augmenting path algorithm for dense and sparse linear assignment problems.
\newblock In \emph{DGOR/NSOR: Papers of the 16th Annual Meeting of DGOR in Cooperation with NSOR/Vortr{\"a}ge der 16. Jahrestagung der DGOR zusammen mit der NSOR}, pages 622--622. Springer.

\bibitem[{Komatsuzaki et~al.(2022)Komatsuzaki, Puigcerver, Lee-Thorp, Ruiz, Mustafa, Ainslie, Tay, Dehghani, and Houlsby}]{komatsuzaki2022sparse}
Aran Komatsuzaki, Joan Puigcerver, James Lee-Thorp, Carlos~Riquelme Ruiz, Basil Mustafa, Joshua Ainslie, Yi~Tay, Mostafa Dehghani, and Neil Houlsby. 2022.
\newblock Sparse upcycling: Training mixture-of-experts from dense checkpoints.
\newblock \emph{arXiv preprint arXiv:2212.05055}.

\bibitem[{Li et~al.(2023)Li, Zhang, Yadav, Sung, Cheng, Bansal, and Chen}]{li2023merge}
Pingzhi Li, Zhenyu Zhang, Prateek Yadav, Yi-Lin Sung, Yu~Cheng, Mohit Bansal, and Tianlong Chen. 2023.
\newblock Merge, then compress: Demystify efficient smoe with hints from its routing policy.
\newblock \emph{arXiv preprint arXiv:2310.01334}.

\bibitem[{Li et~al.(2022)Li, You, Bhojanapalli, Li, Rawat, Reddi, Ye, Chern, Yu, Guo et~al.}]{li2022lazy}
Zonglin Li, Chong You, Srinadh Bhojanapalli, Daliang Li, Ankit~Singh Rawat, Sashank~J Reddi, Ke~Ye, Felix Chern, Felix Yu, Ruiqi Guo, et~al. 2022.
\newblock The lazy neuron phenomenon: On emergence of activation sparsity in transformers.
\newblock \emph{arXiv preprint arXiv:2210.06313}.

\bibitem[{Liu et~al.(2023)Liu, Dettmers, Lin, Stoyanov, and Li}]{liu2023towards}
Zeyu~Leo Liu, Tim Dettmers, Xi~Victoria Lin, Veselin Stoyanov, and Xian Li. 2023.
\newblock Towards a unified view of sparse feed-forward network in pretraining large language model.
\newblock \emph{arXiv preprint arXiv:2305.13999}.

\bibitem[{Lu et~al.(2024)Lu, Liu, Xu, Zhou, Huang, Zhang, Yan, and Li}]{lu2024not}
Xudong Lu, Qi~Liu, Yuhui Xu, Aojun Zhou, Siyuan Huang, Bo~Zhang, Junchi Yan, and Hongsheng Li. 2024.
\newblock Not all experts are equal: Efficient expert pruning and skipping for mixture-of-experts large language models.
\newblock \emph{arXiv preprint arXiv:2402.14800}.

\bibitem[{Merity et~al.(2016)Merity, Xiong, Bradbury, and Socher}]{merity2016pointer}
Stephen Merity, Caiming Xiong, James Bradbury, and Richard Socher. 2016.
\newblock Pointer sentinel mixture models.
\newblock \emph{arXiv preprint arXiv:1609.07843}.

\bibitem[{Pham et~al.(2024)Pham, Do, Nguyen, Nguyen, Liu, Sartipi, Nguyen, Ramasamy, Li, Hoi et~al.}]{pham2024competesmoe}
Quang Pham, Giang Do, Huy Nguyen, TrungTin Nguyen, Chenghao Liu, Mina Sartipi, Binh~T Nguyen, Savitha Ramasamy, Xiaoli Li, Steven Hoi, et~al. 2024.
\newblock Competesmoe--effective training of sparse mixture of experts via competition.
\newblock \emph{arXiv preprint arXiv:2402.02526}.

\bibitem[{Qiu et~al.(2023)Qiu, Huang, and Fu}]{qiu2023emergent}
Zihan Qiu, Zeyu Huang, and Jie Fu. 2023.
\newblock Emergent mixture-of-experts: Can dense pre-trained transformers benefit from emergent modular structures?
\newblock \emph{arXiv preprint arXiv:2310.10908}.

\bibitem[{Qiu et~al.(2024{\natexlab{a}})Qiu, Huang, and Fu}]{qiu2024unlocking}
Zihan Qiu, Zeyu Huang, and Jie Fu. 2024{\natexlab{a}}.
\newblock Unlocking emergent modularity in large language models.
\newblock In \emph{Proceedings of the 2024 Conference of the North American Chapter of the Association for Computational Linguistics: Human Language Technologies (Volume 1: Long Papers)}, pages 2638--2660.

\bibitem[{Qiu et~al.(2024{\natexlab{b}})Qiu, Huang, Huang, and Fu}]{qiu2024empirical}
Zihan Qiu, Zeyu Huang, Youcheng Huang, and Jie Fu. 2024{\natexlab{b}}.
\newblock Empirical study on updating key-value memories in transformer feed-forward layers.
\newblock \emph{arXiv preprint arXiv:2402.12233}.

\bibitem[{Reid et~al.(2024)Reid, Savinov, Teplyashin, Lepikhin, Lillicrap, Alayrac, Soricut, Lazaridou, Firat, Schrittwieser et~al.}]{reid2024gemini}
Machel Reid, Nikolay Savinov, Denis Teplyashin, Dmitry Lepikhin, Timothy Lillicrap, Jean-baptiste Alayrac, Radu Soricut, Angeliki Lazaridou, Orhan Firat, Julian Schrittwieser, et~al. 2024.
\newblock Gemini 1.5: Unlocking multimodal understanding across millions of tokens of context.
\newblock \emph{arXiv preprint arXiv:2403.05530}.

\bibitem[{Shazeer et~al.(2017)Shazeer, Mirhoseini, Maziarz, Davis, Le, Hinton, and Dean}]{shazeer2017outrageously}
Noam Shazeer, Azalia Mirhoseini, Krzysztof Maziarz, Andy Davis, Quoc Le, Geoffrey Hinton, and Jeff Dean. 2017.
\newblock Outrageously large neural networks: The sparsely-gated mixture-of-experts layer.
\newblock \emph{arXiv preprint arXiv:1701.06538}.

\bibitem[{Shen et~al.(2024)Shen, Guo, Cai, and Qin}]{shen2024jetmoe}
Yikang Shen, Zhen Guo, Tianle Cai, and Zengyi Qin. 2024.
\newblock Jetmoe: Reaching llama2 performance with 0.1 m dollars.
\newblock \emph{arXiv preprint arXiv:2404.07413}.

\bibitem[{Shen et~al.(2023)Shen, Zhang, Cao, Tan, Chen, and Gan}]{shen2023moduleformer}
Yikang Shen, Zheyu Zhang, Tianyou Cao, Shawn Tan, Zhenfang Chen, and Chuang Gan. 2023.
\newblock Moduleformer: Learning modular large language models from uncurated data.
\newblock \emph{arXiv preprint arXiv:2306.04640}.

\bibitem[{Shi et~al.(2024)Shi, Yang, Zhu, Wang, Wu, Li, Cai, Yang, and Meng}]{shi2024unchosen}
Chufan Shi, Cheng Yang, Xinyu Zhu, Jiahao Wang, Taiqiang Wu, Siheng Li, Deng Cai, Yujiu Yang, and Yu~Meng. 2024.
\newblock Unchosen experts can contribute too: Unleashing moe models' power by self-contrast.
\newblock \emph{arXiv preprint arXiv:2405.14507}.

\bibitem[{Shleifer et~al.(2021)Shleifer, Weston, and Ott}]{shleifer2021normformer}
Sam Shleifer, Jason Weston, and Myle Ott. 2021.
\newblock Normformer: Improved transformer pretraining with extra normalization.
\newblock \emph{arXiv preprint arXiv:2110.09456}.

\bibitem[{Song et~al.(2024{\natexlab{a}})Song, Han, Zhang, Hu, Shi, Li, Chen, Liu, Li, Yang et~al.}]{song2024prosparse}
Chenyang Song, Xu~Han, Zhengyan Zhang, Shengding Hu, Xiyu Shi, Kuai Li, Chen Chen, Zhiyuan Liu, Guangli Li, Tao Yang, et~al. 2024{\natexlab{a}}.
\newblock Prosparse: Introducing and enhancing intrinsic activation sparsity within large language models.
\newblock \emph{arXiv preprint arXiv:2402.13516}.

\bibitem[{Song et~al.(2024{\natexlab{b}})Song, Xie, Zhang, Wen, Ma, Mi, and Chen}]{song2024turbo}
Yixin Song, Haotong Xie, Zhengyan Zhang, Bo~Wen, Li~Ma, Zeyu Mi, and Haibo Chen. 2024{\natexlab{b}}.
\newblock Turbo sparse: Achieving llm sota performance with minimal activated parameters.
\newblock \emph{arXiv preprint arXiv:2406.05955}.

\bibitem[{Sun et~al.(2024)Sun, Pickett, Nain, and Jones}]{sun2024transformer}
Qi~Sun, Marc Pickett, Aakash~Kumar Nain, and Llion Jones. 2024.
\newblock Transformer layers as painters.
\newblock \emph{arXiv preprint arXiv:2407.09298}.

\bibitem[{Team(2024)}]{qwen_moe}
Qwen Team. 2024.
\newblock \href {https://qwenlm.github.io/blog/qwen-moe/} {Qwen1.5-moe: Matching 7b model performance with 1/3 activated parameters"}.

\bibitem[{Touvron et~al.(2023)Touvron, Lavril, Izacard, Martinet, Lachaux, Lacroix, Rozi{\`e}re, Goyal, Hambro, Azhar et~al.}]{touvron2023llama}
Hugo Touvron, Thibaut Lavril, Gautier Izacard, Xavier Martinet, Marie-Anne Lachaux, Timoth{\'e}e Lacroix, Baptiste Rozi{\`e}re, Naman Goyal, Eric Hambro, Faisal Azhar, et~al. 2023.
\newblock Llama: Open and efficient foundation language models.
\newblock \emph{arXiv preprint arXiv:2302.13971}.

\bibitem[{Wei et~al.(2024{\natexlab{a}})Wei, Zhu, Zhao, Cheng, Li, L{\"u}, Cheng, Zhang, Zhang, Zeng et~al.}]{wei2024skywork}
Tianwen Wei, Bo~Zhu, Liang Zhao, Cheng Cheng, Biye Li, Weiwei L{\"u}, Peng Cheng, Jianhao Zhang, Xiaoyu Zhang, Liang Zeng, et~al. 2024{\natexlab{a}}.
\newblock Skywork-moe: A deep dive into training techniques for mixture-of-experts language models.
\newblock \emph{arXiv preprint arXiv:2406.06563}.

\bibitem[{Wei et~al.(2024{\natexlab{b}})Wei, Wang, Liu, Ding, and Zhang}]{wei2024magicoder}
Yuxiang Wei, Zhe Wang, Jiawei Liu, Yifeng Ding, and Lingming Zhang. 2024{\natexlab{b}}.
\newblock Magicoder: Empowering code generation with oss-instruct.
\newblock In \emph{Forty-first International Conference on Machine Learning}.

\bibitem[{Wu et~al.(2022)Wu, Liu, Chen, Chen, Dai, and Yuan}]{wu2022residual}
Lemeng Wu, Mengchen Liu, Yinpeng Chen, Dongdong Chen, Xiyang Dai, and Lu~Yuan. 2022.
\newblock Residual mixture of experts.
\newblock \emph{arXiv preprint arXiv:2204.09636}.

\bibitem[{Wu et~al.(2024)Wu, Luo, Chen, Li, Zhao, Yu, Wang, Wang, Wang, Qiao et~al.}]{wu2024yuan}
Shaohua Wu, Jiangang Luo, Xi~Chen, Lingjun Li, Xudong Zhao, Tong Yu, Chao Wang, Yue Wang, Fei Wang, Weixu Qiao, et~al. 2024.
\newblock Yuan 2.0-m32: Mixture of experts with attention router.
\newblock \emph{arXiv preprint arXiv:2405.17976}.

\bibitem[{Xue et~al.(2024)Xue, Zheng, Fu, Ni, Zheng, Zhou, and You}]{xue2024openmoe}
Fuzhao Xue, Zian Zheng, Yao Fu, Jinjie Ni, Zangwei Zheng, Wangchunshu Zhou, and Yang You. 2024.
\newblock Openmoe: An early effort on open mixture-of-experts language models.
\newblock \emph{arXiv preprint arXiv:2402.01739}.

\bibitem[{Zhang et~al.(2022)Zhang, Shen, Huang, Zhou, Rong, and Xiong}]{zhang2022mixture}
Xiaofeng Zhang, Yikang Shen, Zeyu Huang, Jie Zhou, Wenge Rong, and Zhang Xiong. 2022.
\newblock Mixture of attention heads: Selecting attention heads per token.
\newblock \emph{arXiv preprint arXiv:2210.05144}.

\bibitem[{Zhang et~al.(2024)Zhang, Liu, Cheng, Xu, and Gao}]{zhang2024diversifying}
Zeliang Zhang, Xiaodong Liu, Hao Cheng, Chenliang Xu, and Jianfeng Gao. 2024.
\newblock Diversifying the expert knowledge for task-agnostic pruning in sparse mixture-of-experts.
\newblock \emph{arXiv preprint arXiv:2407.09590}.

\bibitem[{Zhang et~al.(2021)Zhang, Lin, Liu, Li, Sun, and Zhou}]{zhang2021moefication}
Zhengyan Zhang, Yankai Lin, Zhiyuan Liu, Peng Li, Maosong Sun, and Jie Zhou. 2021.
\newblock Moefication: Transformer feed-forward layers are mixtures of experts.
\newblock \emph{arXiv preprint arXiv:2110.01786}.

\bibitem[{Zoph et~al.(2022)Zoph, Bello, Kumar, Du, Huang, Dean, Shazeer, and Fedus}]{zoph2022st}
Barret Zoph, Irwan Bello, Sameer Kumar, Nan Du, Yanping Huang, Jeff Dean, Noam Shazeer, and William Fedus. 2022.
\newblock St-moe: Designing stable and transferable sparse expert models.
\newblock \emph{arXiv preprint arXiv:2202.08906}.

\bibitem[{Zuo et~al.(2021)Zuo, Liu, Jiao, Kim, Hassan, Zhang, Zhao, and Gao}]{zuo2021taming}
Simiao Zuo, Xiaodong Liu, Jian Jiao, Young~Jin Kim, Hany Hassan, Ruofei Zhang, Tuo Zhao, and Jianfeng Gao. 2021.
\newblock Taming sparsely activated transformer with stochastic experts.
\newblock \emph{arXiv preprint arXiv:2110.04260}.

\end{thebibliography}

\clearpage
\appendix

\section*{Appendix}

\section{Model Selection}
\label{append:models}

Our experiments are conducted on Mixtral 8x7B, DeepSeekMoE, and Grok-1.
We choose these models due to their widespread use and impressive performance across various domains.
Additionally, these models are complementary in several crucial attributes, such as training scheme, activation functions, top-k settings, and the number of experts, as listed in Tab~\ref{tab:models} and Tab~\ref{tab:models2}.
This allows for a comparative analysis with controlled variables and encompasses a wide range of parameter sizes, from rather small (16B) to relatively huge (314B).
Hence, we believe that the findings derived from these four models are fairly robust, despite the limited number of models examined.

\begin{table*}[thbp]
    \small
    \centering
    % \vskip -0.3in
    \begin{tabular}{lccccc}
        \hline
        \textbf{Model} & \textbf{Training scheme} & \textbf{Activation} & \textbf{\# Total layers} & \textbf{\# Total params} & \textbf{\# Activated params} \\
        \hline
        Mixtral & unknown (upcycling) & SiLU & 32 & 46.7B & 12.9B \\
        Mixtral-22 & unknown (upcycling) & SiLU & 56 & 141B & 39B \\
        Mistral & from scratch & SiLU & 32 & 7.3B & 7.3B \\
        DeepSeek & from scratch & SiLU & 28 & 16.4B & 0.3B \\
        Grok & from scratch & GeLU & 64 & 314B & 78.5B \\
        \hline
    \end{tabular}
    % \vskip -0.1in
    \caption{\footnotesize Additional information of chosen models.}
    \label{tab:models2}
    % \vskip -0.2in
\end{table*}

\section{Extended Related Work}
\label{append:related_work}
\noindent\textbf{MoE LLMs.}
MoEs have garnered significant attention in recent years due to their ability to efficiently scale model capacity with minimal computational overhead.
Most current transformer-based MoE LLMs adopt a typical architecture design that replaces the original FFN with multiple expert networks and a sparse gating network~\cite{wei2024skywork, wu2024yuan, dai2024deepseekmoe, xue2024openmoe, jiang2024mixtral, zoph2022st}. 
JetMoE~\cite{shen2024jetmoe} and ModuleFormer~\cite{shen2023moduleformer} incorporate Mixture of Attention Heads~\cite{zhang2022mixture} into their model, achieving further sparsity.
A recent survey~\cite{cai2024survey} provides a comprehensive review of both the algorithmic and system design aspects of MoEs.
For this study, we select four representative candidates among current open-sourced MoE LLMs for analysis to gain intriguing insights.

\section{Projection of Expert Matrices in Low-dimensional Space} 

\subsection{Matrix-level} \label{append:mat_pca}

\begin{figure*}[htbp]
    \centering
    $\vcenter{\hbox{\includegraphics[width=.1\linewidth]{mixtral.pdf} }}$
    $\vcenter{\hbox{\includegraphics[width=.7\linewidth]{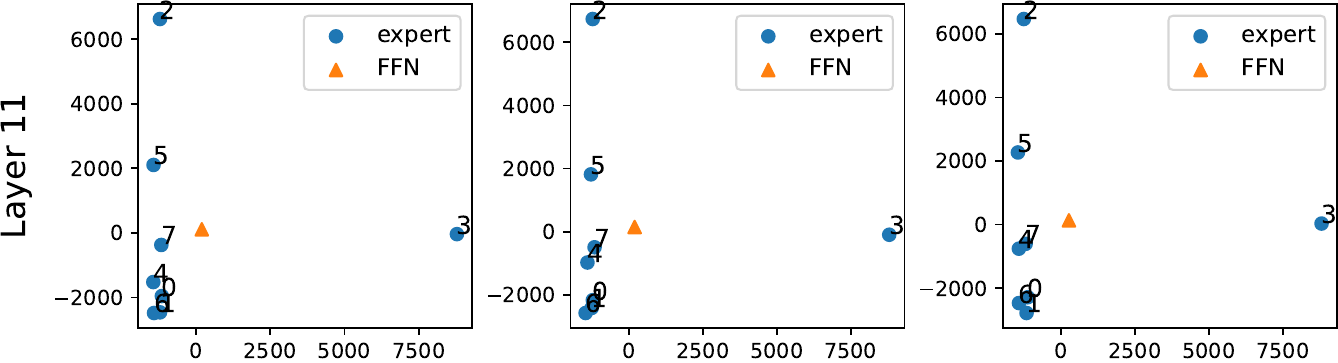} }}$ \\
    \vspace{1.mm}\rule{.9\linewidth}{0.5pt}\vspace{1.mm} \\
    $\vcenter{\hbox{\includegraphics[width=.1\linewidth]{deepseek.pdf} }}$
    $\vcenter{\hbox{\includegraphics[width=.7\linewidth]{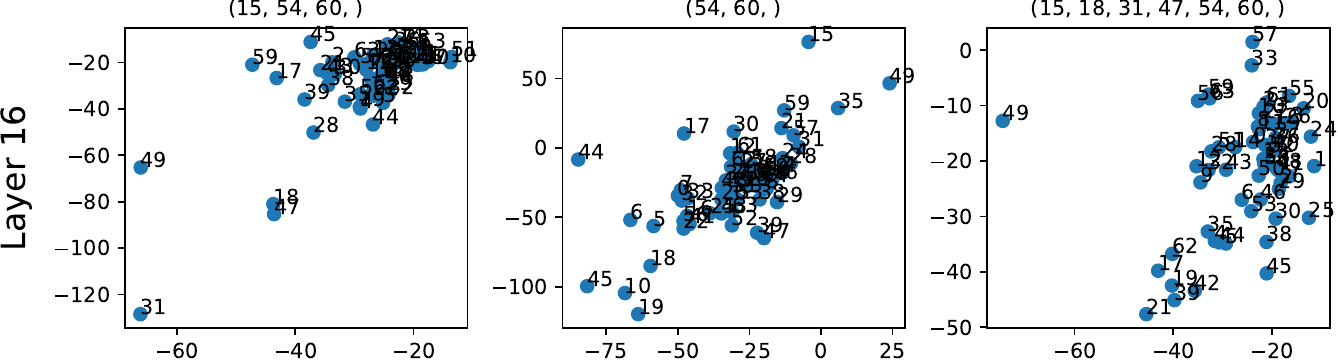} }}$ \\
    \vspace{1.mm}\rule{.9\linewidth}{0.5pt}\vspace{1.mm} \\
    $\vcenter{\hbox{\includegraphics[width=.1\linewidth]{grok.pdf} }}$
    $\vcenter{\hbox{\includegraphics[width=.7\linewidth]{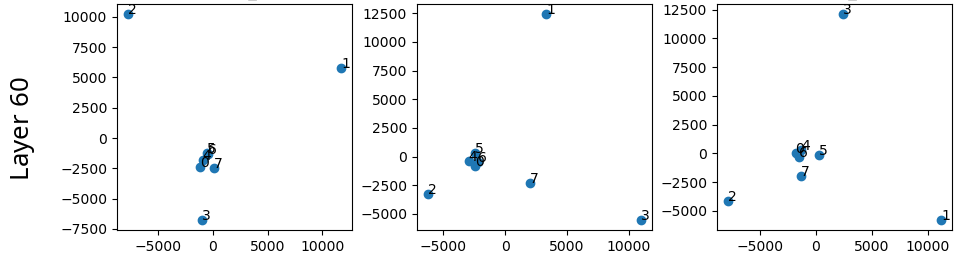} }}$ 
    \caption{\footnotesize Projection of expert matrices in 2D space. Each layer contains three graphs, corresponding to $W_{\text{up}}$, $W_{\text{act}}$, and $W_{\text{down}}$, respectively. For DeepSeek, the indices of the removed outliers are listed on top of each graph.}
    \label{fig:mat_pca}
\end{figure*}

To better understand the relationships among experts, we employ principal components analysis (PCA) to project the flattened vectors of weight matrices into two-dimensional space. 
The vectors are standardized before applying PCA.
Fig.~\ref{fig:mat_pca} depicts the resulting 2D projection.

\noindent\textbf{Mixtral and Mistral.}
Consistent with the observations in \S~\ref{exp:mat_sim}, the figures for the three matrices appear similar.
Generally, about half of the Mixtral experts cluster closely together and near the Mistral FFN, while the others locate much farther away. 
Moreover, the outliers correspond to the dark crosses. 

\noindent\textbf{DeepSeek.}
Only routed experts are considered due to differences in hidden sizes.
Because several outliers exist, causing the remaining data points to be densely gathered, we remove them using the DBSCAN algorithm with $\epsilon=50$ and plot the rest in Fig.~\ref{fig:mat_pca}.
It can be observed that the experts distribute rather densely, especially for $W_{\text{up}}$.
Although the distribution of experts varies for three matrices, the figures for $W_{\text{up}}$ and $W_{\text{down}}$ are more similar than those of the gate matrix. 

\noindent\textbf{Grok.}
Typically, about half of the Grok experts densely gather for $W_{\text{up}}$ and $W_{\text{down}}$.
The other half turns out to be outliers even though no dark cross were observed before. .
Furthermore, the outliers of the three matrices partially coincide.

\subsection{Neuron-level} 
\label{append:neuron_pca}

\begin{figure*}[htbp]
    \centering
    $\vcenter{\hbox{\includegraphics[width=.1\linewidth]{mixtral.pdf} }}$ \\
    $\vcenter{\hbox{\includegraphics[width=.49\linewidth]{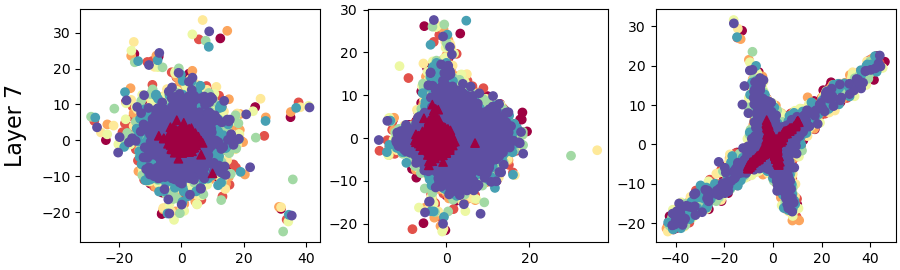} }}$ 
    $\vcenter{\hbox{\includegraphics[width=.49\linewidth]{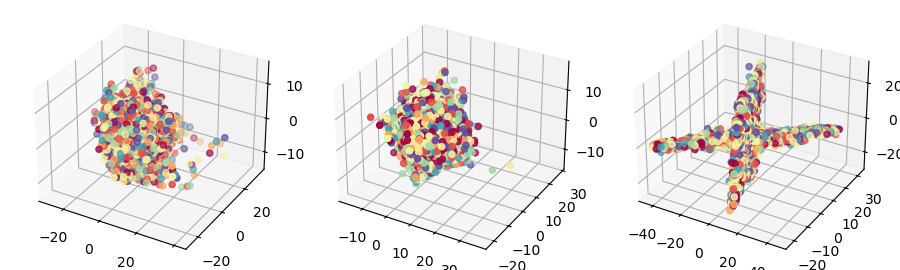} }}$ \\
    \vspace{1.mm}\rule{\linewidth}{0.5pt}
    $\vcenter{\hbox{\includegraphics[width=.1\linewidth]{deepseek.pdf} }}$ \\
    $\vcenter{\hbox{\includegraphics[width=.49\linewidth]{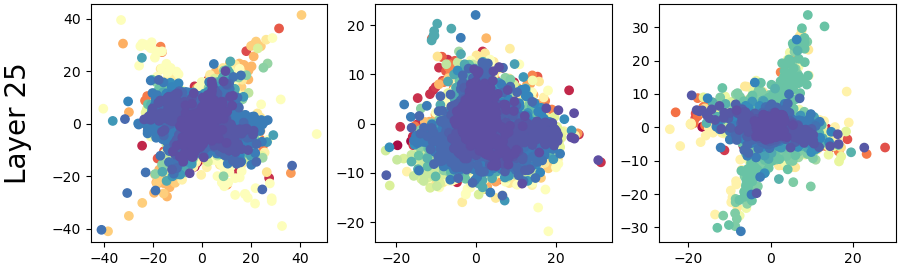} }}$ 
    $\vcenter{\hbox{\includegraphics[width=.49\linewidth]{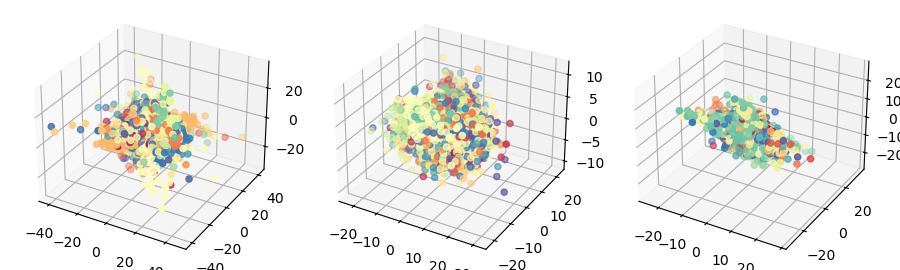} }}$ \\
    \vspace{1.mm}\rule{\linewidth}{0.5pt}
    $\vcenter{\hbox{\includegraphics[width=.1\linewidth]{grok.pdf} }}$ \\
    $\vcenter{\hbox{\includegraphics[width=.49\linewidth]{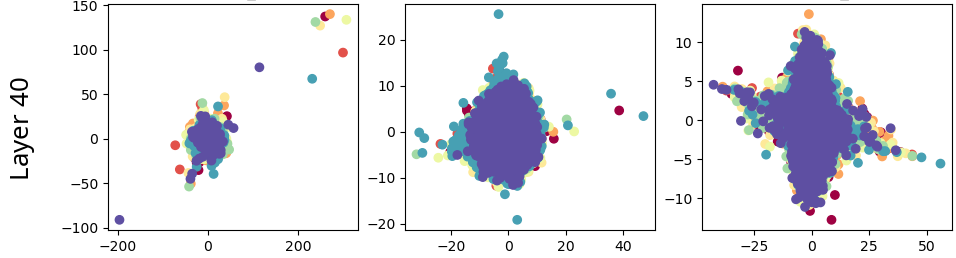} }}$ 
    $\vcenter{\hbox{\includegraphics[width=.49\linewidth]{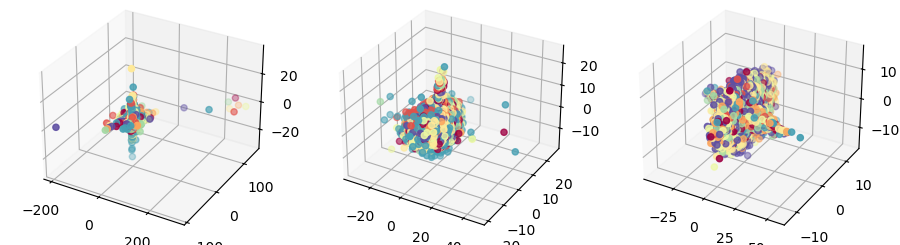} }}$
    \caption{\footnotesize Projection of expert neurons in 2D/3D space. Each layer contains three graphs, corresponding to $W_{\text{up}}$, $W_{\text{act}}$, and $W_{\text{down}}$, respectively.}
    \label{fig:neuron_pca}
\end{figure*}

To project the neurons into a 2D or 3D space, each row vector of $W_{\text{up}}$ and $W_{\text{act}}$, or each column vector of $W_{\text{down}}$, is treated as a single data point.
Standardization is then applied, following by PCA.
The visualization of the principal components is illustrated in Fig.~\ref{fig:neuron_pca}.
Different colors refer to neurons belonging to different experts.

\noindent\textbf{Common.}
The vast majority of neurons gather in the low-dimensional space.
In some layers, the distribution of neurons forms a special shape, such as a cross or a thick line, which appears the most often for $W_{\text{down}}$, followed by $W_{\text{up}}$, and finally $W_{\text{act}}$.
Compared to ellipses, these shapes indicate that the neurons are relatively more similar. 

\noindent\textbf{Mixtral and Mistral.}
The neurons in the Mistral FFN distribute more densely than those of the Mixtral experts.
Notably, the distribution shape of neurons in the FFN and experts are usually alike, even for the outliers. 

\noindent\textbf{DeepSeek and Grok.}
The number of outliers is a bit greater tahn that observed in Mixtral.

\section{Averaging Expert Neurons}
\label{append:averaging}

To investigate expert correlation at the neuron level, the averaging approach simply averages the rows (for $W_{\text{up}}$ and $W_{\text{act}}$) or the columns (for $W_{\text{down}}$) of the weight matrices and then calculates the similarity of the resulting vectors across experts.
Fig.~\ref{fig:gate_sim} displays the graphs.

\noindent\textbf{Common.}
The heat maps of $W_{\text{up}}$ and $W_{\text{down}}$ are nearly identical to those presented in \S~\ref{exp:mat_sim}.
Yet the similarities of $W_{\text{act}}$ significantly increase.

\noindent\textbf{Mixtral and Mistral.}
% After averaging the neurons, the heat maps of $W_{\text{up}}$ and $W_{\text{down}}$ are almost the same as in Section~\ref{exp:mat_sim}.
% Yet the similarities of $W_{\text{act}}$ significantly increase.
% , indicating they might be underestimated in the matrix-level experiment due to the position issue.
The dark crosses sometimes disappear. 
In the figures for $W_{\text{act}}$, the similarities between the experts and the Mistral FFN are often lower than the similarities among the experts themselves (\textit{i.e.}, $S_{\text{ee}}>S_{\text{ef}}$), which is contrary to previous observations.
This can happen if the expert neurons in different positions are alike.
For instance, given three vectors $f=(0, 0)$, $e_1=(1, 0)$, and $e_2=(0, 1)$, the vector similarity $S_{e_1e_2}$ is lower than $S_{e_1f}$ and $S_{e_2f}$. 
If averaging the elements, we have $\bar{f}=(0)$, $\bar{e}_1=(0.5)$, and $\bar{e}_2=(0.5)$, then $S_{e_1e_2}$ becomes the highest.

\noindent\textbf{DeepSeek.}
The growth of the $W_{\text{act}}$ similarity values is directly proportional to the layer depth.

\noindent\textbf{Gork.}
In the heat map of $W_{\text{act}}$, dark crosses frequently appear in various positions. 

\section{Gate Embedding}
\label{append:gate_sim}

\begin{figure*}[thbp]
    \centering
    $\vcenter{\hbox{\includegraphics[width=.09\linewidth]{mixtral.pdf} }}$
    $\vcenter{\hbox{\includegraphics[width=.44\linewidth]{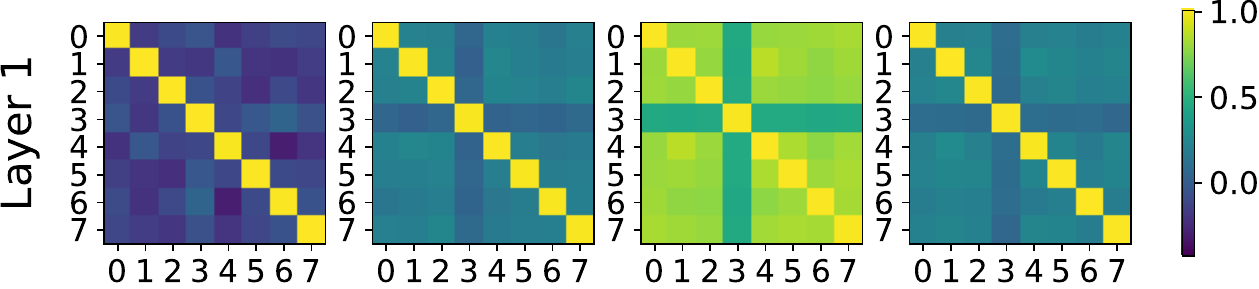} }}$
    $\vcenter{\hbox{\includegraphics[width=.44\linewidth]{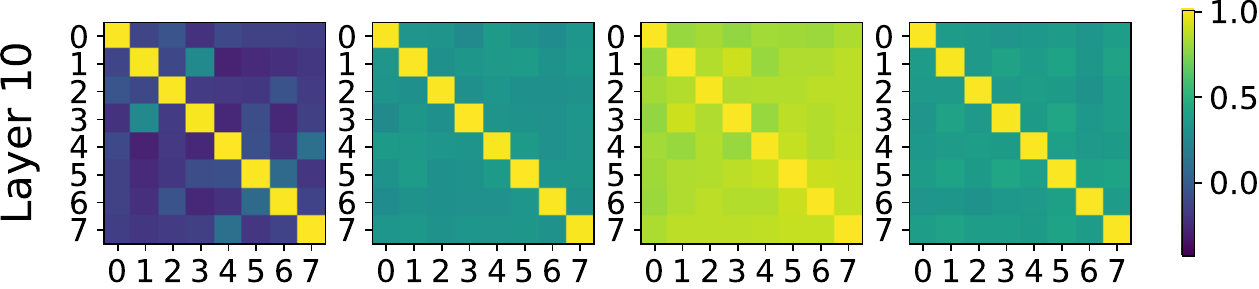} }}$ \\
    $\vcenter{\hbox{\includegraphics[width=.09\linewidth]{blank.pdf} }}$
    $\vcenter{\hbox{\includegraphics[width=.44\linewidth]{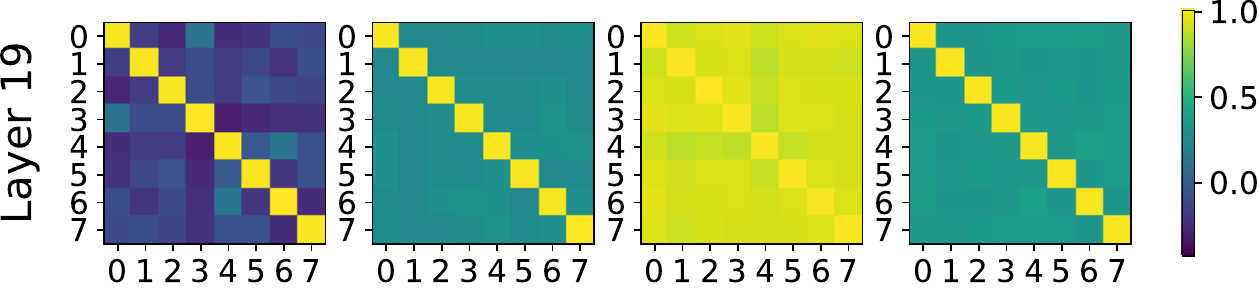} }}$
    $\vcenter{\hbox{\includegraphics[width=.44\linewidth]{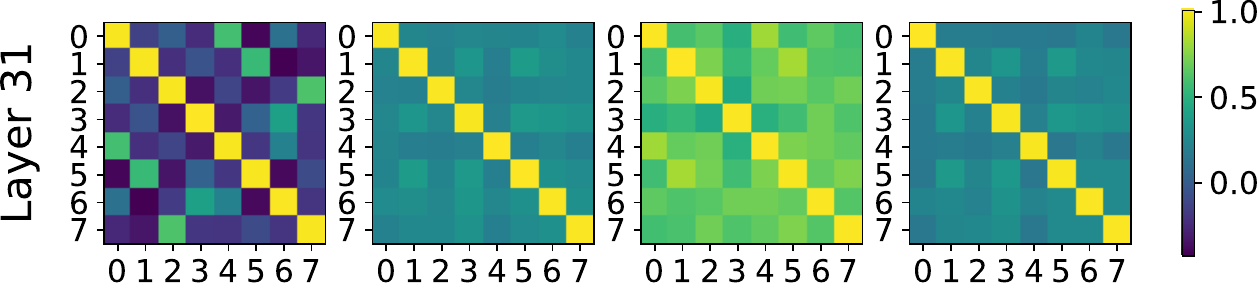} }}$ \\
    \vspace{1.mm}\rule{\linewidth}{0.5pt}\vspace{1.mm} \\
    $\vcenter{\hbox{\includegraphics[width=.09\linewidth]{mixtral_22.pdf} }}$
    $\vcenter{\hbox{\includegraphics[width=.44\linewidth]{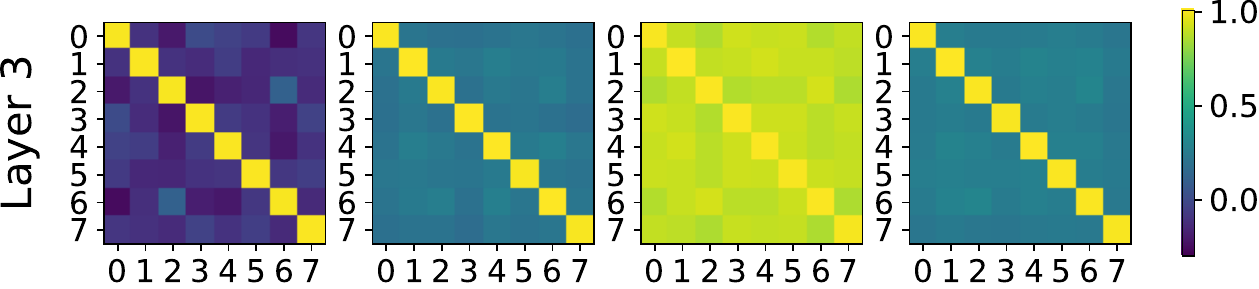} }}$
    $\vcenter{\hbox{\includegraphics[width=.44\linewidth]{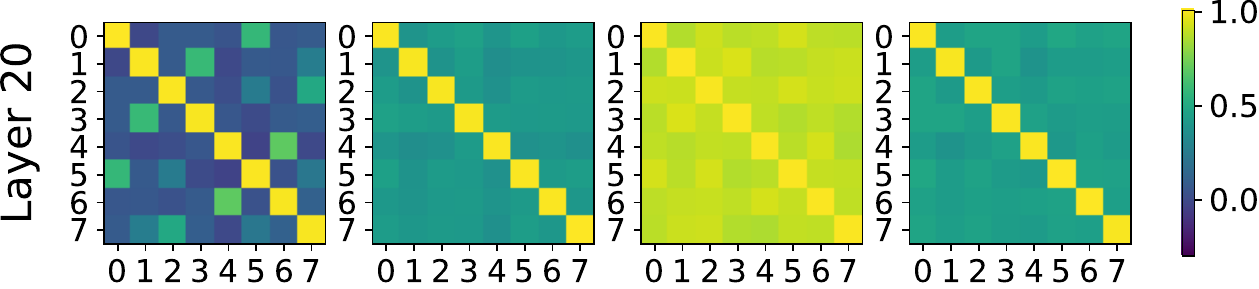} }}$ \\
    $\vcenter{\hbox{\includegraphics[width=.09\linewidth]{blank.pdf} }}$
    $\vcenter{\hbox{\includegraphics[width=.44\linewidth]{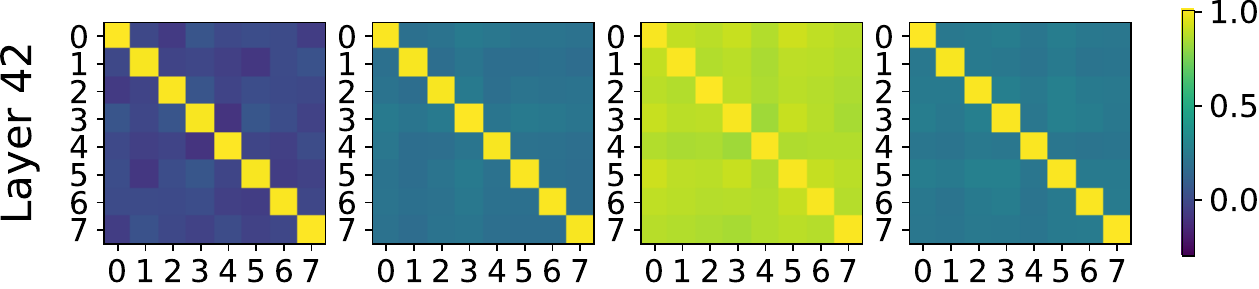} }}$
    $\vcenter{\hbox{\includegraphics[width=.44\linewidth]{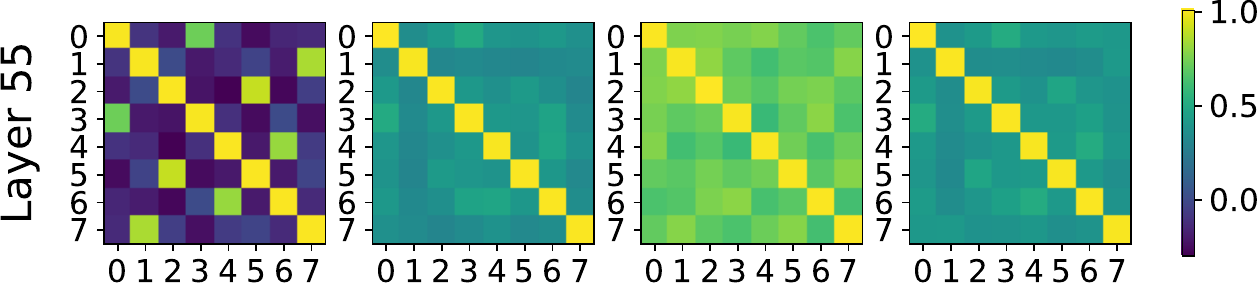} }}$ \\
    \vspace{1.mm}\rule{\linewidth}{0.5pt}\vspace{1.mm} \\
    $\vcenter{\hbox{\includegraphics[width=.09\linewidth]{deepseek.pdf} }}$
    $\vcenter{\hbox{\includegraphics[width=.44\linewidth]{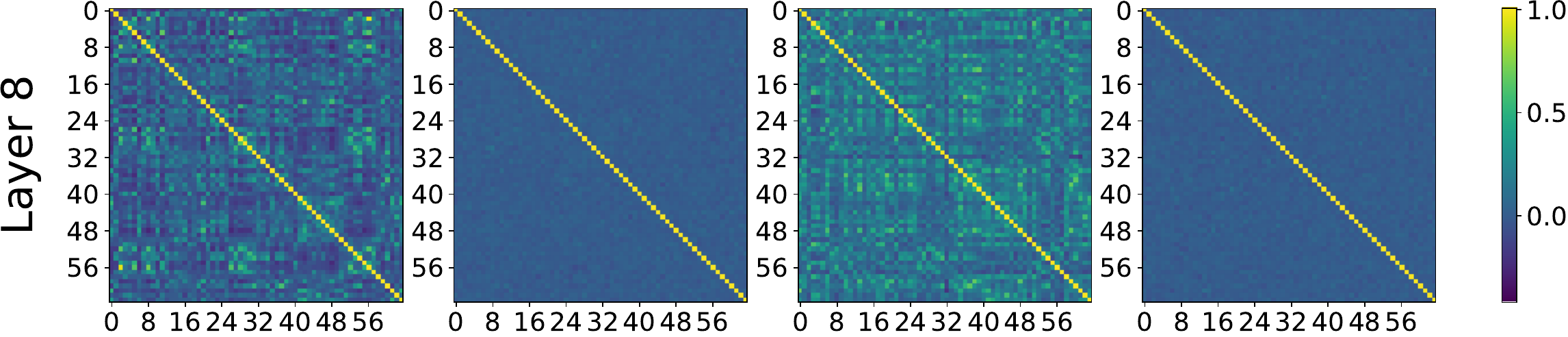} }}$
    $\vcenter{\hbox{\includegraphics[width=.44\linewidth]{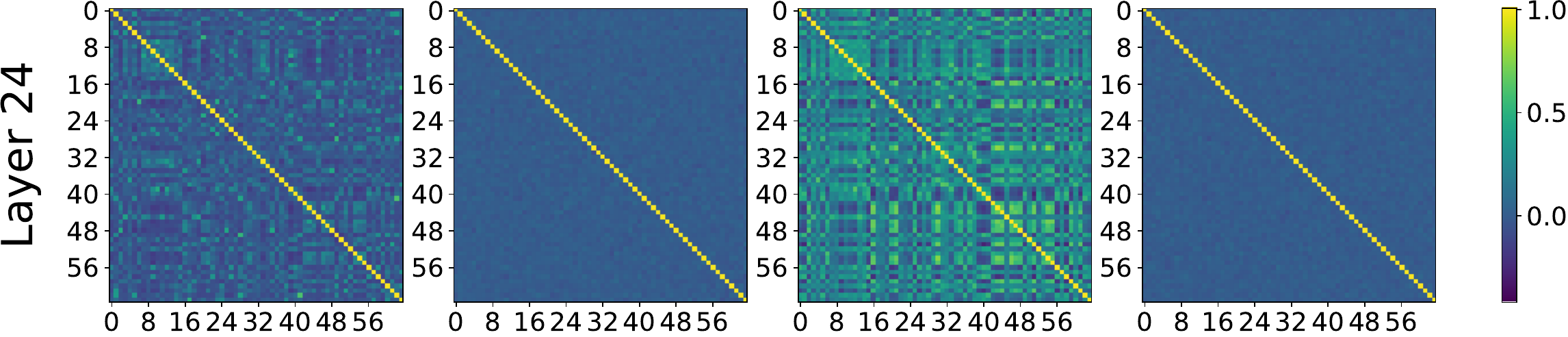} }}$ \\
    \vspace{1.mm}\rule{\linewidth}{0.5pt}\vspace{1.mm} \\
    $\vcenter{\hbox{\includegraphics[width=.09\linewidth]{grok.pdf} }}$
    $\vcenter{\hbox{\includegraphics[width=.44\linewidth]{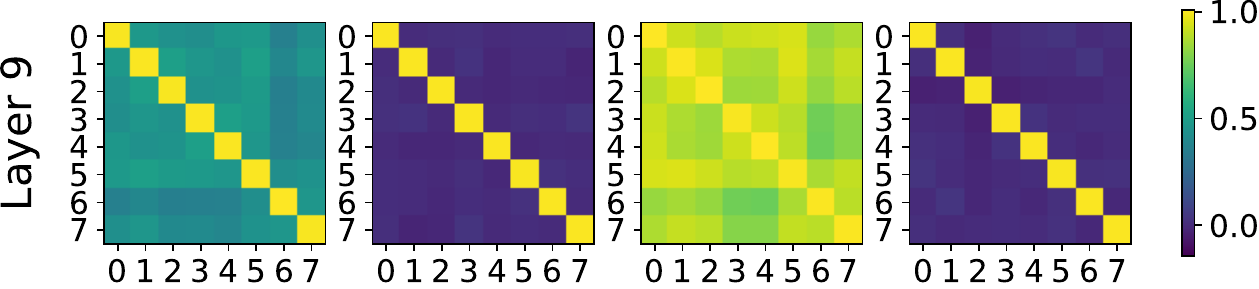} }}$
    $\vcenter{\hbox{\includegraphics[width=.44\linewidth]{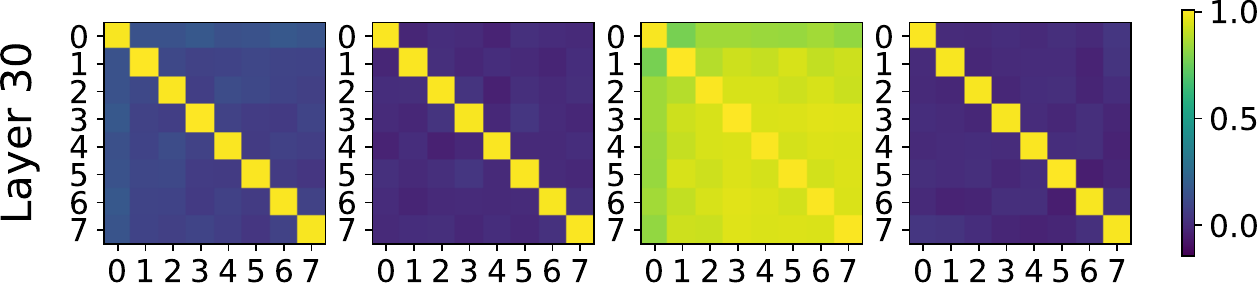} }}$ \\
    $\vcenter{\hbox{\includegraphics[width=.09\linewidth]{blank.pdf} }}$
    $\vcenter{\hbox{\includegraphics[width=.44\linewidth]{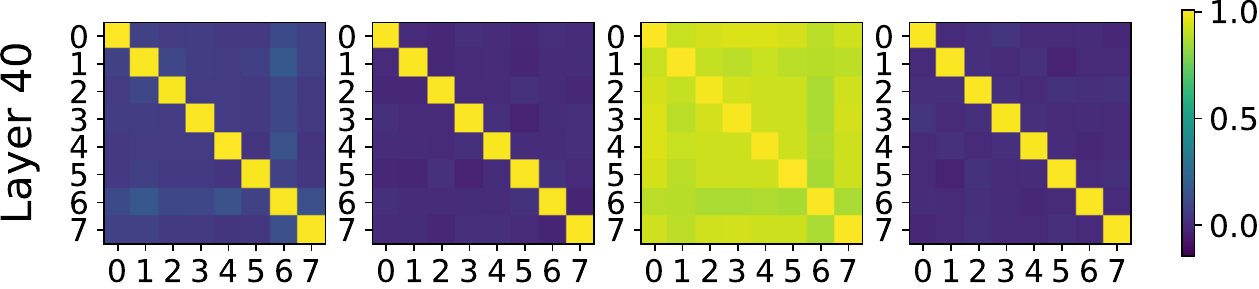} }}$
    $\vcenter{\hbox{\includegraphics[width=.44\linewidth]{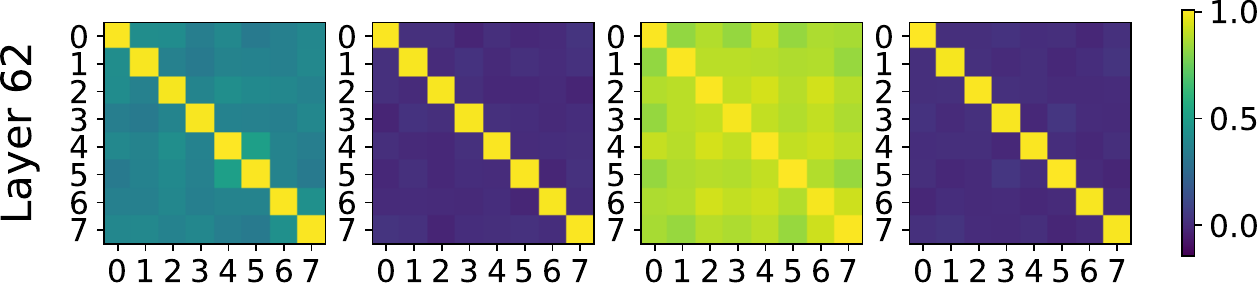} }}$ \\
    \caption{\footnotesize Similarity heat maps of gate embedding (leftmost graph of each layer) along with the neuron-level similarity heat maps using averaging method. The tick numbers refer to the expert indices.}
    \label{fig:gate_sim}
\end{figure*}

\begin{table}[htbp]
    \centering
    \small
    \begin{tabular}{crrrr}
        \toprule
        \textbf{Layer} & \textbf{Mixtral} & \textbf{Mixtral-22} & \textbf{DeepSeek} & \textbf{Grok} \\
        \midrule
        0 & 0.82 & -0.01 & -{}-{}-  & 0.89 \\
        1 & -0.44 & 0.23 & 0.75  & -0.10  \\
        2 & 0.26 & 0.62 & 0.78  & -0.28  \\
        3 & 0.54 & 0.76 & 0.71  & 0.66  \\
        4 & 0.48 & 0.47 & 0.77  & 0.52  \\
        5 & 0.70 & 0.49 & 0.77  & 0.37  \\
        6 & 0.84 & 0.28 & 0.69  & 0.28  \\
        7 & 0.74 & 0.69 & 0.73  & 0.17  \\
        8 & 0.42 & 0.66 & 0.66  & 0.51  \\
        9 & 0.66 & 0.84 & 0.66  & 0.84  \\
        10 & 0.53 & 0.59 & 0.63  & 0.28  \\
        11 & 0.32 & 0.61 & 0.60  & 0.30  \\
        12 & 0.14 & 0.55 & 0.54  & 0.46  \\
        13 & 0.51 & 0.48 & 0.60  & 0.14  \\
        14 & 0.66 & 0.62 & 0.56  & 0.00  \\
        15 & 0.40 & 0.59 & 0.58  & 0.54  \\
        16 & 0.39 & 0.68 & 0.53  & 0.32  \\
        17 & 0.53 & 0.65 & 0.55  & 0.30  \\
        18 & 0.35 & 0.66 & 0.57  & 0.10  \\
        19 & 0.17 & 0.72 & 0.57  & -0.17  \\
        20 & 0.51 & 0.77 & 0.58  & 0.24  \\
        21 & 0.63 & 0.67 & 0.62  & 0.58  \\
        22 & 0.36 & 0.65 & 0.62  & 0.46  \\
        23 & 0.51 & 0.35 & 0.62  & 0.14  \\
        24 & 0.48 & 0.30 & 0.68  & 0.10  \\
        25 & 0.66 & 0.13 & 0.62  & 0.00  \\
        26 & 0.81 & 0.27 & 0.58  & -0.10  \\
        27 & 0.63 & 0.22 & 0.46  & -0.26  \\
        28 & 0.73 & 0.38 & -{}-{}-  & -0.66  \\
        29 & 0.75 & 0.52 & -{}-{}-  & -0.41  \\
        30 & 0.84 & 0.38 & -{}-{}-  & -0.83  \\
        31 & 0.57 & 0.45 & -{}-{}-  & -0.76  \\
        32 & -{}-{}- & 0.37 & -{}-{}-  & -0.24  \\
        33 & -{}-{}- & 0.28  & -{}-{}-  & -0.53  \\
        34 & -{}-{}- & 0.72 & -{}-{}-  & -0.46  \\
        35 & -{}-{}- & 0.69 & -{}-{}-  & 0.14  \\
        36 & -{}-{}- & 0.34 & -{}-{}-  & 0.17  \\
        37 & -{}-{}- & 0.46 & -{}-{}-  & -0.46  \\
        38 & -{}-{}- & 0.31 & -{}-{}-  & -0.17  \\
        39 & -{}-{}- & 0.26 & -{}-{}-  & -0.26  \\
        40 & -{}-{}- & 0.48 & -{}-{}-  & -0.70  \\
        41 & -{}-{}- & 0.41 & -{}-{}-  & 0.17  \\
        42 & -{}-{}- & 0.46 & -{}-{}-  & 0.00  \\
        43 & -{}-{}- & 0.33 & -{}-{}-  & -0.17  \\
        44 & -{}-{}- & 0.44 & -{}-{}-  & -0.22  \\
        45 & -{}-{}- & 0.50 & -{}-{}-  & 0.14  \\
        46 & -{}-{}- & 0.43 & -{}-{}-  & -0.47  \\
        47 & -{}-{}- & 0.34 & -{}-{}-  & -0.44  \\
        48 & -{}-{}- & 0.50 & -{}-{}-  & -0.17  \\
        49 & -{}-{}- & 0.42 & -{}-{}-  & -0.14  \\
        50 & -{}-{}- & 0.43 & -{}-{}-  & 0.17  \\
        51 & -{}-{}- & 0.51 & -{}-{}-  & 0.22  \\
        52 & -{}-{}- & 0.67 & -{}-{}-  & 0.10  \\
        53 & -{}-{}- & 0.32 & -{}-{}-  & 0.33  \\
        54 & -{}-{}- & 0.68 & -{}-{}-  & -0.24  \\
        55 & -{}-{}- & 0.20 & -{}-{}-  & -0.57  \\
        56 & -{}-{}- & -{}-{}- & -{}-{}-  & -0.24  \\
        57 & -{}-{}- & -{}-{}- & -{}-{}-  & -0.37  \\
        58 & -{}-{}- & -{}-{}- & -{}-{}-  & 0.00  \\
        59 & -{}-{}- & -{}-{}- & -{}-{}-  & -0.69  \\
        60 & -{}-{}- & -{}-{}- & -{}-{}-  & -0.17  \\
        61 & -{}-{}- & -{}-{}- & -{}-{}-  & 0.35  \\
        62 & -{}-{}- & -{}-{}- & -{}-{}-  & 0.30  \\
        63 & -{}-{}- & -{}-{}- & -{}-{}-  & 0.10  \\
        \bottomrule
    \end{tabular}
    \caption{\footnotesize Pearson correlation coefficients ($R$) of the paired dataset $(X, Y_{\text{act}})$.}
    \label{tab:R}
\end{table}

Since the gate embedding matrix $W_g$ determines the gate decision, there may be a relationship between $W_g$ and the experts. 
To investigate this, we measure the similarities between the gate embedding vectors, $W_g[n,:]$ for $n \in [1, N]$, and compare them with the neuron-level averaging heat maps of experts presented in \S~\ref{exp:neuron_sim}.
The qualitative analysis of the combined graphs shown in Fig.~\ref{fig:gate_sim} is detailed in this section. 
The table containing the $R$ values for each layer (Tab.~\ref{tab:R}) is appended at the end. 

\noindent\textbf{Mixtral.}
Focusing on the heat maps of $W_g$, the similarities typically range from 0.2 to 0.4, with a noticeable increase in the last layer.
Moreover, dark crosses are rarely found.
Surprisingly, the patterns in the heat maps of $W_g$ and of expert neurons in $W_{\text{act}}$ are partially alike in some layers.
This implies that the way a gate selects experts might be relevant to how an expert activates its neurons.

\noindent\textbf{DeepSeek.}
Unlike the almost all-zero heat maps of $W_{\text{up}}$ and $W_{\text{down}}$, the similarities of gate neurons sometimes exceed 0.4.
In addition, the heat maps of $W_g$ and $W_{\text{act}}$ show similar patterns.
However, the overall similarities of $W_g$ decrease with depth while the similarities of $W_{\text{act}}$ gradually grow.
This indicates that as the layer depth increases, the gate ``looks'' at the input feature in more diverse ways when assigning scores to different experts, even as the neuron activations of the experts become more similar.

\noindent\textbf{Grok.}
Both dark and bright crosses commonly exist in the heat maps of $W_g$, whose patterns are similar to those of $W_{\text{act}}$.
Specially, their patterns show opposite color tendency (\textit{i.e.}, deep color positions in one heat map becomes light color in another) starting form the intermediate layers.
The similarities of $W_g$ decrease when the layer depth increases, except for the last few layers.

\section{Additional Datasets}
\label{append:dataset}

\begin{figure*}[t!]
    \centering
    $\vcenter{\hbox{\includegraphics[width=.12\linewidth]{mixtral.pdf}}}$ \\
    $\vcenter{\hbox{\includegraphics[width=.115\linewidth]{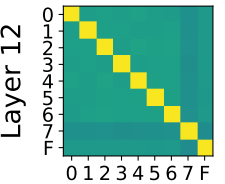}}}$ 
    $\vcenter{\hbox{\includegraphics[width=.1\linewidth]{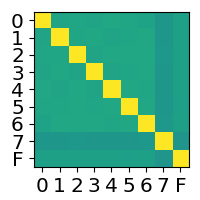}}}$ 
    $\vcenter{\hbox{\includegraphics[width=.1\linewidth]{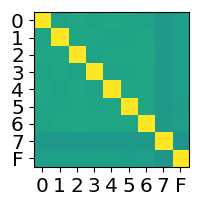}}}$
    $\vcenter{\hbox{\includegraphics[width=.13\linewidth]{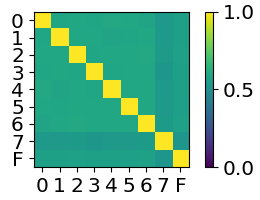}}}$
    \hspace{2mm}
    $\vcenter{\hbox{\includegraphics[width=.115\linewidth]{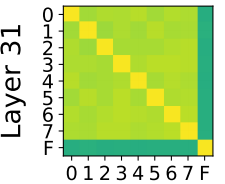} }}$ 
    $\vcenter{\hbox{\includegraphics[width=.1\linewidth]{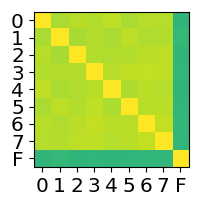}}}$ 
    $\vcenter{\hbox{\includegraphics[width=.1\linewidth]{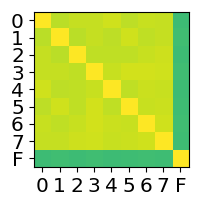}}}$
    $\vcenter{\hbox{\includegraphics[width=.13\linewidth]{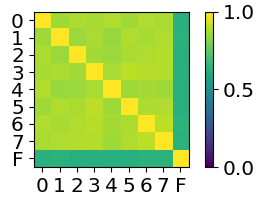}}}$ \\
    
    $\vcenter{\hbox{\includegraphics[width=.12\linewidth]{deepseek.pdf}}}$ \\
    $\vcenter{\hbox{\includegraphics[width=.11\linewidth]{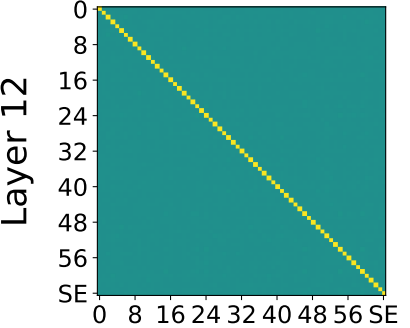}}}$ 
    $\vcenter{\hbox{\includegraphics[width=.1\linewidth]{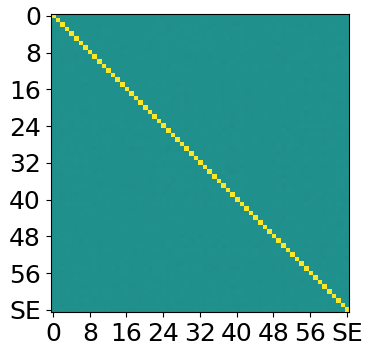}}}$ 
    $\vcenter{\hbox{\includegraphics[width=.1\linewidth]{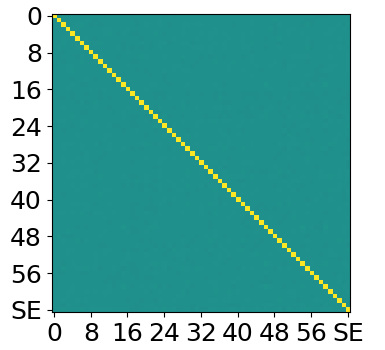}}}$
    $\vcenter{\hbox{\includegraphics[width=.135\linewidth]{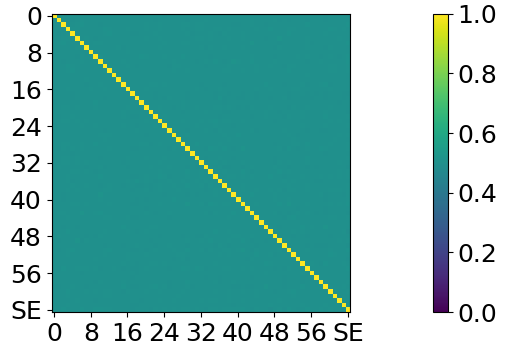}}}$
    \hspace{2mm}
    $\vcenter{\hbox{\includegraphics[width=.11\linewidth]{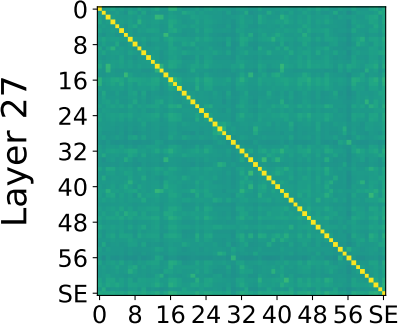} }}$ 
    $\vcenter{\hbox{\includegraphics[width=.1\linewidth]{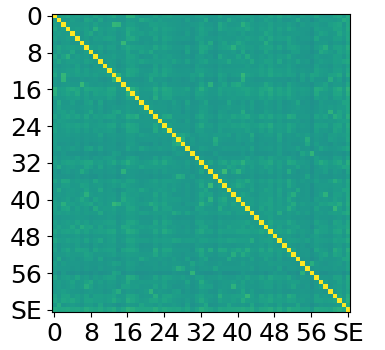}}}$ 
    $\vcenter{\hbox{\includegraphics[width=.1\linewidth]{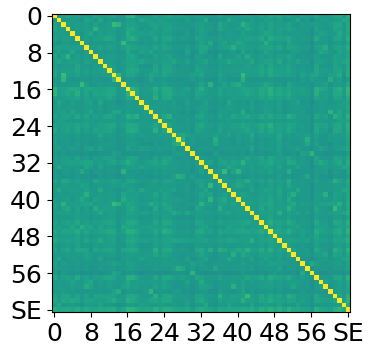}}}$
    $\vcenter{\hbox{\includegraphics[width=.135\linewidth]{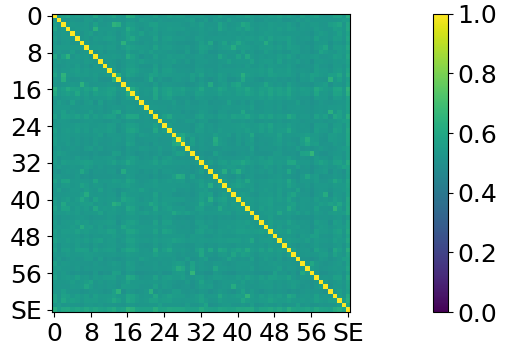}}}$ \\
    \caption{\footnotesize Average similarity heat maps of expert output features using (1) the long sequence, (2) WikiText-103, (3) GSM8K, and (4) Magicoder-Evol-Instruct-110K. The tick numbers refer to the expert indices. ``F'' and ``SE'' denote the Mistral FFN and the DeepSeek shared expert, respectively.}
    \label{fig:out_avgsim_extra}
\end{figure*}

To ensure the universality of our findings, we repeat the experiments that require the long input (§ 5.1, § 5.2) using additional datasets.
Specifically, we utilize the entire test set of WikiText-103~\cite{merity2016pointer} (266K tokens) and of a math dataset GSM8K~\cite{cobbe2021gsm8k} (84K tokens), and 1000 sequences from the code dataset Magicoder-Evol-Instruct-110K~\cite{wei2024magicoder} (188K tokens).
As shown in Fig.~\ref{fig:out_avgsim_extra},~\ref{fig:reg_norm_mixtral},~\ref{fig:reg_norm_deepseek}, the new figures of Mixtral and DeepSeek align with the previous results illustrated in the main context, even when using datasets of specific subjects like math and code (we did not test on the Grok model due to limited computation resources). 
These supplementary results demonstrate that our findings are general and not limited to the initial input sources.

\section{Norms of Expert Outputs and Gate Scores} 
\label{append:norm}

\begin{figure*}[htbp]
    \centering
    $\vcenter{\hbox{\includegraphics[width=.12\linewidth]{mixtral.pdf} }}$ \hspace{6.cm}
    $\vcenter{\hbox{\includegraphics[width=.12\linewidth]{mixtral_22.pdf} }}$ \\
    $\vcenter{\hbox{\includegraphics[width=.48\linewidth]{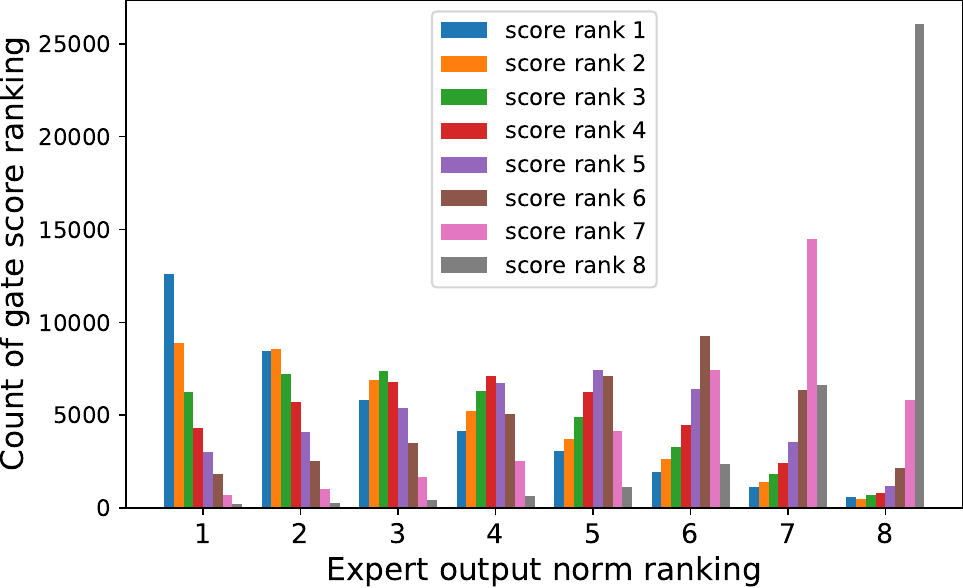} }}$ 
    $\vcenter{\hbox{\includegraphics[width=.48\linewidth]{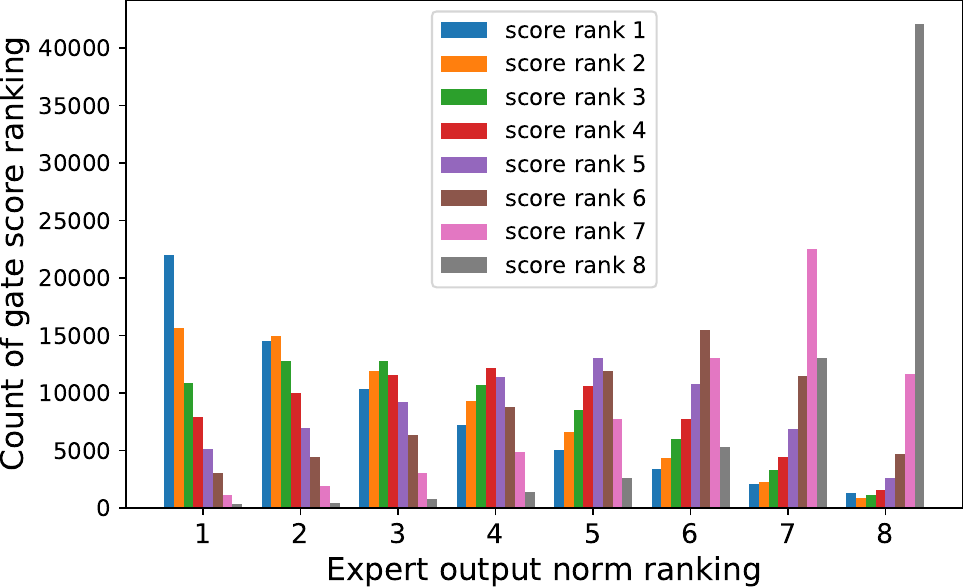} }}$ \\
    $\vcenter{\hbox{\includegraphics[width=.12\linewidth]{grok.pdf} }}$ \\
    $\vcenter{\hbox{\includegraphics[width=.48\linewidth]{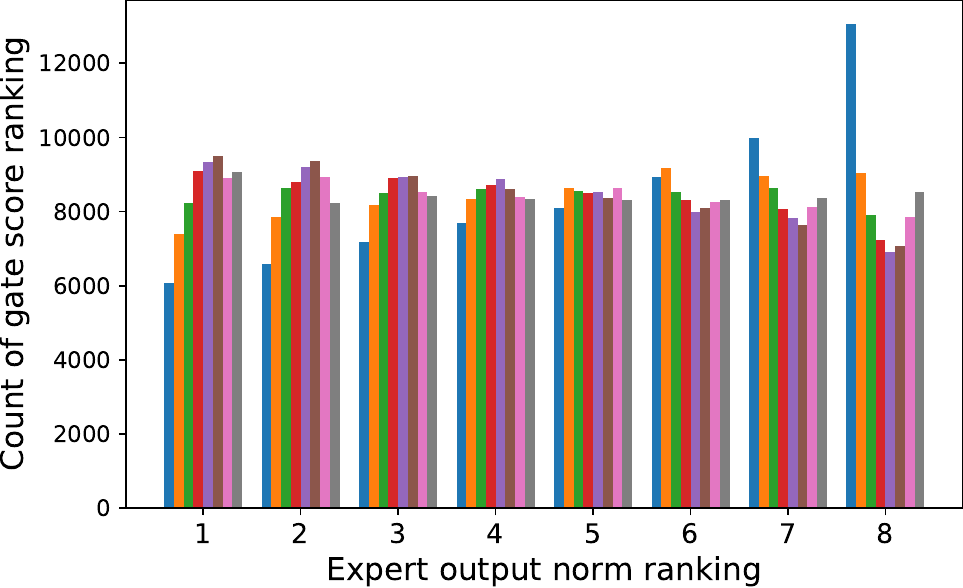} }}$ \\
    $\vcenter{\hbox{\includegraphics[width=.12\linewidth]{deepseek.pdf} }}$ \\
    $\vcenter{\hbox{\includegraphics[width=.95\linewidth]{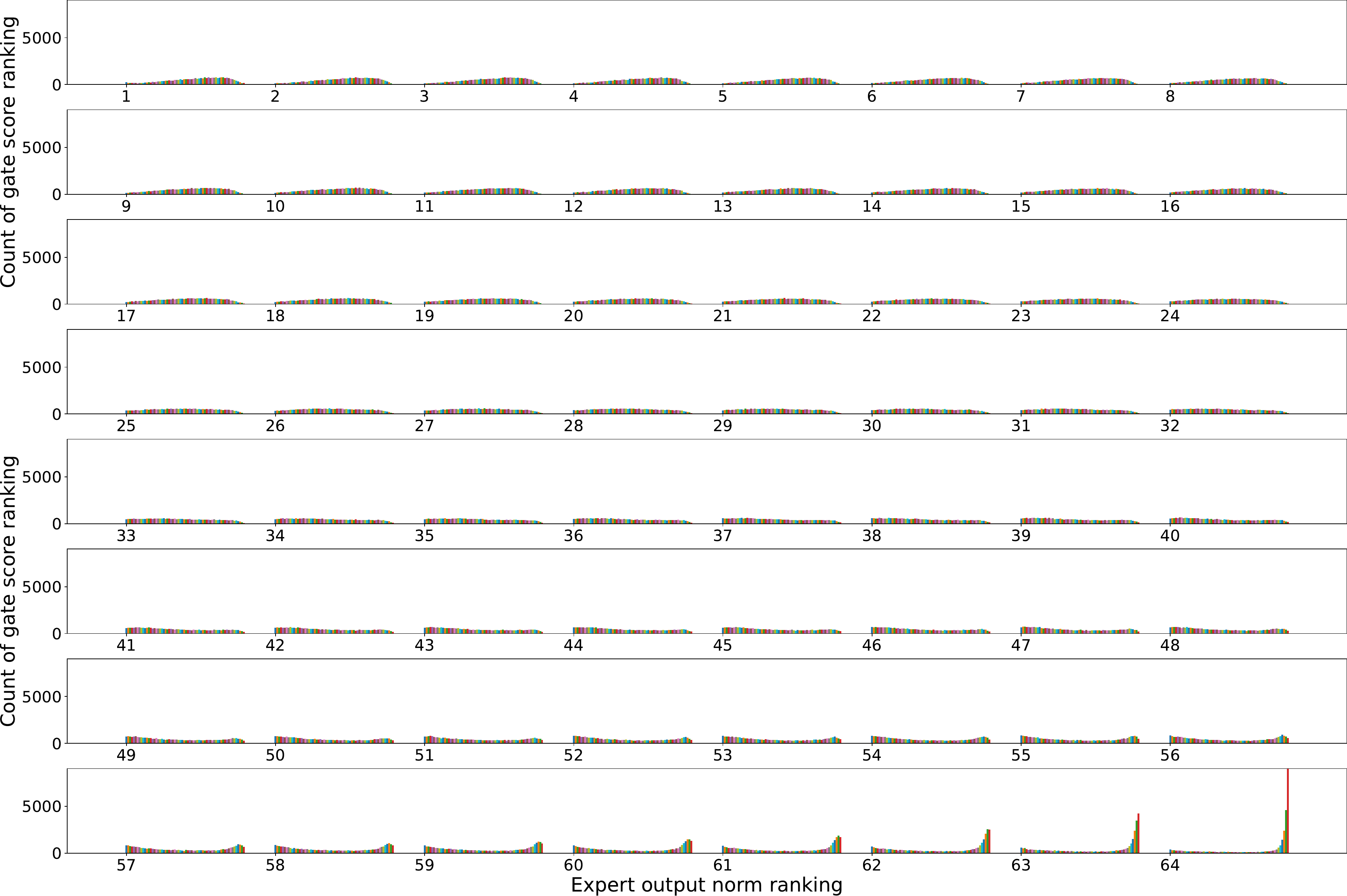} }}$ \\
    \caption{\footnotesize Counts of the gate score ranking for each norm ranking using the long input. The larger the rank number, the larger the norm or score.}
    \label{fig:reg_norm}
\end{figure*}

\begin{figure}[thbp]
    \centering
    \includegraphics[width=.95\linewidth]{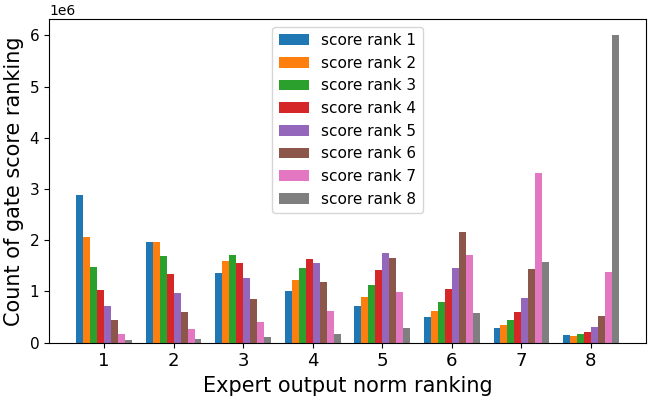}
    \includegraphics[width=.95\linewidth]{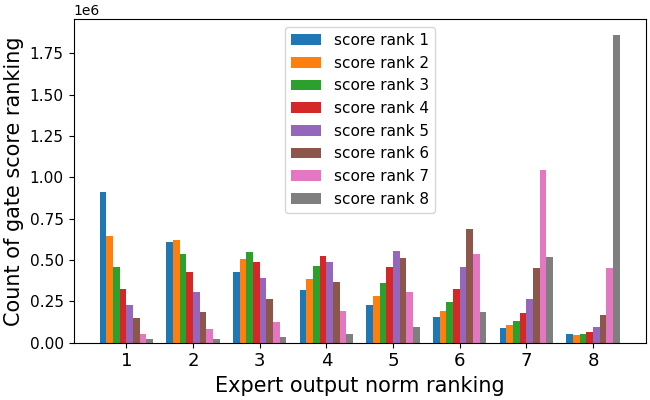}
    \includegraphics[width=.95\linewidth]{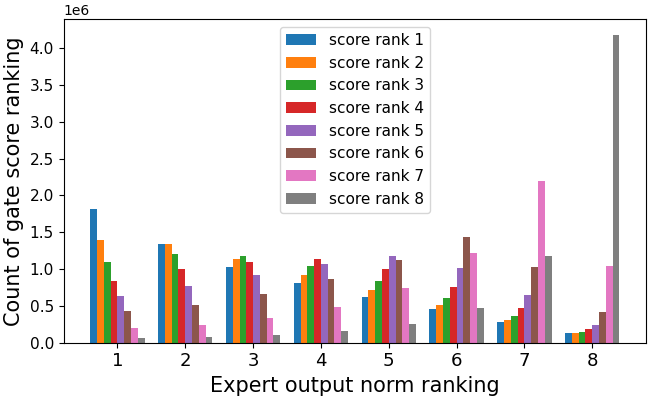}
    \caption{\footnotesize Counts of the gate score ranking for Mixtral expert ourput norm rankings using additional datasets, namely WikiText-103 (\textbf{top}), GSM8K (\textbf{middle}), and Magicoder-Evol-Instruct-110K (\textbf{bottom}). The larger the rank number, the larger the norm or score.}
    \label{fig:reg_norm_mixtral}
\end{figure}

\begin{table*}[thbp]
    \small
    \centering
    \begin{tabular}{ccccccc}
        \toprule
        \textbf{non-MoE layer idx} & \textbf{PPL} $\downarrow$ & \textbf{Bench avg} $\uparrow$ & \textbf{HellaSwag} & \textbf{MMLU} & \textbf{GSM8K} & \textbf{CEval} \\
        \midrule
        none & 8.51 & 39.84 & 61.75 & 41.14 & 13.42 & 43.04 \\
        1 & 8.50 & 39.42 & 61.88 & 40.55 & 11.83 & 43.41 \\
        6 & 8.49 & 38.55 & 62.17 & 39.28 & 12.96 & 39.78 \\
        12 & 8.51 & 38.21 & 61.46 & 40.24 & 10.54 & 40.61 \\
        18 & 8.59 & 37.99 & 61.73 & 39.50 & 11.68 & 39.03 \\
        24 & 8.58 & 38.94 & 61.33 & 40.45 & 12.05 & 41.91 \\
        \bottomrule
    \end{tabular}
    \caption{\footnotesize Model performance on various benchmarks for the dynamic expert numbers experiment. ``Bench avg'' refers to the average performance over the four evaluated benchmarks.}
    \label{tab:dynamic_moe}
\end{table*}

\begin{figure}[thbp]
    \centering
    % $\vcenter{\hbox{\includegraphics[width=.09\linewidth]{mixtral.pdf} }}$ 
    % \hspace{6.2cm}
    % $\vcenter{\hbox{\includegraphics[width=.09\linewidth]{grok.pdf} }}$ \\
    $\vcenter{\hbox{\includegraphics[width=.95\linewidth]{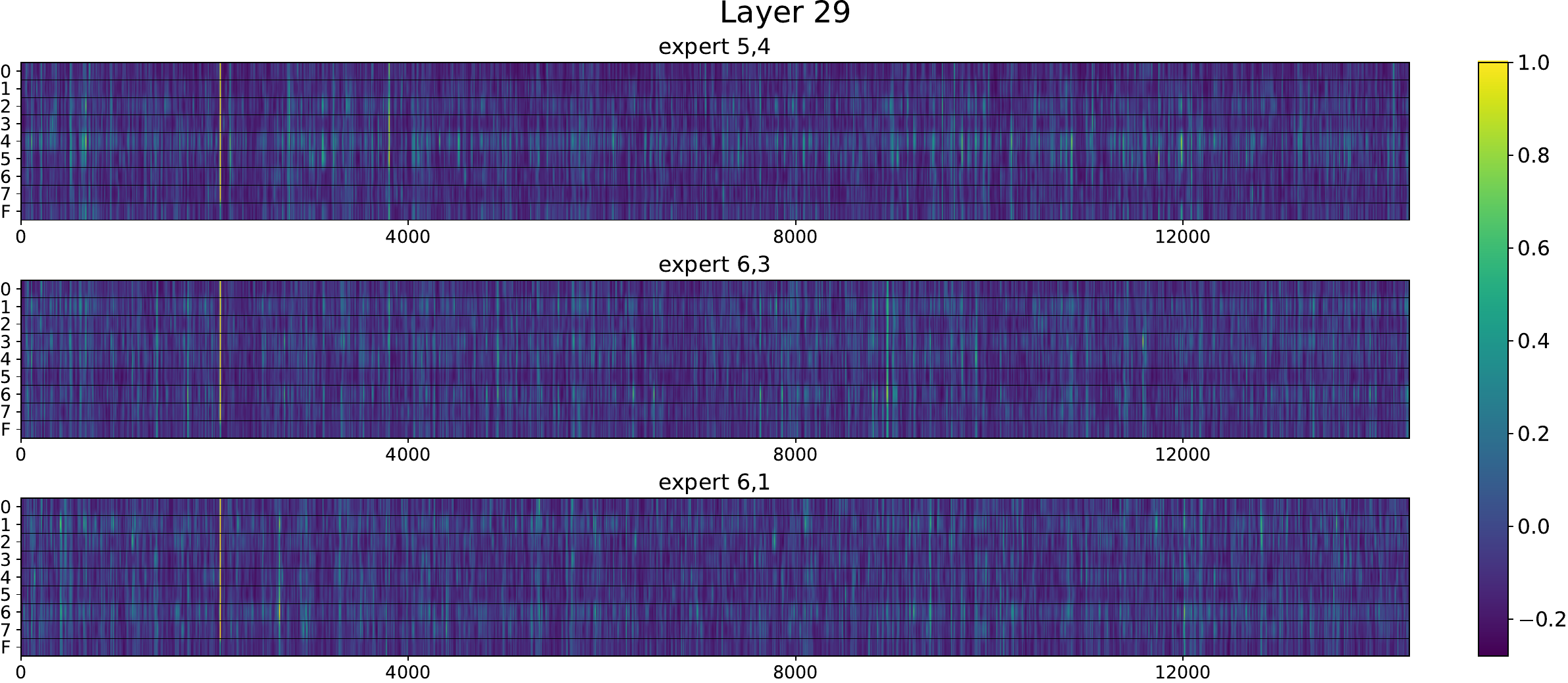} }}$
    % $\vcenter{\hbox{\includegraphics[width=.48\linewidth]{exp-intermidiate/grok/layer_31.pdf} }}$ \\
    % \vspace{1mm}\rule{\linewidth}{0.5pt}\vspace{1mm}
    % $\vcenter{\hbox{\includegraphics[width=.09\linewidth]{deepseek.pdf} }}$ 
    % $\vcenter{\hbox{\includegraphics[width=.8\linewidth]{exp-intermidiate/deepseek/layer_4.pdf} }}$ \\
    % $\vcenter{\hbox{\includegraphics[width=.09\linewidth]{blank.pdf} }}$
    % $\vcenter{\hbox{\includegraphics[width=.8\linewidth]{exp-intermidiate/deepseek/layer_24.pdf} }}$ \\
    \caption{\footnotesize Intermediate state values of Mixtral experts. The top $k$ experts are shown on top of each heat map. Each number in the vertical axis refers to an expert index while the horizontal axis represents the number of neurons.}
    \label{fig:inter}
\end{figure}

\begin{figure*}[htbp]
    \centering
    \includegraphics[width=.7\linewidth]{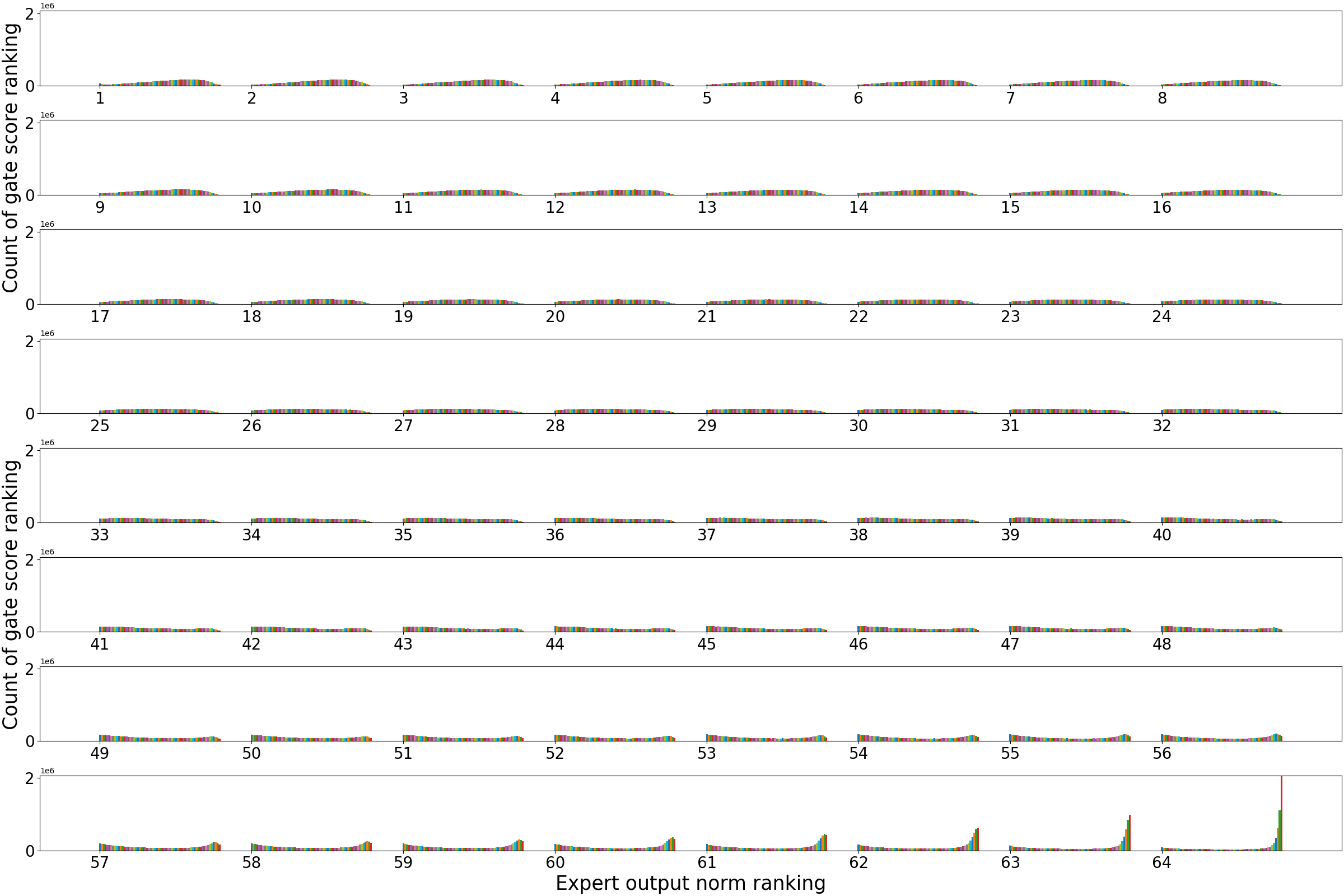} \\
    \vspace{2mm}
    \includegraphics[width=.7\linewidth]{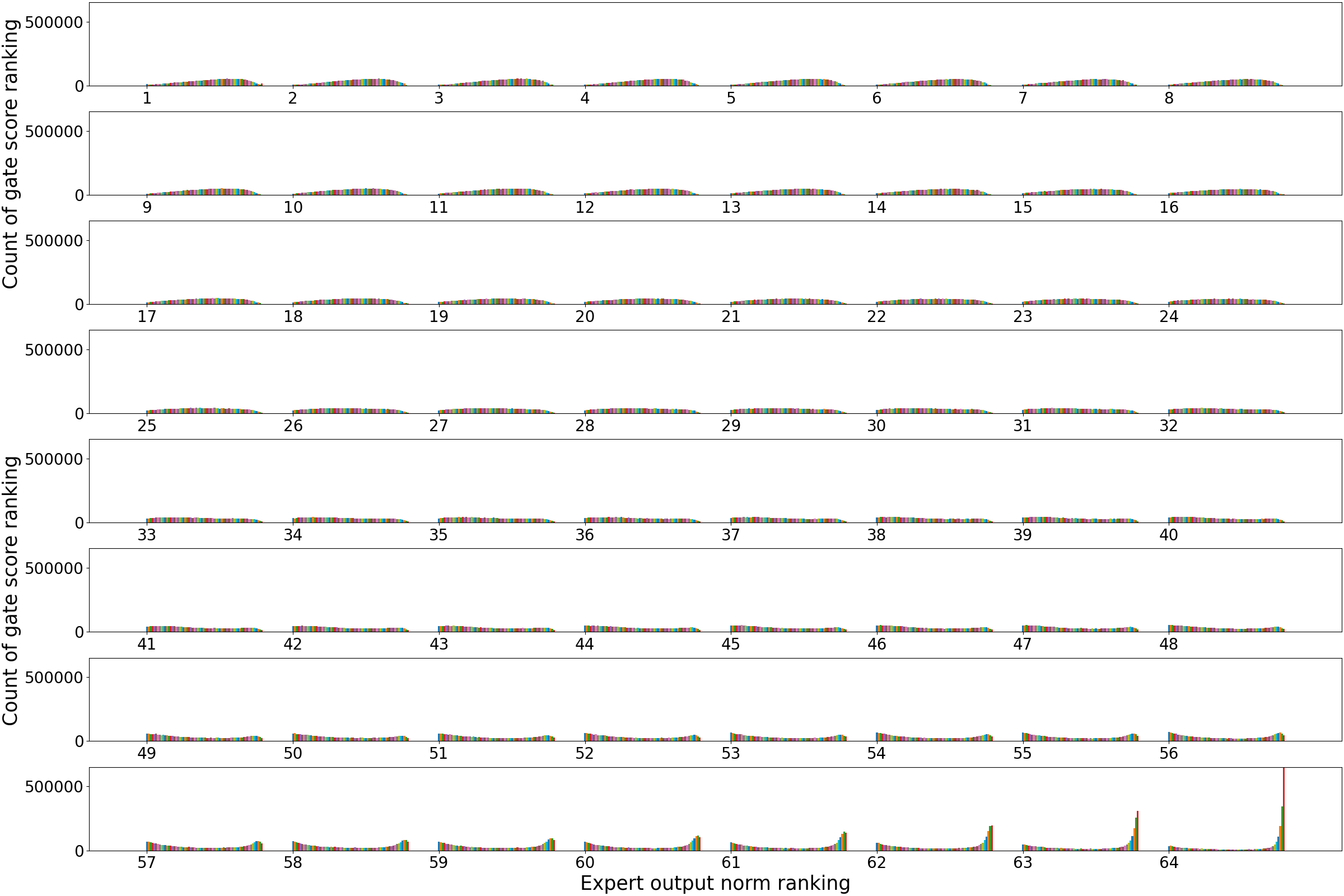} \\
    \vspace{2mm}
    \includegraphics[width=.7\linewidth]{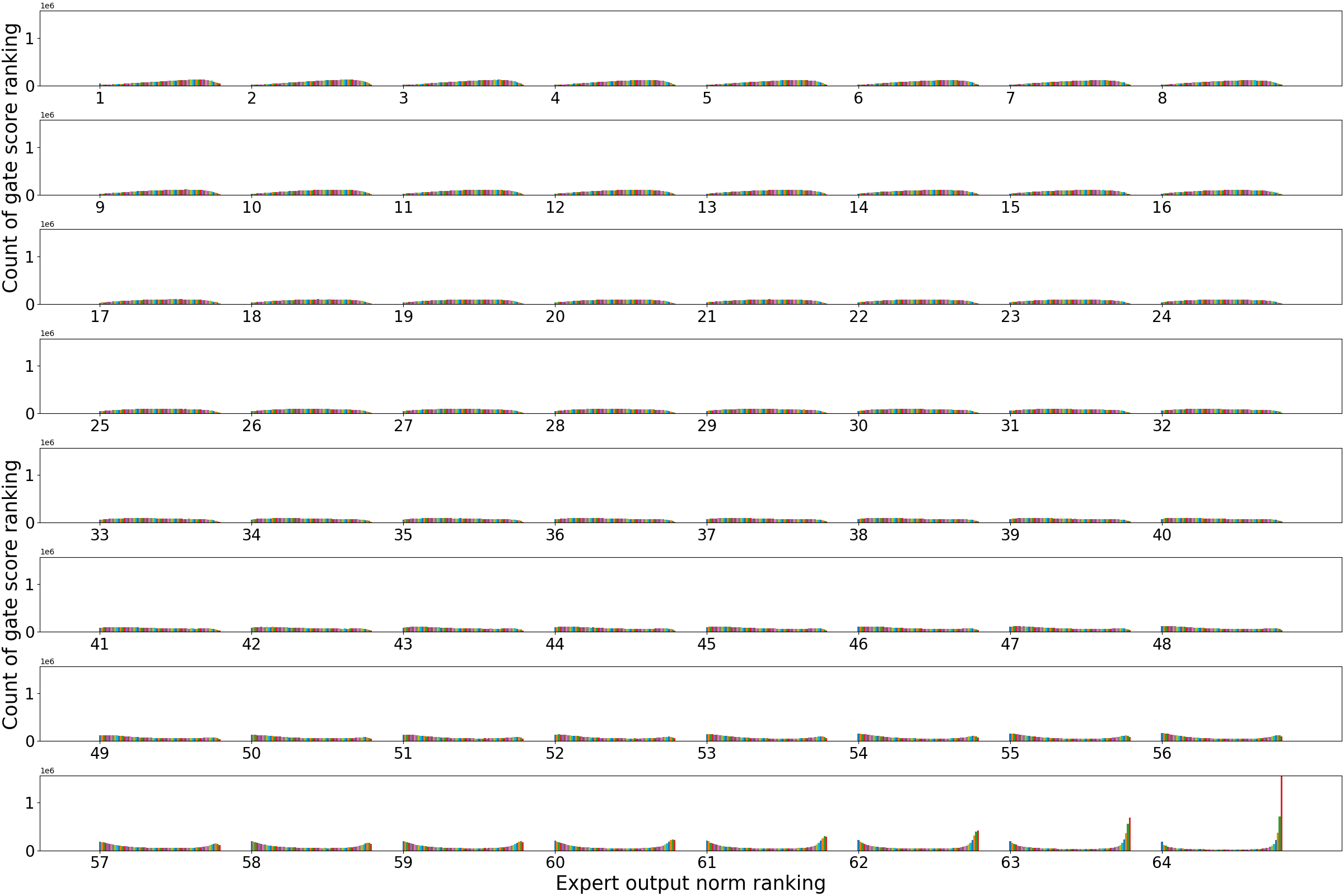}
    \caption{\footnotesize Counts of the gate score ranking for DeepSeek expert ourput norm rankings using additional datasets, namely WikiText-103 (\textbf{top}), GSM8K (\textbf{middle}), and Magicoder-Evol-Instruct-110K (\textbf{bottom}) The larger the rank number, the larger the norm or score.}
    \label{fig:reg_norm_deepseek}
\end{figure*}

In \S~\ref{exp:out_norm}, we notice that in some MoE models, the two experts chosen by the gate usually produce feature vectors with the highest norms.
To further investigate this, we repeat the experiment using the long input and additional datasets, and the statistical results are shown in Fig.~\ref{fig:reg_norm},~\ref{fig:reg_norm_mixtral},~\ref{fig:reg_norm_deepseek}.

\noindent\textbf{Mixtral.}
It is evident that the expert which outputs the largest norm is most frequently assigned the highest score.
Surprisingly, for every $i$, the $i$-th highest score is most likely assigned to the expert with the $i$-th highest output.

\noindent\textbf{DeepSeek.}
For the experts that generate the first few largest norms (rank 60$^{\text{th}}$ to 64$^{\text{th}}$), they are most likely to receive the highest scores.
But we do not observe a similar relationship for the rest of the experts.
On the contrary, the gate assigns relatively high scores more frequently than low scores to the experts with the smallest norms.
For experts ranked 49$^{\text{th}}$ to 59$^{\text{th}}$ in terms of output norms, they tend to receive either low scores or high scores.

\noindent\textbf{Grok.}
In contrast to the previous models, the output norms of the Grok experts tend to have an inverse relationship with the scores. 
More generally, the experts with the first few highest outputs are frequently assigned either low scores or high scores.
One possible explanation could be the relatively low activation ratios of GeLU (see Append~\ref{exp:intermediate}), which may result in a weaker dependence on the norm for gate decisions.

\begin{table}[thbp]
    \small
    \centering
    \begin{tabular}{lc}
        \toprule
        num\_layers & 24 \\
        vocab\_size & 151936 \\
        hidden\_size & 1024 \\
        head\_dim & 64 \\
        q\_head & 16 \\
        kv\_head & 4 \\
        moe\_hidden\_dim & 640 \\
        num\_shared\_expert & 4 \\
        num\_routed\_expert & 64 \\
        topk & 4 \\
        \bottomrule
    \end{tabular}
    \caption{\footnotesize Model architecture for the dynamic expert numbers experiment.}
    \label{tab:dynamic_moe_architecture}
\end{table}

\section{Intermediate States of Experts}
\label{exp:intermediate}

While \S~\ref{exp:out_sim} focused on the final outputs of experts, we continue our analysis here by examining their intermediate outputs to examine the inner states of the experts.
Given an input $x$, the intermediate state of an expert refers to the output of $\sigma(W_{\text{act}}x) \in \mathbb{R}^{d_{\text{hid}}} $, where $\sigma$ denotes an activation function.
These intermediate vectors control the activation of neurons, so we simply record them for analysis with the short input used. 
Mixtral, Mistral, and DeepSeek utilize SiLU as the activation function, while Grok adopts GeLU.
Fig.~\ref{fig:inter} depicts the magnitude of the vectors for Mixtral across three tokens. 

\noindent\textbf{Common.}
Each figure contains some horizontal lines, indicating the presence of an ``outlier'' expert with either the highest or lowest activation values. 
Nonetheless, there is no clear relationship between these phenomena and the gate decisions.
% Some horizontal lines are observed, meaning that most of the activation values of an expert are higher or lower than the other experts at the same time.
% This indices gate function can't capture the full 
% the horizontal lines seem to be irrelevant to the gate decision.

\noindent\textbf{Mixtral and Mistral.}
\label{growth}
For a single token, we found that, on average, the absolute activation value of 99.6\% elements in each expert exceeds 0.001 after applying the SiLU activation function.
This high ratio indicates that the vast majority of neurons in an expert are activated.
%, which might be a property of SiLU.
In Fig.~\ref{fig:inter}, some vertical lines across all experts are commonly found, meaning that the $W_{\text{act}}$ matrices of different experts assign similar activation values to neurons with the same indices. 
In addition, the magnitude of the intermediate states grows along with layer depth, which aligns with the observation in \S~\ref{exp:out_sim}.
% , which corresponds to the previous observation that the output norms of experts gradually increase (\S~\ref{exp:out_norm}).
% This is probably due to larger gradient the deep layers encountered during back-propagation.

\noindent\textbf{DeepSeek.}
On average, 99.7\% of the neurons in each expert have an absolute activation value exceeding 0.001 after applying SiLU.
Vertical lines rarely exist in the DeepSeek model.
Similarly, the elements in the intermediate state vectors get larger as the layer goes deeper.

\noindent\textbf{Grok.}
With GeLU as the activation function, only 25.3\% neurons per Grok expert attain an absolute activation value greater than 0.001. 
The activation values are generally smaller than those in Mixtral and DeepSeek.
\citet{li2022lazy, song2024prosparse} suggest such difference largely stems from the distinct activation functions used.
% which might also due to the property difference of activation function. 
Interestingly,~\citet{song2024turbo} further utilize the sparsity in experts within SMoE to achieve SOTA performance when activating the same number of parameters.

\section{Chosen Experts}
\label{exp:chosen_exp}

\begin{figure*}[htbp]
    \centering
    $\vcenter{\hbox{\includegraphics[width=.1\linewidth]{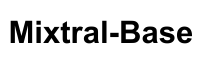} }}$ \\
    $\vcenter{\hbox{\includegraphics[width=.7\linewidth]{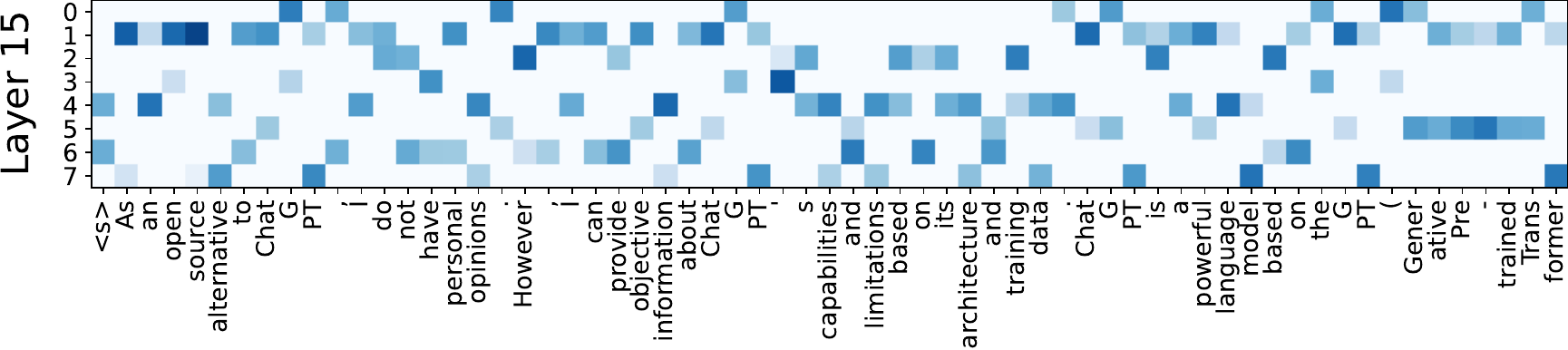} }}$ \\
    $\vcenter{\hbox{\includegraphics[width=.13\linewidth]{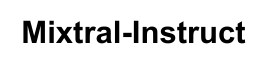} }}$ \\
    $\vcenter{\hbox{\includegraphics[width=.7\linewidth]{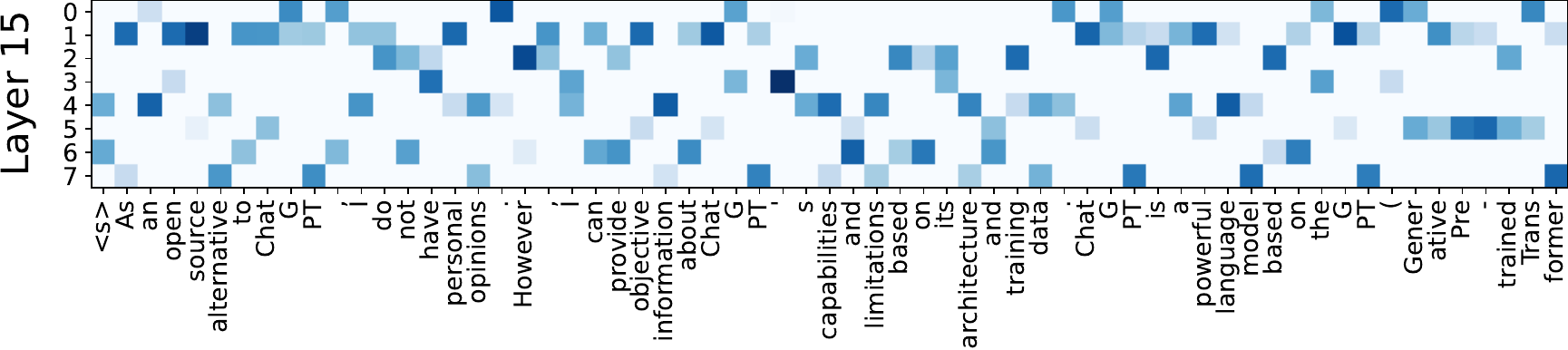} }}$ \\
    $\vcenter{\hbox{\includegraphics[width=.09\linewidth]{grok.pdf} }}$ \\
    $\vcenter{\hbox{\includegraphics[width=.7\linewidth]{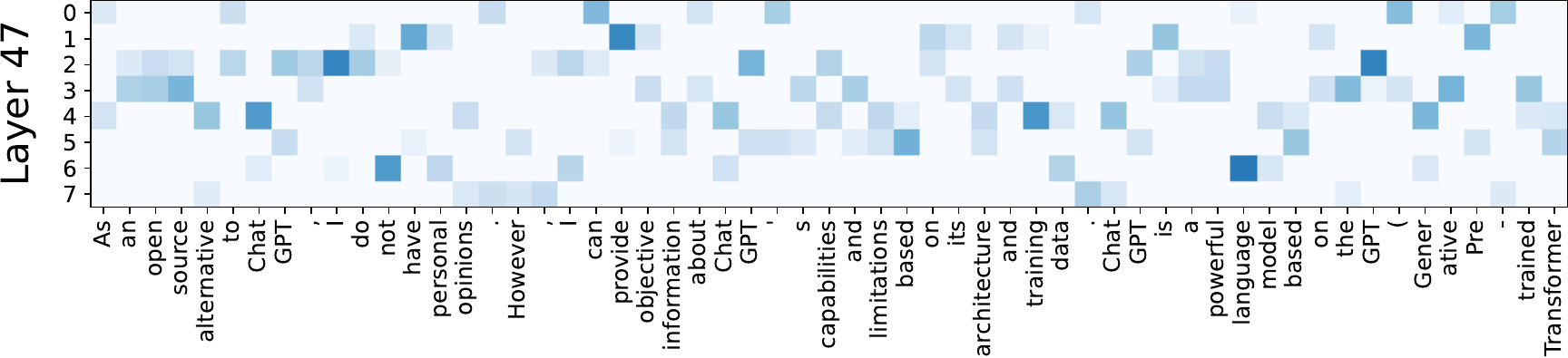} }}$ \\
    $\vcenter{\hbox{\includegraphics[width=.09\linewidth]{deepseek.pdf} }}$ \\
    $\vcenter{\hbox{\includegraphics[width=.68\linewidth]{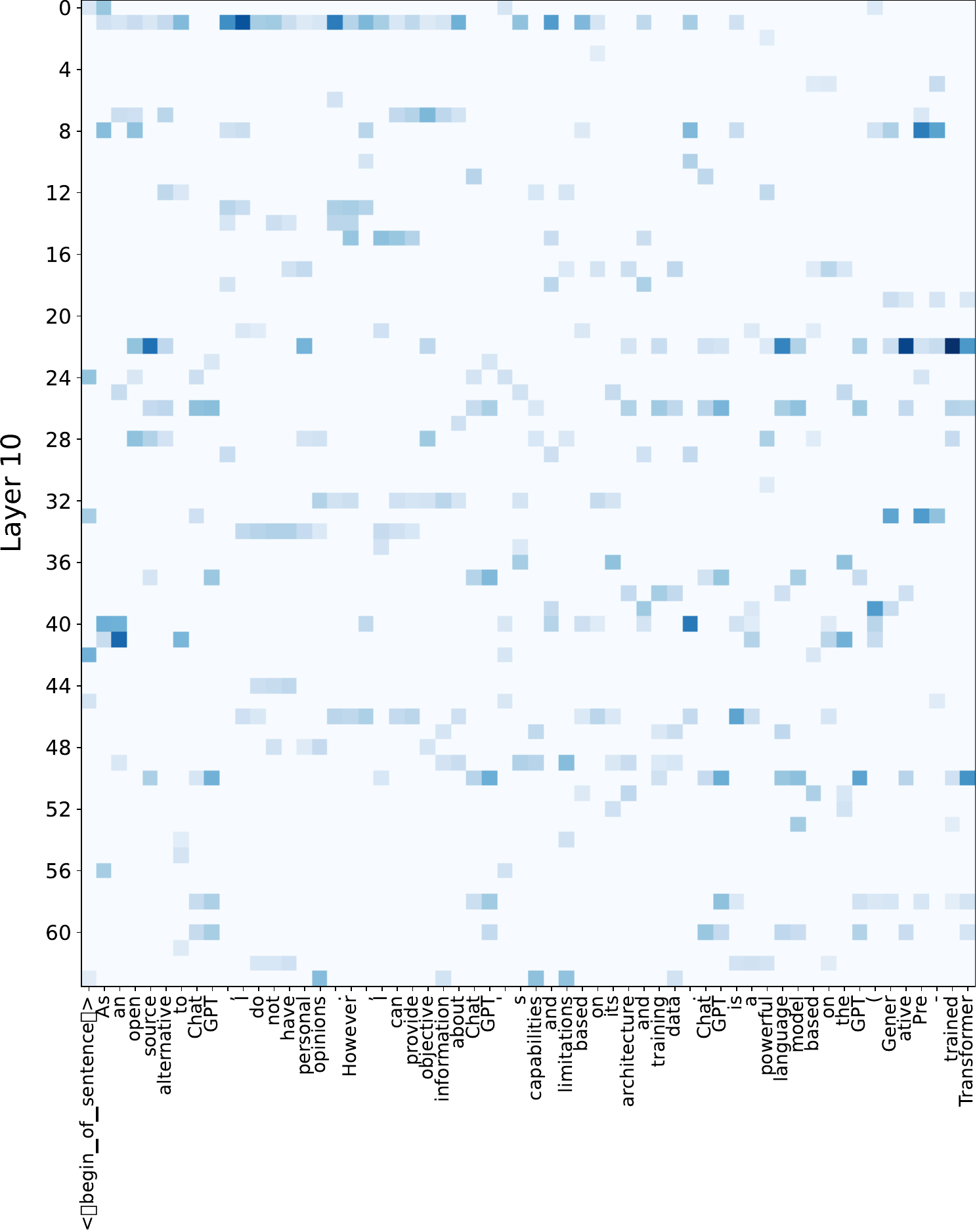} }}$ \\
    \caption{\footnotesize Routing patterns of different models. Deeper colors mean higher gate scores assigned to the corresponding experts. Only scores of the top $k$ experts are illustrated.}
    \label{fig:gate_choice}
\end{figure*}

This experiment aims to examine the routing patterns.
We feed an input prompt with about 64 tokens into the MoE models and record the gate scores (after applying softmax) for the selected experts for each token.
In addition to the base model of Mixtral (Mixtral-Base), we also include its instruct version (Mixtral-Instruct) in this experiment.
The results are depicted in Fig.~\ref{fig:gate_choice}.

\noindent\textbf{Mixtral.}
In Mixtral-Base, the experts are selected fairly evenly across tokens, and it is common to see sequences of more than four tokens routed to the same expert.
But the ``special expert'' with the dark cross in previous similarity graphs turns out to be an exception.
These special experts are chosen less frequently and tend to receive relatively low scores.
The routing pattern of Mixtral-Instruct is largely identical to that of Mixtral-Base, which indicates fine-tuning has little impact on gate decisions.

\noindent\textbf{DeepSeek.}
In some layers, there is an expert selected by most tokens.
However, no distinct characteristics for these experts are observed in the previous similarity heat maps.
Note that the gate scores for DeepSeek are typically lower than those for Mixtral because DeepSeek applies softmax before the top-k operation, while Mixtral adopts the reverse way.

\noindent\textbf{Grok.}
The expert selection is rather even and some relatively high scores exist in the deeper (>30$^{\text{th}}$) layers.
Same as DeepSeek, softmax is applied before the top-k operation for Grok.

\end{document}